\documentclass[journal]{IEEEtran}

\usepackage{amsmath,amssymb,amsfonts,amsthm}
\usepackage{bm}
\usepackage{graphicx}
\usepackage{stfloats}
\usepackage{cite}
\usepackage{booktabs}
\usepackage{multirow}
\usepackage{array}
\usepackage{algorithm}
\usepackage{algorithmic}
\usepackage{xcolor}
\usepackage{url}
\usepackage{hyperref}
\usepackage{makecell}
\newtheorem{definition}{Definition}
\newtheorem{theorem}{Theorem}
\newtheorem{lemma}{Lemma}

\newtheorem{remark}{Remark}
\newtheorem{assumption}{Assumption}

\newcommand{\R}{\mathbb{R}}

\newcommand{\X}{\mathcal{X}}
\newcommand{\Y}{\mathcal{Y}}

\newcommand{\OmegaSet}{\Omega}

\newcommand{\argmin}{\operatorname*{arg\,min}}
\newcommand{\WSpTFI}{\texorpdfstring{WS$_{p}$TF-I}{WSpTF-I}}
\newcommand{\WSpTFII}{\texorpdfstring{WS$_{p}$TF-II}{WSpTF-II}}
\newcommand{\WSpMFI}{\texorpdfstring{WS$_{p}$MF-I}{WSpMF-I}}
\newcommand{\WSpMFII}{\texorpdfstring{WS$_{p}$MF-II}{WSpMF-II}}

\begin{document}

\setcounter{topnumber}{8}
\setcounter{dbltopnumber}{8}
\setcounter{totalnumber}{12}
\renewcommand{\topfraction}{0.98}
\renewcommand{\dbltopfraction}{0.98}
\renewcommand{\textfraction}{0.02}
\renewcommand{\floatpagefraction}{0.90}
\renewcommand{\dblfloatpagefraction}{0.90}

\title{Robust Low-Rank Tensor Completion via Factorized Weighted Tensor Schatten-$p$ Norm Minimization}

\author{
Binghao Wang, Feng Zhang, Wendong Wang, Jianjun Wang, Member, IEEE%
\thanks{This work was supported in part by the National Key Research and Development Program of China under Grant 2023YFA1008502; in part by Fundamental Research Funds for the Central Universities under Grant SWU-KR25013; and in part by National Natural Science Foundation of China under Grant 12101512. (Corresponding author: Feng Zhang.)}
\thanks{Binghao Wang, Feng Zhang,  and Jianjun Wang are with the School of Mathematics and Statistics, Southwest University, Chongqing 400715, China (e-mail:  2496961443@qq.com, zfmath@swu.edu.cn, wjj@swu.edu.cn).}
\thanks{Wendong Wang is with the College of Artificial Intelligence, Southwest University, Chongqing 400715, China.}
}

\maketitle

\begin{abstract}
Low-rank tensor factorization provides a flexible framework for completing
multidimensional data from incomplete and corrupted observations. However,
unweighted spectral regularizers impose a common shrinkage profile across
singular components, which may excessively attenuate dominant low-rank
components, and factorized variants either lack component-specific weighting
or require costly singular value decompositions (SVDs). This paper proposes
two weighted Schatten-$p$ tensor factorization models, termed \WSpTFI{} and
\WSpTFII{}, under the tensor-tensor product (t-product) framework to address
these limitations. \WSpTFI{} is motivated by a factorized weighted tensor
Schatten-$p$ norm identity and permits flexible, possibly asymmetric factor
exponents. \WSpTFII{} constructs a regularizer from transform-domain
column-pair energies, yielding SVD-free main factor updates and a
column-pruning mechanism for reducing redundant rank components. This paper further develops an iteratively
reweighted alternating direction method of multipliers (ADMM)-type scheme for
\WSpTFI{} and an iteratively reweighted least squares (IRLS)--block successive
upper-bound minimization (BSUM) scheme for \WSpTFII{}. Theoretical analysis
establishes the weighted factorization relation and provides a conditional
limiting Karush--Kuhn--Tucker (KKT)
characterization for \WSpTFI{} under the stated boundedness, exact-update, and
factor-block residual assumptions. For \WSpTFII{}, the actual damped quadratic block updates yield a
quantitative sufficient-decrease mechanism for the fixed-$\delta$ smoothed
factor objective. This implies asymptotic regularity, and every accumulation
point of the fixed-dimensional tail is stationary after pruning becomes
inactive.
Experiments on synthetic tensor completion, color-image restoration,
hyperspectral inpainting, and printed-circuit-board defect detection
demonstrate competitive reconstruction quality and robustness under various degradation conditions.
\end{abstract}


\section{Introduction}

Low-rank modeling assumes that high-dimensional observations can be represented by a small number of latent degrees of freedom. For matrices, this principle has given rise to a spectrum of methods, from rank minimization and its nuclear-norm relaxation to nonconvex Schatten-$p$ surrogates and factorized formulations \cite{RechtFazelParrilo2010,NieEtAl2012Schatten,
ShangEtAl2016Tractable,XuEtAl2017UnifiedConvex}, which together lay the foundation for low-rank tensor completion. For third-order data, the t-product and the
associated tensor singular value decomposition (t-SVD) provide an
algebraic framework in which low-rank structure is characterized by
the tubal rank \cite{KilmerMartin2011,KernfeldEtAl2015}. Given an observed
tensor $\Y$, an index set $\Omega$ of available entries, and the projection
operator $\mathcal P_{\Omega}$ onto those entries, the ideal low-tubal-rank
tensor completion problem can be written as
\begin{equation}
\min_{\X}\ \operatorname{rank}_t(\X)
\qquad
\text{s.t.}\qquad
\mathcal P_{\Omega}(\X)=\mathcal P_{\Omega}(\Y) ,
\label{eq:intro-tubal-rank}
\end{equation}
where $\operatorname{rank}_t(\cdot)$ denotes the tubal rank. Because direct tubal-rank minimization is computationally intractable in general, the tensor nuclear norm (TNN) was proposed as a convex surrogate for the tubal rank, computed from the singular values of the transform-domain frontal slices
\cite{ZhangAeron2017,LuEtAl2020TRPCA}.

Under the discrete Fourier transform-based t-product framework, the TNN is defined as
\begin{equation}
\|\X\|_*
=
\frac{1}{n_3}
\sum_{i=1}^{n_3}
\sum_{j}
\sigma_j\!\left(\bar{\X}^{(i)}\right).
\label{eq:intro-tnn}
\end{equation}
where \(\bar{\X}^{(i)}\) denotes the \(i\)th frontal slice of the
transform-domain representation and \(\sigma_j(\cdot)\) denotes the
\(j\)th singular value of a matrix. Although \eqref{eq:intro-tnn}
is convex and tractable, its linear treatment of the singular spectrum
can over-penalize dominant components. A closer nonconvex approximation
to the tubal rank is provided by the tensor Schatten-$p$ norm with $0<p<1$,
\begin{equation*}
\|\X\|_{S_p}
=
\left(
\frac{1}{n_3}
\sum_{i=1}^{n_3}
\sum_{j}
\sigma_j\!\left(\bar{\X}^{(i)}\right)^p
\right)^{1/p},
\end{equation*}
which inherits the rank-approximating behavior of the matrix Schatten-$p$
quasi-norm \cite{NieEtAl2012Schatten,ShangEtAl2016Schatten,
XuEtAl2017UnifiedConvex}. Nonuniform spectral weighting leads to the weighted
tensor Schatten-$p$ norm
\begin{equation}
\|\X\|_{W,S_p}
=
\left(
\frac{1}{n_3}
\sum_{i=1}^{n_3}
\sum_{j}
w_{ij}\,
\sigma_j\!\left(\bar{\X}^{(i)}\right)^p
\right)^{1/p} .
\label{eq:intro-weighted-tensor-schatten}
\end{equation}
where smaller $p$ yields a tighter approximation to the tubal rank and the
weight $w_{ij}$ assigns a distinct shrinkage strength to each singular
component. Weighted and iteratively reweighted spectral regularizers based on
\eqref{eq:intro-weighted-tensor-schatten} have been widely used to reduce the
bias caused by applying the same shrinkage profile to all singular components
\cite{GuEtAl2014WNNM,LuEtAl2016IRNN,XieEtAl2016WSNM,
YangEtAl2022WeightedTensorSchatten}. Equations \eqref{eq:intro-tnn}--
\eqref{eq:intro-weighted-tensor-schatten} thus trace a progression from convex
nuclear regularization to weighted nonconvex spectral modeling. Such
component-wise weighting is especially useful in robust tensor completion,
where missing entries and sparse gross corruption affect different spectral
components unevenly. A useful model should therefore retain dominant low-rank
structure without shrinking all singular values according to the same rule.
Since the spectral regularizers in \eqref{eq:intro-tnn}--\eqref{eq:intro-weighted-tensor-schatten} incur per-slice spectral
computations, the optimization should also remain affordable as the spatial
dimensions and the number of transform slices grow. These two requirements
motivate combining weighted nonconvex spectral regularization with compact
tensor factorization, whose small factors reduce the storage and avoid
spectral computations on the full tensor, rather than treating them as
separate design choices.

A practical limitation of spectral regularizers of the form
\eqref{eq:intro-tnn}--\eqref{eq:intro-weighted-tensor-schatten} is that
directly optimizing them typically requires repeated singular value
decompositions (SVDs) \cite{NieEtAl2012Schatten,
ShangEtAl2016Tractable,XuEtAl2017UnifiedConvex}.
Factorization provides an alternative by representing the unknown tensor
through compact latent factors. In the matrix setting, variational identities
connect nuclear and Schatten-type spectral regularization with regularization
on bilinear factors
\cite{SrebroEtAl2004MMMF,CabralEtAl2013,HastieEtAl2015ALS,
ShangEtAl2020Unified,XuEtAl2017UnifiedConvex}. Under the discrete Fourier transform-based t-product framework, a
two-factor representation of the tensor is
$\X=\mathcal U*\mathcal V\in\R^{n_1\times n_2\times n_3}$, where
$\mathcal U\in\R^{n_1\times r\times n_3}$ and
$\mathcal V\in\R^{r\times n_2\times n_3}$ are factor tensors and $r$ denotes
the factor width. Any $r\ge\operatorname{rank}_t(\X)$. Such compact factorizations have been used for low-tubal-rank completion and related restoration tasks \cite{ZhouEtAl2018TensorFactorization,ChenEtAl2022RLTF}, with the completion setting directly corresponding to \eqref{eq:intro-tubal-rank}; recent works have advanced this direction further. Jiang \emph{et al.} \cite{JiangEtAl2023FactorTensorNorm} constructed factor-tensor norms equivalent to particular Schatten quasi-norms for low-tubal-rank completion, He and Atia \cite{HeAtia2023Correntropy} combined low-tubal-rank factorization with maximum correntropy to improve robustness against large outliers, and Liu \emph{et al.} \cite{LiuEtAl2024FGD} developed a factorized-gradient method with theoretical guarantees under exact and moderately overestimated rank settings. While these methods demonstrate the value of compact tensor factors, they target different aspects of the problem, namely specific Schatten exponents, robust data fidelity, or efficient parameterization. None of them, however, provides a factorized weighted tensor Schatten-$p$ norm identity with flexible factor exponents or a column-pair regularizer connecting SVD-free updates with component-wise rank reduction, both of which are central to the models developed in this paper.

Beyond reducing spectral computation, factorization also provides variational characterizations of tensor spectral regularizers. For the TNN defined in \eqref{eq:intro-tnn}, a classical factorization identity yields
\begin{equation}
\|\X\|_*
=
\frac{1}{2}
\min_{\X=\mathcal U*\mathcal V}
\left(
\|\mathcal U\|_{F}^{2}
+
\|\mathcal V\|_{F}^{2}
\right),
\label{eq:intro-tensor-nuclear-factorization}
\end{equation}
which provides a direct bridge between tensor spectral regularization and
compact factor models \cite{DuEtAl2021UTF}. More
generally, for $p,p_1,p_2>0$ satisfying
$1/p=1/p_1+1/p_2$, the two-factor Schatten relation becomes
\begin{equation}
\frac{1}{p}\|\X\|_{S_p}^{p}
=
\min_{\X=\mathcal U*\mathcal V}
\left\{
\frac{1}{p_1}
\|\mathcal U\|_{S_{p_1}}^{p_1}
+
\frac{1}{p_2}
\|\mathcal V\|_{S_{p_2}}^{p_2}
\right\}.
\label{eq:intro-two-factor-tensor-schatten}
\end{equation}
Equation~\eqref{eq:intro-tensor-nuclear-factorization} is recovered by
$p=1$ and $p_1=p_2=2$ using the scaled Parseval relation of the adopted
transform, while asymmetric choices of $p_1$ and $p_2$ are
also permitted \cite{ShangEtAl2020Unified,XuEtAl2017UnifiedConvex}.  Nevertheless, \eqref{eq:intro-two-factor-tensor-schatten} is unweighted and hence cannot provide the shrinkage represented by \eqref{eq:intro-weighted-tensor-schatten}. This motivates our first formulation, weighted Schatten-$p$ tensor factorization I
(\WSpTFI{}), which establishes a factorized weighted tensor Schatten-$p$ norm
relation with flexible, possibly asymmetric factor exponents. The resulting model preserves a direct connection between
the weighted spectrum of the reconstructed tensor and the spectral norm regularizers
on its compact factors.

Even though factorization reduces the variable dimensions, Schatten norm regularizers on the factors may still require factor SVDs \cite{ShangEtAl2020Unified,GiampourasEtAl2020Variational}. Column-separable variational formulations avoid this by regularizing paired factor components directly \cite{GiampourasEtAl2019AIRLS,FanEtAl2019FGSR,
GiampourasEtAl2020Variational,OrnhagEtAl2020Bilinear}. Motivated by this perspective, our second formulation, weighted Schatten-$p$
tensor factorization II (\WSpTFII{}), extends paired-column Schatten
regularization to t-product tensor factors. Its exact column-pair representation replaces factor-spectral norm regularizers with transform-domain column-pair energies, yielding SVD-free main factor updates and a natural criterion for pruning redundant factor components. In contrast to nonconvex sparsity-inducing penalties \cite{WangEtAl2024SparsityTensor}, \WSpTFII{} is derived from an exact paired-column variational characterization, while \WSpTFI{} retains an explicit weighted spectral interpretation, making the two formulations complementary.

\begin{figure}[!t]
  \centering
  \includegraphics[width=\columnwidth]{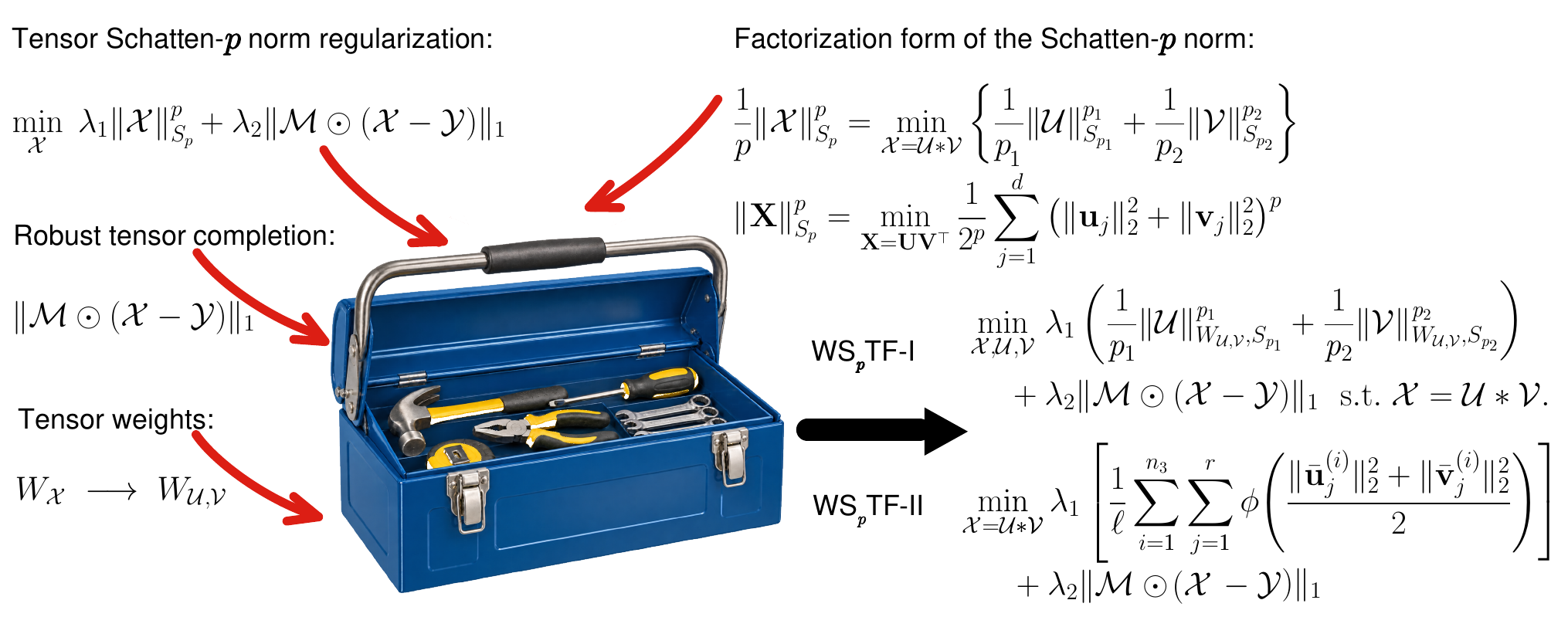}
  \caption{Overview of the complementary \WSpTFI{} and \WSpTFII{} formulations for robust tensor completion.}
  \label{fig:wsptf_overview}
\end{figure}

As illustrated in Fig.~\ref{fig:wsptf_overview}, the proposed framework starts from a common low-tubal-rank factorization and then branches into two complementary regularization strategies, corresponding to weighted factor-spectral regularization in \WSpTFI{} and paired-component regularization with rank pruning in \WSpTFII{}. The main contributions are summarized as follows.
\begin{itemize}
  \item We establish a factorized weighted tensor Schatten-$p$ norm identity with flexible factor exponents, which yields \WSpTFI{} and preserves a direct connection between the weighted spectrum of the completed tensor and the spectral norm regularizers on its compact factors.

  \item We establish a column-pair weighted Schatten-$p$ tensor factorization identity based on transform-domain column-pair energies, which yields \WSpTFII{} and admits SVD-free main factor updates and supports finite pruning of redundant rank components.

  \item We develop an iteratively reweighted alternating direction method of
  multipliers (ADMM)-type scheme for \WSpTFI{} and an iteratively reweighted
  least squares (IRLS)--block successive upper-bound minimization (BSUM) scheme
  for \WSpTFII{}. Extensive experiments on synthetic tensor completion and
  representative visual tensor completion tasks show that the proposed
  formulations achieve consistently strong completion quality and robustness,
  with our two algorithms attaining leading accuracy in several challenging
  settings.

  \item We develop convergence
  guarantees for \WSpTFI{}
  and \WSpTFII{}: a conditional limiting Karush--Kuhn--Tucker (KKT)
  characterization of the frozen-weight
  problem for
  \WSpTFI{}, and monotonic descent with stationary accumulation points for the
  fixed-$\delta$ smoothed objective of \WSpTFII{}.
  
\end{itemize}

The remainder of the paper is organized as follows. Section~\ref{sec:related-work} reviews
closely related spectral and factorized low-rank models. Section~\ref{sec:notations}
introduces the tensor algebra and spectral functionals used in the paper.
Sections~\ref{sec:wsptf} and~\ref{sec:vwsptf} present \WSpTFI{} and \WSpTFII{}, respectively.
Section~\ref{sec:experiments} reports the experiments, and Section~\ref{sec:conclusion} concludes the paper.

\section{Related Work}
\label{sec:related-work}

\subsection{Schatten-$p$ and Weighted Norms for Robust Matrix and Tensor Completion}

Robust low-rank recovery aims to infer a low-dimensional component from
incomplete observations that may additionally contain dense noise or sparse
gross corruption. For matrices, the nuclear norm is the standard convex
surrogate for rank, but its linear treatment of the singular spectrum applies
the same shrinkage rule to all singular directions and can therefore bias the
dominant components \cite{CandesEtAl2011RPCA}. Nonconvex Schatten-$p$
penalties with $0<p<1$ provide a closer approximation to rank and have been
studied in low-rank recovery, minimum-rank approximation, and low-rank
representation \cite{NieEtAl2012Schatten,LiuEtAl2014ExactRank,
ZhangEtAl2018LRRSchatten,Foucart2018ConcaveMirsky}. Their tighter spectral
approximation can better preserve large singular values, although the
resulting problems are nonconvex and typically retain an SVD-based
computational burden.

A complementary strategy is to make the spectral shrinkage itself
nonuniform. Weighted nuclear-norm minimization assigns different penalties to
different singular values \cite{GuEtAl2014WNNM}, while reweighted low-rank
schemes update those penalties from the current estimate
\cite{PengEtAl2014Reweighted,LuEtAl2016IRNN,HuangEtAl2020GIRNN}. Combining
nonuniform weights with a nonconvex Schatten exponent leads to weighted
Schatten-$p$ regularization, which has been used in image denoising and
background subtraction \cite{XieEtAl2016WSNM}. These models form a useful
continuum: the exponent controls the nonconvex approximation to rank, whereas
the weights adapt the shrinkage across singular directions.

The same ideas extend naturally to multidimensional data. The t-product
framework can be induced by a general invertible linear transform along the
third mode \cite{KernfeldEtAl2015}; under the standard scaled-orthogonality
condition on the transform matrix, transform-induced TNNs and
recovery models can be defined consistently with the Frobenius geometry
\cite{LuPengWei2019TransformTNN}. Under this transform-based t-SVD/t-product
framework, low tubal rank is described through the spectra of transform-domain
frontal slices, leading to tensor nuclear-norm completion and tensor robust principal
component analysis \cite{ZhangAeron2017,LuEtAl2020TRPCA}. Weighted tensor
Schatten-$p$ regularization further combines nonuniform and nonconvex
shrinkage across these slice spectra \cite{YangEtAl2022WeightedTensorSchatten},
while truncated tensor Schatten-$p$ penalties provide another nonconvex
alternative for completion \cite{LiuEtAl2025TTSP}. These spectral approaches
are closely related to the modeling objective of this work, but their direct
optimization still acts on singular spectra. Repeated SVDs can therefore
become a dominant cost, motivating compact factorized representations.

\subsection{Low-Rank Factorization and Variational Schatten-$p$ Models}

Factorization addresses this issue by parameterizing a low-rank matrix as
$\mathbf X=\mathbf U\mathbf V^\top$ and optimizing over smaller factors. The
classical variational characterization of the nuclear norm establishes a
direct connection between spectral regularization and
Frobenius-regularized bilinear factorization. Early low-norm and
maximum-margin formulations used this viewpoint in collaborative prediction
\cite{SrebroEtAl2004MMMF,RennieSrebro2005}, and it was subsequently adopted
in matrix completion, robust low-rank decomposition, and online robust principal component analysis
\cite{WenEtAl2012LowRankFactorization,CabralEtAl2013,
FengEtAl2013ORPCA,ShangEtAl2015RBF}. Alternating least-squares methods further
showed that explicit low-rank factors can be optimized efficiently for large
incomplete matrices \cite{HastieEtAl2015ALS}, while theoretical studies
established recovery and optimality properties for nonconvex factorized
models \cite{SunLuo2016Factorization,HaeffeleVidal2020}.

The variational connection extends beyond the nuclear norm. Bi-trace and
tri-trace penalties are equivalent to the Schatten-$1/2$ and
Schatten-$1/3$ quasi-norms, respectively
\cite{ShangEtAl2016Schatten,ShangEtAl2016Tractable}. More generally, a
Schatten-$p$ penalty can be represented through two factor penalties whose
exponents satisfy $1/p=1/p_1+1/p_2$
\cite{XuEtAl2017UnifiedConvex}; analogous multi-factor
extensions satisfy $1/p=\sum_k1/p_k$ and can place individual factor
penalties in smoother Schatten regimes \cite{ShangEtAl2020Unified}. Such identities reduce the
dimensions on which spectral operations are performed, but they do not
necessarily eliminate factor SVDs.

This remaining bottleneck has motivated column-wise variational models.
Giampouras \emph{et al.} \cite{GiampourasEtAl2019AIRLS} proposed an
alternating IRLS formulation in which a joint column-sparsity regularizer is
closely related to a weighted nuclear penalty. Factor group-sparse regularization
then connected column sparsity with Schatten-$p$ penalties and enabled
SVD-free recovery for particular exponent families \cite{FanEtAl2019FGSR}.
A more general column-separable representation valid for every
$p\in(0,1]$ was subsequently derived in
\cite{GiampourasEtAl2020Variational}. Related bilinear parameterizations
convert broader singular-value regularizers into differentiable factor
objectives \cite{OrnhagEtAl2020Bilinear}, while column $\ell_{2,0}$
regularization provides another factor-space mechanism for promoting low
rank \cite{TaoEtAl2022ColumnL20}. These developments illustrate the tradeoff
between spectral fidelity, factor separability, rank adaptivity, and the need
for explicit SVDs.

Recent work has transferred these ideas directly to low-tubal-rank tensor
recovery. Jiang \emph{et al.} \cite{JiangEtAl2023FactorTensorNorm}
introduced factor tensor norms equivalent to the Schatten-$1/2$ and
Schatten-$2/3$ quasi-norms, replacing large spectral problems by smaller
factor problems. He and Atia \cite{HeAtia2023Correntropy} combined
low-tubal-rank factorization with maximum correntropy to improve robustness
to large outliers. Liu \emph{et al.} \cite{LiuEtAl2024FGD} developed
factorized gradient descent and analyzed exact-rank as well as
over-parameterized recovery. In a related nonconvex direction, Wang
\emph{et al.} \cite{WangEtAl2024SparsityTensor} proposed sparsity-inducing
low-tubal-rank regularizers with closed-form thresholding rules.

Table~\ref{tab:related_factorization_comparison} provides a compact
comparison of representative factorization and variational models discussed
above that admit an explicit factorized representation of a low-rank norm,
quasi-norm, rank surrogate, or spectral regularizer. When several methods use
the same factorization identity, only one representative method is retained
to avoid redundant formulas. RLTF~\cite{ChenEtAl2022RLTF} is included because
it factorizes a reweighted tensor nuclear norm into reweighted Frobenius norms
of two compact tensor factors. The second column reports the corresponding
factorized form rather than the full recovery optimization problem. A check
mark indicates that the corresponding property is explicitly supported by
the model or algorithm. Here, ``SVD-free'' means that the main iterative
factor updates do not require an explicit matrix SVD or t-SVD.

\begin{table*}[!t]
\centering
\caption{Comparison of representative factorized low-rank norm and variational models discussed in Section II-B.}
\label{tab:related_factorization_comparison}
\scriptsize
\renewcommand{\arraystretch}{1.18}
\setlength{\tabcolsep}{2.2pt}
\begin{tabular}{
  @{}
  >{\raggedright\arraybackslash}p{0.190\textwidth}
  >{\raggedright\arraybackslash}p{0.500\textwidth}
  cccc
  @{}
}
\toprule
\textbf{Method}
&
\makecell{\textbf{Norm / regularizer}\\\textbf{factorization}}
&
\makecell{\textbf{Flexible}\\\textbf{Schatten-$p$}}
&
\textbf{Weighted}
&
\textbf{SVD-free}
&
\makecell{\textbf{Rank}\\\textbf{adaptive}}
\\
\midrule

Cabral \emph{et al.}~\cite{CabralEtAl2013}
&
\(
\begin{aligned}
\|\mathbf X\|_*
&=
\min_{\mathbf X=\mathbf U\mathbf V^\top}
\frac12\bigl(
\|\mathbf U\|_F^2+\|\mathbf V\|_F^2
\bigr)
\end{aligned}
\)
&  &  &  &  \\\addlinespace[2pt]

RBF~\cite{ShangEtAl2015RBF}
&
\(
\begin{aligned}
\|\mathbf L\|_*
&=\|\mathbf U\mathbf V^\top\|_*=\|\mathbf V\|_*,
\mathbf U^\top\mathbf U=\mathbf I
\end{aligned}
\)
&  &  &  & $\checkmark$ \\\addlinespace[2pt]

Bi-trace / Tri-trace~\cite{ShangEtAl2016Tractable}
&
\(
\begin{aligned}
\|\mathbf X\|_{S_{1/2}}
&=
\min_{\mathbf X=\mathbf U\mathbf V^\top}
\|\mathbf U\|_*\|\mathbf V\|_*,
\\
\|\mathbf X\|_{S_{1/3}}
&=
\min_{\mathbf X=\mathbf U\mathbf V\mathbf W^\top}
\|\mathbf U\|_*\|\mathbf V\|_*\|\mathbf W\|_*
\end{aligned}
\)
&
& & &
\\
\addlinespace[2pt]

F/N hybrid~\cite{ShangEtAl2016Schatten}
&
\(
\begin{aligned}
\|\mathbf X\|_{S_{2/3}}
&=
\min_{\mathbf X=\mathbf U\mathbf V^\top}
\|\mathbf U\|_F\|\mathbf V\|_*
\end{aligned}
\)
&
& & &
\\
\addlinespace[2pt]

Bi-Schatten-\(p\) norm surrogate~\cite{XuEtAl2017UnifiedConvex}
&
\(
\begin{aligned}
\frac1p\|\mathbf X\|_{S_p}^{p}
&=
\min_{\mathbf X=\mathbf U\mathbf V^\top}
\left(
\frac1{p_1}\|\mathbf U\|_{S_{p_1}}^{p_1}
+
\frac1{p_2}\|\mathbf V\|_{S_{p_2}}^{p_2}
\right),
\frac1p=\frac1{p_1}+\frac1{p_2}
\end{aligned}
\)
&
$\checkmark$ & & &
\\
\addlinespace[2pt]

Unified scalable Schatten formulation~\cite{ShangEtAl2020Unified}
&
\(
\begin{aligned}
\|\mathbf X\|_{S_p}
&=
\min_{\mathbf X=\mathbf U\mathbf V^\top}
\|\mathbf U\|_{S_{p_1}}\|\mathbf V\|_{S_{p_2}},
\frac1p=\frac1{p_1}+\frac1{p_2}
\end{aligned}
\)
&
$\checkmark$ & & &
\\
\addlinespace[2pt]

AIRLS-MC~\cite{GiampourasEtAl2019AIRLS}
&
\(
\begin{aligned}
\|\mathbf X\|_{*,\mathbf w}
&\le
\frac12\sum_{j=1}^{d}
\bigl(
\|\mathbf u_j\|_2^2+\|\mathbf v_j\|_2^2
\bigr)^{p/2},
\mathbf X=\mathbf U\mathbf V^\top
\end{aligned}
\)
&  & $\checkmark$ & $\checkmark$ & $\checkmark$ \\\addlinespace[2pt]

FGSR$_{1/2}$/FGSR$_{2/3}$~\cite{FanEtAl2019FGSR}
&
\(
\begin{aligned}
\|\mathbf X\|_{S_{1/2}}^{1/2}
&=
\frac12
\min_{\mathbf A\mathbf B=\mathbf X}
\bigl(
\|\mathbf A\|_{2,1}+\|\mathbf B^\top\|_{2,1}
\bigr),
\\
\|\mathbf X\|_{S_{2/3}}^{2/3}
&=
\frac{2}{3\alpha^{1/3}}
\min_{\mathbf A\mathbf B=\mathbf X}
\left(
\|\mathbf A\|_{2,1}
+\frac{\alpha}{2}\|\mathbf B\|_F^2
\right)
\end{aligned}
\)
&  &  & $\checkmark$ & $\checkmark$ \\\addlinespace[2pt]

Variational Schatten-\(p\) form~\cite{GiampourasEtAl2020Variational}
&
\(
\begin{aligned}
\|\mathbf X\|_{S_p}^{p}
&=
\min_{\mathbf X=\mathbf U\mathbf V^\top}
\frac1{2^p}
\sum_{j=1}^{d}
\bigl(
\|\mathbf u_j\|_2^2+\|\mathbf v_j\|_2^2
\bigr)^p
\end{aligned}
\)
&
$\checkmark$ & & $\checkmark$ & $\checkmark$
\\
\addlinespace[2pt]Column \(\ell_{2,0}\)-norm factorization~\cite{TaoEtAl2022ColumnL20}
&
\(
\begin{aligned}
\operatorname{rank}(\mathbf X)
&=
\min_{\mathbf X=\mathbf U\mathbf V^\top}
\frac12
\bigl(
\|\mathbf U\|_{2,0}+\|\mathbf V\|_{2,0}
\bigr)
\end{aligned}
\)
&
& & & $\checkmark$
\\
\addlinespace[2pt]

RLTF~\cite{ChenEtAl2022RLTF}
&
\(
\begin{aligned}
\|\X\|_{w,*}
&=
\min_{\X=\mathcal U*\mathcal V}
\frac12
\left(
\|\mathcal U\|_{w,F}^{2}
+
\|\mathcal V\|_{w,F}^{2}
\right),
\\[-1pt]
w_{ij}
&=
c\!\left(
\sigma_j(\bar{\mathbf U}^{(i)})
\sigma_j(\bar{\mathbf V}^{(i)})
+\epsilon
\right)^{p-1},
\quad 0<p\le1
\end{aligned}
\)
&
& $\checkmark$ & &
\\
\addlinespace[2pt]\midrule
\WSpTFI{} (Ours)
&
\(
\begin{aligned}
\frac1p\|\X\|_{W_{\X},S_p}^{p}
&=
\min_{\X=\mathcal U*\mathcal V}
\left[
\frac1{p_1}
\|\mathcal U\|_{W_{\mathcal U,\mathcal V},S_{p_1}}^{p_1}
+
\frac1{p_2}
\|\mathcal V\|_{W_{\mathcal U,\mathcal V},S_{p_2}}^{p_2}
\right],
\\[-1pt]
&\hspace{13mm}
\frac1p=\frac1{p_1}+\frac1{p_2}
\end{aligned}
\)
&
$\checkmark$ & $\checkmark$ & &
\\
\addlinespace[2pt]

\WSpTFII{} (Ours)
&
\(
\begin{aligned}
\frac1p\|\X\|_{W_{\X},S_p}^{p}
&=
\min_{\X=\mathcal U*\mathcal V}
\frac1\ell
\sum_{i=1}^{n_3}\sum_{j=1}^{r}
\phi\!\left(
\frac{
\|\bar{\mathbf u}_j^{(i)}\|_2^2+
\|\bar{\mathbf v}_j^{(i)}\|_2^2
}{2}
\right)
\end{aligned}
\)
&
$\checkmark$ & $\checkmark$ & $\checkmark$ & $\checkmark$
\\
\bottomrule
\end{tabular}
\end{table*}

The present work is closest to these factorized and nonconvex tensor methods,
but it emphasizes two exact factorization connections to a common weighted
Schatten-$p$ norm-based spectral regularizer. \WSpTFI{} establishes a factor-spectral
relation between the spectrum of the reconstructed tensor and the spectra of
two compact factors while allowing flexible factor exponents. \WSpTFII{}
derives an exact transform-domain paired-column variational representation of
the same spectral functional up to scaling on the common admissible parameter
region; its main factor updates are SVD-free and its component structure
supports finite rank pruning. Accordingly, the proposed models complement
recent factor-norm, correntropy, factorized-gradient, and sparsity-inducing
approaches through different factor-space representations and optimization
mechanisms rather than through different target spectral regularizers.

\section{Notations and Preliminaries}
\label{sec:notations}

This section introduces the notation and the transform-based tensor algebra used throughout the paper. The t-product can be defined through any invertible linear transform applied along the third mode of the tensor, such as the discrete Fourier transform, the discrete cosine transform (DCT), and random orthogonal matrices \cite{KernfeldEtAl2015}. The optimization formulas and the norm identities below rely on a scaled-orthogonality condition on the transform matrix, which preserves Frobenius inner products and norms up to a positive factor. The theoretical development does not depend on any particular transform family.

\subsection{Basic Notation}

Scalars, vectors, matrices, and third-order tensors are denoted by lowercase
letters, bold lowercase letters, bold uppercase letters, and calligraphic
letters, respectively. The real field is $\R$, $\mathbf I$ denotes an identity
matrix of compatible size, and $\langle\cdot,\cdot\rangle$ denotes the
Euclidean/Frobenius inner product. For
$\mathcal A\in\R^{n_1\times n_2\times n_3}$,
$\mathcal A^{(i)}$ and $\mathcal A(i,j,:)$ denote its $i$th frontal slice and
$(i,j)$th tube, respectively; $\bar{\mathcal A}$ denotes its transform-domain
representation; and $\sigma_j(\mathbf A)$ denotes the $j$th singular value of
a matrix $\mathbf A$. The symbols $\X$ and $\Y$ denote the underlying
low-rank/clean tensor and its observation, while $\mathcal U$ and $\mathcal V$
denote factor tensors and $\mathcal N$ or $\mathcal S$ denotes a corruption or
residual tensor when used.

The Frobenius and entrywise $\ell_1$ norms are
\[
\|\mathcal A\|_F^2=\sum_{i,j,k}|\mathcal A_{ijk}|^2,
\qquad
\|\mathcal A\|_1=\sum_{i,j,k}|\mathcal A_{ijk}|.
\]
The Hadamard product is $\odot$. The observed index set is $\OmegaSet$.
For matrix data, $\mathbf M\in\{0,1\}^{n_1\times n_2}$ denotes the binary
observation mask, where $M_{ij}=1$ if $(i,j)\in\OmegaSet$ and $M_{ij}=0$
otherwise. For tensor data,
$\mathcal M\in\{0,1\}^{n_1\times n_2\times n_3}$ denotes the binary
observation mask, where $\mathcal M_{ijk}=1$ if $(i,j,k)\in\OmegaSet$ and
$\mathcal M_{ijk}=0$ otherwise. Accordingly, the projection onto observed
tensor entries is $\mathcal{P}_{\OmegaSet}(\mathcal A)=\mathcal M\odot\mathcal A$.
The symbols $\operatorname{rank}(\mathbf A)$ and
$\operatorname{rank}_t(\mathcal A)$ denote matrix rank and tubal rank,
respectively. For a proper lower-semicontinuous function $f$, $\partial f$
denotes the limiting (Mordukhovich) subdifferential; at a differentiable point,
$\partial f(x)=\{\nabla f(x)\}$.

\subsection{Transform-Based Tensor Algebra}

Let $\mathcal{L}$ denote a linear transform applied along the third mode,
induced by a matrix $\mathbf L\in\R^{n_3\times n_3}$. Throughout the
theoretical development from this section onward, we assume that
\begin{equation}
\mathbf L^*\mathbf L
=
\mathbf L\mathbf L^*
=
\ell\mathbf I,
\qquad \ell>0,
\label{eq:transform-condition}
\end{equation}
where $\mathbf L^*$ denotes the matrix adjoint. The scalar $\ell$ is therefore not
an additional free parameter. It is the common squared scaling induced by
$\mathbf L$. In particular, all singular values of $\mathbf L$ are
$\sqrt{\ell}$, and
\begin{equation}
\ell
=
\frac{1}{n_3}\operatorname{tr}(\mathbf L^*\mathbf L)
=
\frac{\|\mathbf L\|_F^2}{n_3}.
\label{eq:ell-definition}
\end{equation}
Condition~\eqref{eq:transform-condition} implies that $\mathbf L$ is
nonsingular and $\mathbf L^{-1}=\ell^{-1}\mathbf L^*$.
For $\mathcal{A}\in\R^{n_1\times n_2\times n_3}$, write
\[
\bar{\mathcal{A}}=\mathcal{L}(\mathcal{A}),
\qquad
\mathcal{A}=\mathcal{L}^{-1}(\bar{\mathcal{A}}).
\]
It follows directly from \eqref{eq:transform-condition} that
\begin{equation}
  \langle\mathcal A,\mathcal B\rangle
  =
  \frac{1}{\ell}
  \langle\bar{\mathcal A},\bar{\mathcal B}\rangle,
  \qquad
  \|\mathcal A\|_F=\frac{1}{\sqrt{\ell}}\|\bar{\mathcal A}\|_F .
  \label{eq:transform-normalization}
\end{equation}
Thus, the factors $1/\ell$ appearing in the transform-domain TNN,
Schatten-$p$, and weighted Schatten-$p$ norms below are determined by the
normalization of $\mathcal L$ in \eqref{eq:transform-condition}; they are not
independent tuning parameters. The same relation also permits the quadratic
factor subproblems in Sections~\ref{sec:wsptf} and~\ref{sec:vwsptf} to retain
independent transform-domain frontal-slice forms.

For
$\mathcal{A}\in\R^{n_1\times n\times n_3}$ and
$\mathcal{B}\in\R^{n\times n_2\times n_3}$, their t-product
$\mathcal A*\mathcal B$ is defined by \cite{LuPengWei2019TransformTNN}
\begin{equation*}
  \overline{(\mathcal A*\mathcal B)}^{(i)}
  =
  \bar{\mathcal A}^{(i)}\bar{\mathcal B}^{(i)},
  \quad i=1,\ldots,n_3,
\end{equation*}
followed by the inverse transform.

The tensor transpose is defined slice-wise by
\[
\overline{(\mathcal A^\top)}^{(i)}
=
(\bar{\mathcal A}^{(i)})^\top.
\]

\begin{definition}[f-diagonal tensor \cite{LuPengWei2019TransformTNN}]
\label{def:f-diagonal-tensor}
A tensor $\mathcal S\in\R^{n_1\times n_2\times n_3}$ is called
f-diagonal if every transform-domain frontal slice
$\bar{\mathcal S}^{(i)}$ is diagonal.
\end{definition}

\begin{definition}[Identity tensor \cite{LuPengWei2019TransformTNN}]
\label{def:identity-tensor}
The identity tensor $\mathcal I\in\R^{n\times n\times n_3}$ is defined by
$\bar{\mathcal I}^{(i)}=\mathbf I_n$ for every $i$.
\end{definition}

\begin{definition}[Orthogonal tensor \cite{LuPengWei2019TransformTNN}]
\label{def:orthogonal-tensor}
A square tensor $\mathcal Q\in\R^{n\times n\times n_3}$ is called
orthogonal if
$\mathcal Q^\top*\mathcal Q=\mathcal Q*\mathcal Q^\top=\mathcal I$.
\end{definition}

\begin{definition}[t-SVD \cite{LuPengWei2019TransformTNN}]
\label{def:transform-tsvd}
Every
$\mathcal A\in\R^{n_1\times n_2\times n_3}$ admits
\[
\mathcal A=\mathcal U*\mathcal S*\mathcal V^\top,
\]
where
$\mathcal U\in\R^{n_1\times n_1\times n_3}$ and
$\mathcal V\in\R^{n_2\times n_2\times n_3}$ are orthogonal, and
$\mathcal S\in\R^{n_1\times n_2\times n_3}$ is f-diagonal. Equivalently,
\[
\bar{\mathcal A}^{(i)}
=
\bar{\mathcal U}^{(i)}
\bar{\mathcal S}^{(i)}
(\bar{\mathcal V}^{(i)})^\top
\]
for every transform-domain frontal slice.
\end{definition}

\begin{definition}[Tensor singular values \cite{LuPengWei2019TransformTNN}]
\label{def:tensor-singular-values}
Let $\mathcal A\in\R^{n_1\times n_2\times n_3}$ and
$\bar{\mathcal A}=\mathcal L(\mathcal A)$. For each
$i=1,\ldots,n_3$, let
$\sigma_j(\bar{\mathcal A}^{(i)})$ denote the $j$th singular value of
the $i$th transform-domain frontal slice $\bar{\mathcal A}^{(i)}$,
where $j=1,\ldots,\min\{n_1,n_2\}$. The collection
$\{\sigma_j(\bar{\mathcal A}^{(i)})\}_{i,j}$ is referred to as the
tensor singular values of $\mathcal A$ under the adopted transform.
\end{definition}

\subsection{Tubal Rank and Spectral Norms}

The singular values used in this paper are the ordinary matrix singular
values $\sigma_j(\bar{\mathcal A}^{(i)})$ of the transform-domain frontal slices.

\begin{definition}[Tubal rank \cite{LuPengWei2019TransformTNN}]
\label{def:tubal-rank}
The tubal rank of $\mathcal A$ is defined as
\begin{equation}
  \operatorname{rank}_t(\mathcal A)
  =
  \max_{1\le i\le n_3}
  \operatorname{rank}(\bar{\mathcal A}^{(i)}).
  \label{eq:transform-tubal-rank}
\end{equation}
\end{definition}

\begin{definition}[TNN \cite{LuPengWei2019TransformTNN}]
\label{def:tensor-nuclear-norm}
Under the normalization adopted here, the TNN is defined as
\begin{equation*}
  \|\mathcal A\|_*
  =
  \frac{1}{\ell}
  \sum_{i=1}^{n_3}
  \sum_{j=1}^{\min\{n_1,n_2\}}
  \sigma_j(\bar{\mathcal A}^{(i)}).
\end{equation*}
\end{definition}

\begin{definition}[Tensor Schatten-$p$ norm \cite{LiuEtAl2020WeightedTSchatten}]
\label{def:tensor-schatten-p-norm}
For $p>0$, the tensor Schatten-$p$ norm is defined by
\begin{equation}
  \|\mathcal A\|_{S_p}
  =
  \left(
  \frac{1}{\ell}
  \sum_{i=1}^{n_3}
  \sum_{j=1}^{\min\{n_1,n_2\}}
  \sigma_j(\bar{\mathcal A}^{(i)})^p
  \right)^{1/p} .
  \label{eq:tensor-schatten-p-norm}
\end{equation}
For $p\ge1$, this quantity is a norm under the adopted normalization; for
$0<p<1$, it is understood in the usual Schatten-$p$ quasi-norm sense.
\end{definition}

\begin{definition}[Weighted tensor Schatten-$p$ norm \cite{LiuEtAl2020WeightedTSchatten}]
\label{def:weighted-tensor-schatten-p-norm}
Let
$\mathcal A\in\R^{m_1\times m_2\times n_3}$ and
$d_{\mathcal A}=\min\{m_1,m_2\}$. Let $W=[w_{ij}]$ be a nonnegative
weight array, where $w_{ij}$ is the weight assigned to the $j$th singular
value of the $i$th transform-domain frontal slice, with
$i=1,\ldots,n_3$ and $j=1,\ldots,d_{\mathcal A}$. For $p>0$, define the
weighted tensor Schatten-$p$ norm by
\begin{equation}
  \|\mathcal A\|_{W,S_p}
  =
  \left(
  \frac{1}{\ell}
  \sum_{i=1}^{n_3}
  \sum_{j=1}^{d_{\mathcal A}}
  w_{ij}\,
  \sigma_j(\bar{\mathcal A}^{(i)})^p
  \right)^{1/p} .
  \label{eq:weighted-tensor-schatten-norm}
\end{equation}
When $p=1$, we also write
$\|\mathcal A\|_{W,*}:=\|\mathcal A\|_{W,S_p}$ and refer to it as the weighted
tensor nuclear norm. When all $w_{ij}=1$,
\eqref{eq:weighted-tensor-schatten-norm} reduces to
\eqref{eq:tensor-schatten-p-norm}.
\end{definition}

The weight rules used by \WSpTFI{} are introduced in
Section~\ref{sec:wsptf}, where their dependence on the current tensor or
factor pair is explicit. The variational column-pair functional used by
\WSpTFII{} is introduced separately in Section~\ref{sec:vwsptf}.

\section{\WSpTFI{}: Weighted Schatten-$p$ Tensor Factorization I}
\label{sec:wsptf}

This section develops weighted Schatten-$p$ tensor factorization I  (\WSpTFI{}).
The theoretical identity is stated for tensor and factor weights induced by
the current singular values, whereas the numerical scheme is explicitly
interpreted as an iteratively reweighted procedure.

\subsection{\WSpTFI{} Factorization Identity}
\label{subsec:wsptf-factorization}

The identity developed below can be viewed as a weighted extension
of the unweighted two-factor tensor Schatten norm relation in
\eqref{eq:intro-two-factor-tensor-schatten}. Let $p,p_1,p_2>0$ satisfy
\begin{equation}
  \frac{1}{p}=\frac{1}{p_1}+\frac{1}{p_2}.
  \label{eq:p-relation}
\end{equation}

\begin{lemma}[Product singular-value inequality, \cite{HornJohnson1991}]
\label{lem:wsptf-product-singular-value}
Let $\mathbf A\in\R^{m\times r}$ and
$\mathbf B\in\R^{r\times n}$, and let
$d\leq\min\{m,n,r\}$. Suppose
$h:\R_+\rightarrow\R$ is continuous at zero and
$\varphi(x)=h(e^x)$ is convex and nondecreasing on $\R$. Then
\[
  \sum_{j=1}^{d}h(\sigma_j(\mathbf A\mathbf B))
  \leq
  \sum_{j=1}^{d}
  h(\sigma_j(\mathbf A)\sigma_j(\mathbf B)).
\]
\end{lemma}

\begin{theorem}[Factorized weighted tensor Schatten-$p$ norm]
\label{thm:tensor-factorization}
Let $\X\in\R^{n_1\times n_2\times n_3}$ satisfy
$\operatorname{rank}_t(\X)\le r$. Assume $0<p\le1$, $0<q\le1$, $c>0$, $\varepsilon>0$, and
$q\ge1-\frac{4p^2}{1+4p}$, and let $p_1,p_2>0$ satisfy
\eqref{eq:p-relation}. Define the tensor weights
\[
w_{ij}^{\X}
=
c\left(\sigma_j(\bar{\X}^{(i)})+\varepsilon\right)^{q-1},
\]
and, for every feasible factorization $\X=\mathcal U*\mathcal V$, define the
shared factor weights
\begin{equation*}
w_{ij}^{\mathcal U,\mathcal V}
=
c\left(
\sigma_j(\bar{\mathcal U}^{(i)})
\sigma_j(\bar{\mathcal V}^{(i)})
+\varepsilon
\right)^{q-1}.
\end{equation*}
The factor singular-value sequences are zero-padded when necessary. Then
\begin{equation*}
\begin{aligned}
  \frac{1}{p}\|\X\|_{W_{\X},S_p}^{p}
  =
  \min_{\substack{
    \X=\mathcal U*\mathcal V\\
    \mathcal U\in\R^{n_1\times r\times n_3}\\
    \mathcal V\in\R^{r\times n_2\times n_3}
  }}
  \bigg\{
  &\frac{1}{p_1}
  \|\mathcal U\|_{W_{\mathcal U,\mathcal V},S_{p_1}}^{p_1}\\
  &+
  \frac{1}{p_2}
  \|\mathcal V\|_{W_{\mathcal U,\mathcal V},S_{p_2}}^{p_2}
  \bigg\}.
\end{aligned}
\end{equation*}
\end{theorem}

The proof is provided in Appendix~\ref{app:proof-thm-tensor-factorization}.
The lower bound on $q$ is a sufficient condition ensuring that
$h(e^x)$, with
$h(t)=c(t+\varepsilon)^{q-1}t^p$, is nondecreasing and convex, as required by
Lemma~\ref{lem:wsptf-product-singular-value}.

\begin{remark}[Explanation and matrix reduction]
Theorem~\ref{thm:tensor-factorization} gives an exact factorized
representation of the weighted tensor Schatten-$p$ norm regularizer in terms
of two compact tensor factors with flexible exponents.
This identity preserves the weighted spectral meaning of the completed tensor
while moving the optimization variables to lower-dimensional factors, thereby
providing the mathematical foundation of \WSpTFI{}.
When $n_3=1$ and the transform is the identity, the t-product reduces to
ordinary matrix multiplication and the result becomes
\[
\frac{1}{p}\|\mathbf X\|_{W_{\mathbf X},S_p}^{p}
=
\min_{\mathbf X=\mathbf U\mathbf V}
\left\{
\frac{1}{p_1}\|\mathbf U\|_{W_{\mathbf U,\mathbf V},S_{p_1}}^{p_1}
+
\frac{1}{p_2}\|\mathbf V\|_{W_{\mathbf U,\mathbf V},S_{p_2}}^{p_2}
\right\},
\]
which recovers the corresponding factorized weighted Schatten-$p$ matrix
construction as a special case.
\end{remark}

\begin{remark}[Interpretation of the parameter condition]
For $p=1$, the sufficient lower bound is $q\ge0.2$, while for $p=1/2$ it is
$q\ge2/3$. The condition is used in the factorization proof. Moreover, the
tensor weights and factor weights are not identical for an arbitrary feasible
factor pair. Equality is attained by the balanced transform-slice factor
construction used in Appendix~\ref{app:proof-thm-tensor-factorization}, which
distributes each reconstructed-slice singular value between the two factors
according to the prescribed factor exponents. 
\end{remark}

\subsection{Model Formulation}
\label{subsec:wsptf-model}

Consider an incomplete and corrupted observation $\Y$ of a low-rank tensor
$\X$ under the observation mask $\mathcal M$. Let $\lambda_1>0$ and
$\lambda_2>0$ be regularization parameters controlling, respectively, the
strength of the low-rank prior and the sparse data-fidelity term. The ideal
robust formulation combines tubal rank and sparse residual counting,
\begin{equation*}
  \min_{\X}\;
  \lambda_1\operatorname{rank}_{t}(\X)
  +
  \lambda_2\|\mathcal M\odot(\X-\Y)\|_0 .
\end{equation*}
A standard convex surrogate replaces these two terms by the tensor nuclear
norm (TNN) and the entrywise $\ell_1$ loss,
\begin{equation*}
  \min_{\X}\;
  \lambda_1\|\X\|_*
  +
  \lambda_2\|\mathcal M\odot(\X-\Y)\|_1 .
\end{equation*}

To reduce the excessive attenuation that can arise when all singular
components share the same shrinkage profile, we instead use the weighted
tensor Schatten-$p$ norm through the norm-based regularizer
\begin{equation}
  \min_{\X}\;
  \lambda_1\|\X\|_{W_{\X},S_p}^{p}
  +
  \lambda_2\|\mathcal M\odot(\X-\Y)\|_1 ,
  \label{eq:wsptf-original}
\end{equation}
where $W_{\X}$ is generated from the singular spectrum of $\X$. In the
numerical procedure, the weights are updated from the current factors and
then frozen within the corresponding spectral subproblem; hence the solver is
an \emph{iteratively reweighted ADMM-type scheme}.

Applying Theorem~\ref{thm:tensor-factorization} to the weighted tensor
Schatten-$p$ norm regularizer gives our first factorized model, \WSpTFI{}. After absorbing the scalar factor
$p$ in Theorem~\ref{thm:tensor-factorization} into $\lambda_1$, \WSpTFI{} is
written as
\begin{equation}
\begin{aligned}
\min_{\X,\mathcal U,\mathcal V}\;&
\lambda_1\left(
\frac{1}{p_1}
\|\mathcal U\|_{W_{\mathcal U,\mathcal V},S_{p_1}}^{p_1}
+
\frac{1}{p_2}
\|\mathcal V\|_{W_{\mathcal U,\mathcal V},S_{p_2}}^{p_2}
\right)\\
&+\lambda_2\|\mathcal M\odot(\X-\Y)\|_1\\
\mathrm{s.t.}\;&\X=\mathcal U*\mathcal V .
\end{aligned}
\label{eq:wsptf-factorized-model}
\end{equation}
This explicit factorized form connects the weighted tensor Schatten-$p$ norm
in \eqref{eq:wsptf-original} directly to the compact factors used by the
algorithm.

\subsection{Iteratively Reweighted ADMM}
\label{subsec:wsptf-admm}

At a particular reweighted update, $W$ is treated as fixed. Introducing
auxiliary variables gives the
fixed-weight split surrogate
\begin{equation}
\begin{aligned}
\min_{\X,\mathcal U,\mathcal V,\widehat{\mathcal U},\widehat{\mathcal V}}
&\;
\lambda_1\left(
\frac{1}{p_1}\|\widehat{\mathcal U}\|_{W,S_{p_1}}^{p_1}
+
\frac{1}{p_2}\|\widehat{\mathcal V}\|_{W,S_{p_2}}^{p_2}
\right)\\
&+\lambda_2\|\mathcal M\odot(\X-\Y)\|_1\\
\mathrm{s.t.}\;&
\widehat{\mathcal U}=\mathcal U,\quad
\widehat{\mathcal V}=\mathcal V,\quad
\mathcal U*\mathcal V=\X .
\end{aligned}
\label{eq:wsptf-split}
\end{equation}
Here $\widehat{\mathcal U}$ and $\widehat{\mathcal V}$ are two auxiliary
variables introduced as copies of $\mathcal U$ and $\mathcal V$, respectively,
so that the weighted Schatten norm regularizers are decoupled from the bilinear
reconstruction constraint $\mathcal U*\mathcal V=\X$. The matrix $W$ is a
placeholder for the weight matrix frozen during the current update. Across iterations it is replaced by the current iteratively reweighted
estimate.

For fixed $W$, let $\mu>0$ denote the ADMM penalty parameter, and let
$\mathcal Z_1$, $\mathcal Z_2$, and $\mathcal Z_3$ denote the Lagrange
multiplier tensors associated with
$\widehat{\mathcal U}=\mathcal U$,
$\widehat{\mathcal V}=\mathcal V$, and
$\mathcal U*\mathcal V=\X$, respectively. The augmented Lagrangian
function of \eqref{eq:wsptf-split} is
\begin{equation*}
\begin{aligned}
\mathcal L_\mu
={}&
\frac{\lambda_1}{p_1}
\|\widehat{\mathcal U}\|_{W,S_{p_1}}^{p_1}
+
\frac{\lambda_1}{p_2}
\|\widehat{\mathcal V}\|_{W,S_{p_2}}^{p_2}\\
&+
\lambda_2\|\mathcal M\odot(\X-\Y)\|_1\\
&+
\frac{\mu}{2}
\|\widehat{\mathcal U}-\mathcal U
+\mu^{-1}\mathcal Z_1\|_F^2\\
&+
\frac{\mu}{2}
\|\widehat{\mathcal V}-\mathcal V
+\mu^{-1}\mathcal Z_2\|_F^2\\
&+
\frac{\mu}{2}
\|\mathcal U*\mathcal V-\X
+\mu^{-1}\mathcal Z_3\|_F^2 .
\end{aligned}
\end{equation*}

At iteration $k$, let
$\mathcal A^k=\X^k-\mu_k^{-1}\mathcal Z_3^k$. By
\eqref{eq:transform-normalization}, the $\mathcal U$ block decomposes into independent
transform-slice least-squares problems:
\begin{equation}
\begin{aligned}
\bar{\mathcal U}^{(i),k+1}
=&
\left(
\bar{\widehat{\mathcal U}}^{(i),k}
+\mu_k^{-1}\bar{\mathcal Z}_1^{(i),k}
+\bar{\mathcal A}^{(i),k}
(\bar{\mathcal V}^{(i),k})^\top
\right)\\
&\times
\left(
\mathbf I+
\bar{\mathcal V}^{(i),k}
(\bar{\mathcal V}^{(i),k})^\top
\right)^{-1}.
\end{aligned}
\label{eq:wsptf-u-update}
\end{equation}
The $\mathcal V$ update is
\begin{equation}
\begin{aligned}
\bar{\mathcal V}^{(i),k+1}
=&
\left(
\mathbf I+
(\bar{\mathcal U}^{(i),k+1})^\top
\bar{\mathcal U}^{(i),k+1}
\right)^{-1}\\
&\times
\left(
\bar{\widehat{\mathcal V}}^{(i),k}
+\mu_k^{-1}\bar{\mathcal Z}_2^{(i),k}
+(\bar{\mathcal U}^{(i),k+1})^\top
\bar{\mathcal A}^{(i),k}
\right).
\end{aligned}
\label{eq:wsptf-v-update}
\end{equation}

Before updating $\widehat{\mathcal U}$, the current shared weights are
evaluated from the previous spectral factors:
\begin{equation}
w_{ij}^{k}
=
c\left(
\sigma_j(\bar{\widehat{\mathcal U}}^{(i),k})
\sigma_j(\bar{\widehat{\mathcal V}}^{(i),k})
+\varepsilon
\right)^{q-1}.
\label{eq:wsptf-admm-weight}
\end{equation}
They are held fixed while the $\widehat{\mathcal U}$ proximal subproblem is
solved. Let
$\mathcal G_1^k=\mathcal U^{k+1}-\mu_k^{-1}\mathcal Z_1^k$ and compute
\[
\bar{\mathcal G}_1^{(i),k}
=
\mathbf P_i\operatorname{diag}(a_{i1},\ldots,a_{ir_i})\mathbf Q_i^\top .
\]
Here $r_i=\min\{n_1,r\}$, $a_{ij}\ge0$ are the singular values of
$\bar{\mathcal G}_1^{(i),k}$, and
$\mathbf P_i\in\R^{n_1\times r_i}$ and
$\mathbf Q_i\in\R^{r\times r_i}$ collect the corresponding left and right
singular vectors. Then
\[
\bar{\widehat{\mathcal U}}^{(i),k+1}
=
\mathbf P_i
\operatorname{diag}(u_{i1}^{k+1},\ldots,u_{ir_i}^{k+1})
\mathbf Q_i^\top,
\]
where
\begin{equation}
u_{ij}^{k+1}
=
\argmin_{s\ge0}
\left\{
\frac{\lambda_1w_{ij}^k}{p_1}s^{p_1}
+\frac{\mu_k}{2}(s-a_{ij})^2
\right\}.
\label{eq:wsptf-u-singular-subproblem}
\end{equation}
Its positive stationary points satisfy
\begin{equation}
\mu_k(s-a_{ij})+\lambda_1w_{ij}^k s^{p_1-1}=0.
\label{eq:wsptf-u-singular-optimality}
\end{equation}
For $p_1=2$, the correct quadratic shrinkage is
\begin{equation*}
u_{ij}^{k+1}
=
\frac{a_{ij}}{1+\lambda_1w_{ij}^k/\mu_k}.
\end{equation*}
For $p_1=1$, it becomes weighted soft thresholding,
\[
u_{ij}^{k+1}
=
\max\left\{
a_{ij}-\frac{\lambda_1w_{ij}^k}{\mu_k},0
\right\}.
\]
For $p_1>1$, the scalar objective is convex and
\eqref{eq:wsptf-u-singular-optimality} has a unique nonnegative minimizer.
For $0<p_1<1$, the scalar problem is nonconvex; we use a safeguarded one-dimensional solver
to locate all admissible positive stationary candidates and select the
candidate with the smallest objective value after also comparing with
$s=0$. The same rule is used for $p_2$.

After $\widehat{\mathcal U}^{k+1}$ is obtained, a Gauss--Seidel reweighting
step evaluates
\begin{equation}
w_{ij}^{k+1/2}
=
c\left(
\sigma_j(\bar{\widehat{\mathcal U}}^{(i),k+1})
\sigma_j(\bar{\widehat{\mathcal V}}^{(i),k})
+\varepsilon
\right)^{q-1}.
\label{eq:wsptf-half-weight}
\end{equation}
These weights are frozen in the analogous $\widehat{\mathcal V}$ subproblem
\begin{equation}
v_{ij}^{k+1}
=
\argmin_{s\ge0}
\left\{
\frac{\lambda_1w_{ij}^{k+1/2}}{p_2}s^{p_2}
+\frac{\mu_k}{2}(s-b_{ij})^2
\right\},
\label{eq:wsptf-v-singular-subproblem}
\end{equation}
where $b_{ij}$ are the singular values of
$\bar{\mathcal G}_2^{(i),k}$ with
$\mathcal G_2^k=\mathcal V^{k+1}-\mu_k^{-1}\mathcal Z_2^k$.

For the robust tensor variable, define
\[
\mathcal B^k
=
\mathcal U^{k+1}*\mathcal V^{k+1}
+\mu_k^{-1}\mathcal Z_3^k .
\]
The elementwise update is
\begin{equation}
\X^{k+1}
=
\mathcal M\odot
\left[
\Y+\mathcal S_{\lambda_2/\mu_k}(\mathcal B^k-\Y)
\right]
+
(\mathbf1-\mathcal M)\odot\mathcal B^k,
\label{eq:wsptf-x-update}
\end{equation}
where $\mathbf 1$ denotes the all-ones tensor with the same size as
$\mathcal M$, and
$\mathcal S_\tau(z)=\operatorname{sign}(z)\max(|z|-\tau,0)$.

The multipliers are updated by
\begin{equation}
\mathcal Z_1^{k+1}
=
\mathcal Z_1^k+
\mu_k(\widehat{\mathcal U}^{k+1}-\mathcal U^{k+1}),
\label{eq:wsptf-z1-update-opt}
\end{equation}
\[
\mathcal Z_2^{k+1}
=
\mathcal Z_2^k+
\mu_k(\widehat{\mathcal V}^{k+1}-\mathcal V^{k+1}),
\]
and
\begin{equation}
\mathcal Z_3^{k+1}
=
\mathcal Z_3^k+
\mu_k(\mathcal U^{k+1}*\mathcal V^{k+1}-\X^{k+1}),
\label{eq:wsptf-z3-update-opt}
\end{equation}
with $\mu_{k+1}=\rho\mu_k$, $\rho>1$.

For reproducibility, define the relative reconstruction change
\[
r_{\mathrm{rel}}^{k+1}
=
\frac{\|\X^{k+1}-\X^k\|_F}{\max\{1,\|\X^k\|_F\}}.
\]
Define the three normalized feasibility residuals
\begin{equation*}
\begin{aligned}
r_U^{k+1}
&=
\frac{\|\widehat{\mathcal U}^{k+1}
-\mathcal U^{k+1}\|_F}
{\max\{1,\|\mathcal U^{k+1}\|_F\}},\\
r_V^{k+1}
&=
\frac{\|\widehat{\mathcal V}^{k+1}
-\mathcal V^{k+1}\|_F}
{\max\{1,\|\mathcal V^{k+1}\|_F\}},\\
r_X^{k+1}
&=
\frac{\|\mathcal U^{k+1}*\mathcal V^{k+1}
-\X^{k+1}\|_F}
{\max\{1,\|\X^{k+1}\|_F\}} .
\end{aligned}
\end{equation*}
The normalized primal residual is then
\begin{equation*}
r_{\mathrm{pri}}^{k+1}
=
\max\{r_U^{k+1},r_V^{k+1},r_X^{k+1}\}.
\end{equation*}
Given a stopping tolerance $\mathrm{tol}>0$, the iteration stops when
$\max\{r_{\mathrm{rel}}^{k+1},r_{\mathrm{pri}}^{k+1}\}<\mathrm{tol}$.

\begin{algorithm}[!t]
\caption{Iteratively Reweighted \WSpTFI{}-ADMM}
\label{alg:wsptf-admm}
\begin{algorithmic}[1]
\STATE \textbf{Input:} $\Y$, $\mathcal M$,
$\lambda_1,\lambda_2,p_1,p_2,q,c,\varepsilon,\rho,\mu_0$,
initial factor width, maximum iteration number $K$, and stopping tolerance
$\mathrm{tol}$.
\STATE Initialize
$\X^0,\mathcal U^0,\mathcal V^0,
\widehat{\mathcal U}^0,\widehat{\mathcal V}^0,
\mathcal Z_1^0,\mathcal Z_2^0,\mathcal Z_3^0$.
\FOR{$k=0,1,\ldots,K-1$}
  \STATE Update $\mathcal U^{k+1}$ by \eqref{eq:wsptf-u-update}.
  \STATE Update $\mathcal V^{k+1}$ by \eqref{eq:wsptf-v-update}.
  \STATE Evaluate $W^k$ by \eqref{eq:wsptf-admm-weight} and hold it fixed.
  \STATE Update $\widehat{\mathcal U}^{k+1}$ by
  \eqref{eq:wsptf-u-singular-subproblem}, using the exact scalar minimizer.
  \STATE Evaluate $W^{k+1/2}$ by \eqref{eq:wsptf-half-weight} and hold it fixed.
  \STATE Update $\widehat{\mathcal V}^{k+1}$ by
  \eqref{eq:wsptf-v-singular-subproblem}, using the exact scalar minimizer.
  \STATE Update $\X^{k+1}$ by \eqref{eq:wsptf-x-update}.
  \STATE Update the multipliers by
  \eqref{eq:wsptf-z1-update-opt}--\eqref{eq:wsptf-z3-update-opt}.
  \STATE Set $\mu_{k+1}=\rho\mu_k$.
  \STATE Stop if
  $\max\{r_{\mathrm{rel}}^{k+1},r_{\mathrm{pri}}^{k+1}\}<\mathrm{tol}$.
\ENDFOR
\STATE \textbf{Output:} recovered tensor $\X$.
\end{algorithmic}
\end{algorithm}

The complete iteratively reweighted ADMM procedure is summarized in
Algorithm~\ref{alg:wsptf-admm}.

\subsection{Conditional Limiting Analysis}
\label{subsec:wsptf-theoretical-analysis}

Because \WSpTFI{} combines bilinear factorization, nonconvex spectral norm regularizers,
and iteration-dependent weights, the following result is stated
conditionally. 

If the two reweighting sequences converge to $W^\star$, define the associated
limiting fixed-weight split problem by
\begin{equation}
\begin{aligned}
\min_{\X,\mathcal U,\mathcal V,\widehat{\mathcal U},\widehat{\mathcal V}}
&\;
\lambda_1\left(
\frac{1}{p_1}\|\widehat{\mathcal U}\|_{W^\star,S_{p_1}}^{p_1}
+
\frac{1}{p_2}\|\widehat{\mathcal V}\|_{W^\star,S_{p_2}}^{p_2}
\right)\\
&+\lambda_2\|\mathcal M\odot(\X-\Y)\|_1\\
\mathrm{s.t.}\;&
\widehat{\mathcal U}=\mathcal U,\quad
\widehat{\mathcal V}=\mathcal V,\quad
\mathcal U*\mathcal V=\X .
\end{aligned}
\label{eq:wsptf-limiting-fixed-weight}
\end{equation}

\begin{assumption}
\label{ass:wsptf-convergence}
For \WSpTFI{}, assume that:
\begin{enumerate}
  \item $\lambda_1,\lambda_2,c,\varepsilon>0$,
  $0<p\le1$, $0<q\le1$,
  $q\ge1-\frac{4p^2}{1+4p}$, and $p_1,p_2>0$ satisfy
  \eqref{eq:p-relation};
  \item $\mu_{k+1}=\rho\mu_k$ with $\rho>1$;
  \item $\{\mathcal Z_1^k\}$ and $\{\mathcal Z_2^k\}$ are bounded;
  \item every block update, including each scalar spectral subproblem, is
  solved to its global minimum;
  \item the generated primal sequence is bounded;
  \item the factor-block KKT residuals satisfy
  \begin{equation*}
  \begin{aligned}
  \mathcal R_U^k
  &:=
  -\mathcal Z_1^k+
  \mathcal Z_3^k*(\mathcal V^k)^\top
  \longrightarrow0,\\
  \mathcal R_V^k
  &:=
  -\mathcal Z_2^k+
  (\mathcal U^k)^\top*\mathcal Z_3^k
  \longrightarrow0 .
  \end{aligned}
  \end{equation*}
\end{enumerate}
\end{assumption}

\begin{theorem}[Conditional Limiting KKT Characterization of WS$_p$TF-I-ADMM]
\label{thm:wsptf-convergence}
Let
\[
\mathcal S^k=
(\mathcal U^k,\mathcal V^k,
\widehat{\mathcal U}^k,\widehat{\mathcal V}^k,
\X^k,\mathcal Z_1^k,\mathcal Z_2^k,\mathcal Z_3^k)
\]
be generated by Algorithm~\ref{alg:wsptf-admm}. Under
Assumption~\ref{ass:wsptf-convergence}, $\{\mathcal Z_3^k\}$ is bounded and the five primal sequences
$\{\mathcal U^k\}$, $\{\mathcal V^k\}$,
$\{\widehat{\mathcal U}^k\}$,
$\{\widehat{\mathcal V}^k\}$, and $\{\X^k\}$ are Cauchy. The two weight
sequences converge to the same limit $W^\star$, and
\begin{equation*}
\begin{aligned}
\widehat{\mathcal U}^{k+1}
-\mathcal U^{k+1}
&\longrightarrow0,\\
\widehat{\mathcal V}^{k+1}
-\mathcal V^{k+1}
&\longrightarrow0,\\
\mathcal U^{k+1}*\mathcal V^{k+1}
-\X^{k+1}
&\longrightarrow0 .
\end{aligned}
\end{equation*}
Every accumulation point of the full primal-dual sequence satisfies the KKT
conditions of the limiting fixed-weight problem
\eqref{eq:wsptf-limiting-fixed-weight}.
\end{theorem}

The proof is given in Appendix~\ref{app:convergence}. Under
Assumption~\ref{ass:wsptf-convergence}, boundedness, global block optimality,
and the geometrically increasing penalty are used to establish bounded
increments and the Cauchy property, while the vanishing factor-block
residuals supply the two factor-stationarity equations that are not implied
by primal feasibility alone. 

\subsection{Computational Complexity}
\label{subsec:wsptf-complexity}

Let the factor width be
$r\ll\min\{n_1,n_2\}$. The two factor least-squares updates require
\[
\mathcal O\!\left(
n_1n_2n_3r+(n_1+n_2)n_3r^2+n_3r^3
\right)
\]
operations. Compact SVDs in the two spectral proximal updates cost
$\mathcal O((n_1+n_2)n_3r^2)$. The scalar proximal cost is
$\mathcal O(n_3r)$ for closed-form cases and is multiplied by the bounded
number of safeguarded one-dimensional iterations otherwise. Let
$C_{\mathcal L}(n_3)$ denote the arithmetic cost of applying
$\mathcal L$ or $\mathcal L^{-1}$ to one length-$n_3$ tube. The transform
and inverse-transform operations contribute
\[
\mathcal O\!\left(
[n_1n_2+(n_1+n_2)r]\,C_{\mathcal L}(n_3)
\right).
\]
Thus the per-iteration complexity is
\[
\begin{aligned}
\mathcal O\!\big(&
n_1n_2n_3r+(n_1+n_2)n_3r^2+n_3r^3\\
&+[n_1n_2+(n_1+n_2)r]C_{\mathcal L}(n_3)
\big),
\end{aligned}
\]
apart from a bounded scalar root-finding factor. For a dense transform matrix,
$C_{\mathcal L}(n_3)=\mathcal O(n_3^2)$, whereas structured transforms may
admit fast implementations with $C_{\mathcal L}(n_3)=\mathcal O(n_3\log n_3)$.
The storage complexity is
$\mathcal O(n_1n_2n_3+(n_1+n_2)rn_3)$. These expressions are
per-iteration bounds and do not by themselves imply a wall-clock advantage
over \WSpTFII{}; total runtime also depends on the iteration count, scalar
solver, and active rank.

\section{\WSpTFII{}: Weighted Schatten-$p$ Tensor Factorization II}
\label{sec:vwsptf}

\WSpTFII{} complements \WSpTFI{} by replacing factor spectral proximal mappings with
a variational regularizer constructed from transform-domain column-pair energies.
The resulting main factor updates are SVD-free, and redundant components can
be removed through finite column pruning.

\subsection{\WSpTFII{} Variational Form}
\label{subsec:vwsptf-variational-form}

For
$\X=\mathcal U*\mathcal V$, write
\[
\bar{\X}^{(i)}
=
\bar{\mathcal U}^{(i)}\bar{\mathcal V}^{(i)}
=
\sum_{j=1}^{r}
\bar{\mathbf u}_j^{(i)}
(\bar{\mathbf v}_j^{(i)})^\top ,
\]
where $\bar{\mathbf u}_j^{(i)}$ is the $j$th column of
$\bar{\mathcal U}^{(i)}$ and $\bar{\mathbf v}_j^{(i)}$ is the transpose of
the $j$th row of $\bar{\mathcal V}^{(i)}$. Define the balanced pair energy
\[
t_j^{(i)}
=
\frac{
\|\bar{\mathbf u}_j^{(i)}\|_2^2+
\|\bar{\mathbf v}_j^{(i)}\|_2^2
}{2}.
\]
The scalar penalty is
\[
\phi(t)=t^p(t+\epsilon)^{q-1},
\qquad
0<p\le1,\quad0<q\le1,\quad\epsilon>0.
\]
Its derivative is
\begin{equation}
\phi'(t)
=
t^{p-1}(t+\epsilon)^{q-2}
\left[(p+q-1)t+p\epsilon\right].
\label{eq:vwsptf-phi-derivative}
\end{equation}
When $p+q\ge1$, $\phi$ is nondecreasing. Moreover,
\[
\begin{aligned}
\phi''(t)
={}&
t^{p-2}(t+\epsilon)^{q-3}
\Big[
p(p-1)\epsilon^2\\
&\quad
+2p(p+q-2)\epsilon t\\
&\quad
+(p+q-1)(p+q-2)t^2
\Big]
\le0 .
\end{aligned}
\]
so $\phi$ is concave on $\R_+$. The condition $p+q\ge1$ is therefore not
merely a tuning restriction: it keeps the spectral regularizer nondecreasing
while enabling the tangent majorization used below.

Define
\begin{equation*}
R_{p,q,\epsilon}(\mathcal U,\mathcal V)
=
\frac{1}{\ell}
\sum_{i=1}^{n_3}
\sum_{j=1}^{r}
\phi(t_j^{(i)}).
\end{equation*}

\begin{lemma}[Singular-value inequality for concave functions, \cite{Thompson1976}]
\label{lem:ips-concave-singular-value}
Let $\mathbf A,\mathbf B,\mathbf C\in\R^{m\times n}$ with
$m\ge n$ and $\mathbf C=\mathbf A+\mathbf B$. Let
$a_i,b_i,c_i$ be their singular values. For any permutation-invariant,
coordinatewise nondecreasing, concave
$F:\R_+^{2n}\rightarrow\R$,
\[
F(c_1,\ldots,c_n,0,\ldots,0)
\le
F(a_1,\ldots,a_n,b_1,\ldots,b_n).
\]
\end{lemma}

\begin{theorem}[Column variational characterization\hfill\break of \WSpTFII{}]
\label{thm:vwsptf-variational}
Let $\X\in\R^{n_1\times n_2\times n_3}$ and
$r\ge\operatorname{rank}_t(\X)$. Assume $0<p\le1$, $0<q\le1$,
$p+q\ge1$, and $\epsilon>0$.
With $s=\min\{n_1,n_2\}$, define
\[
\rho_{p,q,\epsilon}(\X)
=
\frac{1}{\ell}
\sum_{i=1}^{n_3}
\sum_{j=1}^{s}
\phi(\sigma_j(\bar{\X}^{(i)})).
\]
Then
\begin{equation}
\rho_{p,q,\epsilon}(\X)
=
\min_{\substack{
\X=\mathcal U*\mathcal V\\
\mathcal U\in\R^{n_1\times r\times n_3}\\
\mathcal V\in\R^{r\times n_2\times n_3}
}}
R_{p,q,\epsilon}(\mathcal U,\mathcal V).
\label{eq:vwsptf-variational-equivalence}
\end{equation}
\end{theorem}

\begin{remark}[Explanation and matrix reduction of the column form]
Theorem~\ref{thm:vwsptf-variational} converts the transform-domain spectral
regularizer into an exact sum of paired-column energies. This column-wise
representation is useful because the main factor updates can be formed
without factor SVDs, while the same paired components provide a direct energy
criterion for removing redundant factor columns. When $n_3=1$ and the
transform is the identity, \eqref{eq:vwsptf-variational-equivalence} reduces
to the matrix relation
\[
\rho_{p,q,\epsilon}(\mathbf X)
=
\min_{\mathbf X=\mathbf U\mathbf V}
\sum_{j=1}^{r}
\phi\!\left(
\frac{\|\mathbf u_j\|_2^2+\|\mathbf v_j\|_2^2}{2}
\right),
\]
which is the paired-column matrix counterpart of the proposed tensor
variational construction.
\end{remark}

\begin{remark}[Common weighted Schatten-$p$ regularizer underlying \WSpTFI{} and \WSpTFII{}]
\label{rem:common-spectral-regularizer}
Assume the parameter conditions of
Theorem~\ref{thm:tensor-factorization} and set $\epsilon=\varepsilon$.
Then
\[
q\ge 1-\frac{4p^2}{1+4p}
\quad\Longrightarrow\quad
p+q\ge 1+\frac{p}{1+4p}>1,
\]
so the parameter condition of
Theorem~\ref{thm:vwsptf-variational} is also satisfied. With
$w_{ij}^{\X}=c\left(\sigma_j(\bar{\X}^{(i)})+\epsilon\right)^{q-1}$,
the two spectral functionals satisfy
\begin{equation*}
\|\X\|_{W_{\X},S_p}^{p}
=
c\,\rho_{p,q,\epsilon}(\X).
\end{equation*}
Consequently, Theorems~\ref{thm:tensor-factorization} and
\ref{thm:vwsptf-variational} give two exact factor-space representations of
the same weighted tensor Schatten-$p$ norm-based spectral regularizer, up to
the positive scaling factor $c/p$, which can be absorbed into the
regularization parameter.
Thus, the distinction between \WSpTFI{} and \WSpTFII{} lies in their
factor parameterizations and resulting optimization mechanisms rather than in
the target spectral regularizer.
\end{remark}

The proof is given in Appendix~\ref{app:vwsptf-proof}. When $q=1$,
$\phi(t)=t^p$ and Theorem~\ref{thm:vwsptf-variational} reduces to the
transform-domain variational Schatten-$p$ construction. When $p=q=1$, it further
reduces to the standard quadratic variational nuclear representation. The
condition $r\ge\operatorname{rank}_t(\X)$ is an existence condition for the
variational equality; in computation, $r$ may be deliberately overestimated
and subsequently reduced by pruning.

\subsection{Model Formulation}
\label{subsec:vwsptf-model}

Using the spectral functional from
Theorem~\ref{thm:vwsptf-variational}, consider
\begin{equation*}
\min_{\X}\;
\lambda_1\rho_{p,q,\epsilon}(\X)
+
\lambda_2\|\mathcal M\odot(\X-\Y)\|_1 .
\end{equation*}
By substituting the exact column variational characterization in
Theorem~\ref{thm:vwsptf-variational} and parameterizing
$\X=\mathcal U*\mathcal V$, we obtain our second formulation, \WSpTFII{}:
\begin{equation*}
\min_{\mathcal U,\mathcal V}\;
\lambda_1R_{p,q,\epsilon}(\mathcal U,\mathcal V)
+
\lambda_2
\|\mathcal M\odot(\mathcal U*\mathcal V-\Y)\|_1 .
\end{equation*}
Thus, this paired-column factor objective is precisely the \WSpTFII{} model
used in the subsequent IRLS--BSUM development.

For the optimization and theory, divide by $\lambda_2>0$ and define once
and for all $\lambda:=\frac{\lambda_1}{\lambda_2}$.
The original $\ell_1$ model motivates the robust formulation, whereas the
implemented IRLS-BSUM scheme optimizes the following fixed-$\delta$
smoothing with $\delta>0$:
\begin{equation}
\begin{aligned}
F_\delta(\mathcal U,\mathcal V)
=&
\sum_{(a,b,c)\in\OmegaSet}
\sqrt{
[(\mathcal U*\mathcal V)_{abc}-\Y_{abc}]^2+\delta^2
}\\
&+
\lambda R_{p,q,\epsilon}(\mathcal U,\mathcal V).
\end{aligned}
\label{eq:vwsptf-smoothed-objective}
\end{equation}
As $\delta\to0$, its data term approaches the original entrywise
$\ell_1$ fidelity, but the convergence results below are stated for fixed
$\delta>0$.

\subsection{IRLS-BSUM Optimization}
\label{subsec:vwsptf-irls-bsum}

At iteration $k$, set
\begin{equation}
\X^k=\mathcal U^k*\mathcal V^k,\qquad
\mathcal R^k=\mathcal M\odot(\X^k-\Y).
\label{eq:vwsptf-opt-residual}
\end{equation}
The IRLS coefficients are
\begin{equation}
d_{abc}^k
=
\frac{1}{\sqrt{(\mathcal R_{abc}^k)^2+\delta^2}},
\qquad (a,b,c)\in\OmegaSet,
\label{eq:vwsptf-opt-irls}
\end{equation}
and are set to zero outside $\OmegaSet$. With the resulting tensor $\mathcal D^k$ and an additive constant $C_k$
that is independent of $\mathcal U$ and $\mathcal V$, the smoothed data
term is majorized by
\[
\frac12
\left\|
\sqrt{\mathcal D^k}\odot\mathcal M\odot
(\mathcal U*\mathcal V-\Y)
\right\|_F^2+C_k .
\]

For the regularizer, concavity gives
$\phi(t)\le\phi(t^k)+\phi'(t^k)(t-t^k)$. When $p<1$, the derivative is
singular at zero, so the numerical implementation introduces a small
safeguard parameter $\epsilon_t>0$ and uses
$t_{j,\mathrm{safe}}^{(i),k}=\max\{t_j^{(i),k},\epsilon_t\}$,
and evaluates
\begin{equation}
\begin{aligned}
\omega_j^{(i),k}
={}&
(t_{j,\mathrm{safe}}^{(i),k})^{p-1}
(t_{j,\mathrm{safe}}^{(i),k}+\epsilon)^{q-2}\\
&\times
\left[
(p+q-1)t_{j,\mathrm{safe}}^{(i),k}+p\epsilon
\right].
\end{aligned}
\label{eq:vwsptf-opt-reg-weight}
\end{equation}
On the fixed-dimensional tail covered by the theory,
$t_j^{(i),k}$ is bounded away from zero, so this safeguard is eventually
inactive. Define
\begin{equation}
\mathbf W^{(i),k}
=
\operatorname{diag}
(\omega_1^{(i),k},\ldots,\omega_{r_k}^{(i),k}),
\label{eq:vwsptf-opt-wmat}
\end{equation}
where $r_k$ is the active factor width.

Combining the two tangent majorizers gives the current surrogate, up to an
additive constant $C_k'$ that is independent of $\mathcal U$ and
$\mathcal V$,
\[
\begin{aligned}
G^k(\mathcal U,\mathcal V)
=&
\frac12
\left\|
\sqrt{\mathcal D^k}\odot\mathcal M\odot
(\mathcal U*\mathcal V-\Y)
\right\|_F^2\\
&+
\frac{\lambda}{2\ell}
\sum_{i=1}^{n_3}
\sum_{j=1}^{r_k}
\omega_j^{(i),k}
\left(
\|\bar{\mathbf u}_j^{(i)}\|_2^2+
\|\bar{\mathbf v}_j^{(i)}\|_2^2
\right)\\
&+C_k' .
\end{aligned}
\]
which is tangent to $F_\delta$ at the current iterate whenever the exact
derivatives are used.

The $0.95$ quantile of the nonzero IRLS weights is used only as an initial
curvature estimate,
\begin{equation}
\alpha_0^k
=
Q_{0.95}\left(
\{d_{abc}^k:(a,b,c)\in\OmegaSet\}
\right).
\label{eq:vwsptf-alpha}
\end{equation}
For each block, backtracking starts from this value and multiplies the trial
curvature by the backtracking factor $\gamma>1$ until the quadratic model
satisfies the block
upper-bound inequality for $G^k$. Thus the quantile itself is not assumed to
be a global Hessian bound. Separate accepted values
$\alpha_U^k$ and $\alpha_V^k$ may be obtained for the two blocks.

Let $\eta_U$ and $\eta_V$ denote the damping parameters for the two factor
blocks. We impose $0<\eta_U,\eta_V\le1$.
This range is used below to obtain a quantitative decrease directly from the
accepted damped quadratic updates.

With $\mathcal V^k$ fixed, define
\[
\widetilde{\mathcal R}_U^k
=
\mathcal D^k\odot\mathcal M\odot
(\mathcal U^k*\mathcal V^k-\Y).
\]
For each transform-domain slice,
\[
\mathbf H_U^{(i),k}
=
\alpha_U^k
\bar{\mathcal V}^{(i),k}
(\bar{\mathcal V}^{(i),k})^\top
+\lambda\mathbf W^{(i),k},
\]
\[
\mathbf G_U^{(i),k}
=
\bar{\widetilde{\mathcal R}}_U^{(i),k}
(\bar{\mathcal V}^{(i),k})^\top
+
\lambda\bar{\mathcal U}^{(i),k}\mathbf W^{(i),k},
\]
and
\begin{equation}
\bar{\mathcal U}^{(i),k+1}
=
\bar{\mathcal U}^{(i),k}
-
\eta_U
\mathbf G_U^{(i),k}
(\mathbf H_U^{(i),k})^{-1}.
\label{eq:vwsptf-u-update}
\end{equation}

After $\mathcal U^{k+1}$ is accepted, define
\[
\widetilde{\mathcal R}_V^k
=
\mathcal D^k\odot\mathcal M\odot
(\mathcal U^{k+1}*\mathcal V^k-\Y).
\]
Then
\[
\mathbf H_V^{(i),k}
=
\alpha_V^k
(\bar{\mathcal U}^{(i),k+1})^\top
\bar{\mathcal U}^{(i),k+1}
+\lambda\mathbf W^{(i),k},
\]
\[
\mathbf G_V^{(i),k}
=
(\bar{\mathcal U}^{(i),k+1})^\top
\bar{\widetilde{\mathcal R}}_V^{(i),k}
+
\lambda\mathbf W^{(i),k}\bar{\mathcal V}^{(i),k},
\]
and
\begin{equation}
\bar{\mathcal V}^{(i),k+1}
=
\bar{\mathcal V}^{(i),k}
-
\eta_V
(\mathbf H_V^{(i),k})^{-1}
\mathbf G_V^{(i),k}.
\label{eq:vwsptf-v-update}
\end{equation}

To reduce an overestimated factor width, define the tensor-pair energy
\begin{equation}
E_j^{k+1}
=
\frac{1}{n_3}
\sum_{i=1}^{n_3}
\frac{
\|\bar{\mathbf u}_j^{(i),k+1}\|_2^2+
\|\bar{\mathbf v}_j^{(i),k+1}\|_2^2
}{2}.
\label{eq:vwsptf-column-pruning-energy}
\end{equation}
Let $K_{\mathrm{pr}}$ denote the finite pruning budget and $\tau\ge0$ the
pruning threshold. Pruning is restricted to iterations $k<K_{\mathrm{pr}}$.
A component with $E_j^{k+1}\le\tau$ is proposed for deletion and is accepted only if the resulting
$F_\delta$ does not increase. This acceptance rule makes the finite-pruning
condition used in the convergence analysis explicit.

\begin{algorithm}[!t]
\caption{\WSpTFII{}-IRLS-BSUM}
\label{alg:vwsptf-irls-bsum}
\begin{algorithmic}[1]
\STATE \textbf{Input:} $\Y$, $\mathcal M$, initial width $r_0$,
$\lambda,p,q,\epsilon,\delta,\epsilon_t,
\eta_U,\eta_V\in(0,1],\tau,\gamma$,
finite pruning budget $K_{\mathrm{pr}}$, maximum iteration number $K$,
and stopping tolerance $\mathrm{tol}$.
\STATE Initialize $\mathcal U^0,\mathcal V^0$ and
$\X^0=\mathcal U^0*\mathcal V^0$.
\FOR{$k=0,1,\ldots,K-1$}
  \STATE Compute $\mathcal R^k$ and $\mathcal D^k$ by
  \eqref{eq:vwsptf-opt-residual}--\eqref{eq:vwsptf-opt-irls}.
  \STATE Compute safe pair energies and $\mathbf W^{(i),k}$ by
  \eqref{eq:vwsptf-opt-reg-weight}--\eqref{eq:vwsptf-opt-wmat}.
  \STATE Initialize $\alpha_U^k$ from \eqref{eq:vwsptf-alpha}; increase it
  by factor $\gamma$ until the $\mathcal U$-block quadratic majorizes
  $G^k$, then accept \eqref{eq:vwsptf-u-update}.
  \STATE Initialize $\alpha_V^k$ from \eqref{eq:vwsptf-alpha}; increase it
  by factor $\gamma$ until the $\mathcal V$-block quadratic majorizes
  $G^k$, then accept \eqref{eq:vwsptf-v-update}.
  \STATE Set $\X^{k+1}=\mathcal U^{k+1}*\mathcal V^{k+1}$.
  \IF{$k<K_{\mathrm{pr}}$}
    \STATE Propose deletion of pairs satisfying
    $E_j^{k+1}\le\tau$ using
    \eqref{eq:vwsptf-column-pruning-energy}; accept a deletion only if it
    does not increase $F_\delta$.
  \ENDIF
  \STATE Stop if
  $\|\X^{k+1}-\X^k\|_F/\max\{1,\|\X^k\|_F\}<\mathrm{tol}$.
\ENDFOR
\STATE \textbf{Output:} $\X=\mathcal U*\mathcal V$.
\end{algorithmic}
\end{algorithm}

The complete IRLS--BSUM procedure, including safeguarded reweighting,
backtracking, damping, and finite pruning, is summarized in
Algorithm~\ref{alg:vwsptf-irls-bsum}.

\subsection{Theoretical Analysis}
\label{subsec:vwsptf-theoretical-analysis}

The analysis concerns the fixed-$\delta$ smoothed factor objective
\eqref{eq:vwsptf-smoothed-objective}. For the $k$th outer iteration, write
\[
G^k(\mathcal U,\mathcal V)
=
G(\mathcal U,\mathcal V\mid\mathcal U^k,\mathcal V^k),
\]
and let $Q_U^k$ and $Q_V^k$ denote the accepted quadratic upper bounds for
the two block updates. Let $\mathbf B_U^k$ and $\mathbf B_V^k$ denote the
corresponding full block curvature operators induced by the accepted
transform-slice matrices $\mathbf H_U^{(i),k}$ and $\mathbf H_V^{(i),k}$; their
explicit block-diagonal forms are given in
Appendix~\ref{app:vwsptf-proof}.

\begin{assumption}
\label{ass:vwsptf-convergence}
Assume that:
\begin{enumerate}
  \item $0<p\le1$, $0<q\le1$, $p+q\ge1$,
  $\epsilon>0$, $\delta>0$, $\lambda>0$, and
  $0<\eta_U,\eta_V\le1$.
  \item The generated factor sequence is bounded. Pruning is performed only
  during the finite budget $k<K_{\mathrm{pr}}$, and each accepted pruning
  operation does not increase $F_\delta$. Hence the factor dimensions are
  fixed after the last accepted pruning operation.
  \item If $p<1$, every retained transform-domain pair energy on the
  fixed-dimensional tail is bounded away from the numerical safeguard:
  there exists $\underline t>\epsilon_t$ such that
  $t_j^{(i),k}\ge\underline t$ for all retained pairs and all sufficiently
  large $k$. Hence the safeguard is inactive on the tail and the exact
  derivative $\phi'(t_j^{(i),k})$ is used in the majorizer. This condition
  is unnecessary for $p=1$.
  \item On a compact neighborhood containing the fixed-dimensional tail,
  the block gradients of $G(\cdot\mid x)$ are uniformly Lipschitz, so the
  backtracking procedure terminates finitely. The accepted block quadratics
  majorize the corresponding blocks of $G^k$. Moreover, for constants
  $0<\mu_U\le L_U<\infty$ and $0<\mu_V\le L_V<\infty$ independent of $k$,
  their induced curvature operators satisfy
  $\mu_U\mathbf I\preceq\mathbf B_U^k\preceq L_U\mathbf I$ and
  $\mu_V\mathbf I\preceq\mathbf B_V^k\preceq L_V\mathbf I$
  on the fixed-dimensional tail.
\end{enumerate}
\end{assumption}

\begin{theorem}[Sufficient decrease for the fixed-$\delta$ objective]
\label{thm:vwsptf-monotonic}
Under Assumption~\ref{ass:vwsptf-convergence}, let $k_0$ be large enough
that pruning has stopped and the nondegeneracy condition holds. Define
\[
c_U
=
\mu_U\frac{2-\eta_U}{\eta_U},
\qquad
c_V
=
\mu_V\frac{2-\eta_V}{\eta_V}.
\]
Then $c_U,c_V>0$ and, for every $k\ge k_0$,
\begin{equation}
\begin{aligned}
F_\delta(\mathcal U^k,\mathcal V^k)
-
F_\delta(\mathcal U^{k+1},\mathcal V^{k+1})
\ge{}&
\frac{c_U}{2}
\|\mathcal U^{k+1}-\mathcal U^k\|_F^2\\
&+
\frac{c_V}{2}
\|\mathcal V^{k+1}-\mathcal V^k\|_F^2 .
\end{aligned}
\label{eq:vwsptf-main-sufficient-decrease}
\end{equation}
Consequently, the objective-value sequence converges and
\[
\sum_{k=k_0}^{\infty}
\left(
\|\mathcal U^{k+1}-\mathcal U^k\|_F^2
+
\|\mathcal V^{k+1}-\mathcal V^k\|_F^2
\right)
<\infty.
\]
In particular,
\[
\|\mathcal U^{k+1}-\mathcal U^k\|_F\to0,
\qquad
\|\mathcal V^{k+1}-\mathcal V^k\|_F\to0.
\]
\end{theorem}

\begin{theorem}[Stationarity of fixed-$\delta$ \WSpTFII{} accumulation points]
\label{thm:vwsptf-stationary}
Under Assumption~\ref{ass:vwsptf-convergence}, every accumulation point
$(\mathcal U^\star,\mathcal V^\star)$ of the fixed-dimensional tail is a
stationary point of the factor objective \eqref{eq:vwsptf-smoothed-objective}:
\[
\nabla_{\mathcal U}F_\delta(\mathcal U^\star,\mathcal V^\star)=0,
\qquad
\nabla_{\mathcal V}F_\delta(\mathcal U^\star,\mathcal V^\star)=0.
\]
Equivalently,
$0\in\partial F_\delta(\mathcal U^\star,\mathcal V^\star)$.
\end{theorem}

The proofs are provided in Appendix~\ref{app:vwsptf-proof}. The assumptions
make explicit the roles of finite nonincreasing pruning, positive retained
pair energies when $p<1$, Lipschitz block gradients, and accepted majorizing
curvatures. The quantitative decrease in
Theorem~\ref{thm:vwsptf-monotonic} follows from the actual damped updates
with $0<\eta_U,\eta_V\le1$; the vanishing first-order residuals used for
Theorem~\ref{thm:vwsptf-stationary} are then derived from that decrease and
the update relations rather than imposed as an additional assumption.

\begin{table}[!t]
\centering
\caption{Matrix-based baseline methods and their objective functions.}
\label{tab:matrix_compared_methods}
\scriptsize
\renewcommand{\arraystretch}{1.28}
\setlength{\tabcolsep}{2pt}
\begin{tabular}{
  @{}
  >{\centering\arraybackslash}p{0.165\columnwidth}
  >{\raggedright\arraybackslash}p{0.735\columnwidth}
  @{}
}
\toprule
\multicolumn{1}{c}{\textbf{Compared Methods}} & \multicolumn{1}{c}{\textbf{Algorithm Formulation}} \\
\midrule

RPCA~\cite{CandesEtAl2011RPCA}
&
\(
\displaystyle
\begin{aligned}
\min_{\mathbf X,\mathbf N}\;&
\lambda\|\mathbf X\|_{*}
+\|\mathbf M\odot\mathbf N\|_{1},\\[-1mm]
&\mathrm{s.t.}\;
\mathbf Y=\mathbf X+\mathbf N
\end{aligned}
\)
\\
\addlinespace[2pt]

UNIFY~\cite{CabralEtAl2013}
&
\(
\displaystyle
\begin{aligned}
\min_{\mathbf{U},\mathbf{V},\mathbf{X}}\;&
\frac{\lambda}{2}
\left(
\|\mathbf{U}\|_{F}^{2}
+\|\mathbf{V}\|_{F}^{2}
\right)
+\|\mathbf{M}\odot(\mathbf{Y}-\mathbf{X})\|_{1},
\\[-1mm]
&\mathrm{s.t.}\;
\mathbf{X}=\mathbf{U}\mathbf{V}
\end{aligned}
\)
\\
\addlinespace[2pt]

GIRNN~\cite{HuangEtAl2020GIRNN}
&
\(
\displaystyle
\begin{aligned}
\min_{\mathbf X,\mathbf N}\;&
\lambda\|\mathbf X\|_{w,*}
+\|\mathbf M\odot\mathbf N\|_{1},\\[-1mm]
&\mathrm{s.t.}\;
\mathbf Y=\mathbf X+\mathbf N
\end{aligned}
\)
\\
\addlinespace[2pt]

RLMF~\cite{ChenEtAl2022RLTF}
&
\(
\displaystyle
\begin{aligned}
\min_{\mathbf{U},\mathbf{V}}\;&
\frac{\lambda}{2}
\left(
\|\mathbf{U}\|_{w,F}^{2}
+\|\mathbf{V}\|_{w,F}^{2}
\right)
\\[-1mm]
&+
\|\mathbf{M}\odot(\mathbf{U}\mathbf{V}-\mathbf{Y})\|_{1}
\end{aligned}
\)
\\

\bottomrule
\end{tabular}
\end{table}

\begin{table}[!t]
\centering
\caption{Tensor-based baseline methods and their objective functions.}
\label{tab:tensor_compared_methods}
\scriptsize
\renewcommand{\arraystretch}{1.28}
\setlength{\tabcolsep}{2pt}
\begin{tabular}{
  @{}
  >{\centering\arraybackslash}p{0.165\columnwidth}
  >{\raggedright\arraybackslash}p{0.735\columnwidth}
  @{}
}
\toprule
\multicolumn{1}{c}{\textbf{Compared Methods}} & \multicolumn{1}{c}{\textbf{Algorithm Formulation}} \\
\midrule

SNN~\cite{LiuEtAl2013SNN}
&
\(
\displaystyle
\begin{aligned}
\min_{\X,\mathcal N}\;&
\sum_{i=1}^{3}\lambda_i\|\mathbf X_{(i)}\|_{*}
+\|\mathcal M\odot\mathcal N\|_{1},\\[-1mm]
&\mathrm{s.t.}\;
\Y=\X+\mathcal N
\end{aligned}
\)
\\
\addlinespace[2pt]

TNN~\cite{LuEtAl2020TRPCA}
&
\(
\displaystyle
\begin{aligned}
\min_{\X,\mathcal N}\;&
\lambda\|\X\|_{*}
+\|\mathcal M\odot\mathcal N\|_{1},\\[-1mm]
&\mathrm{s.t.}\;
\Y=\X+\mathcal N
\end{aligned}
\)
\\
\addlinespace[2pt]

RLTF~\cite{ChenEtAl2022RLTF}
&
\(
\displaystyle
\begin{aligned}
\min_{\mathcal{U},\mathcal{V}}\;&
\frac{\lambda}{2}
\left(
\|\mathcal{U}\|_{w,F}^{2}
+\|\mathcal{V}\|_{w,F}^{2}
\right)
\\[-1mm]
&+
\|\mathcal{M}\odot(\mathcal{U}*\mathcal{V}-\Y)\|_{1}
\end{aligned}
\)
\\
\addlinespace[2pt]

WNNTF
&
\(
\displaystyle
\begin{aligned}
\min_{\mathcal{U},\mathcal{V},\X}\;&
\|\mathcal{P}_{\OmegaSet}(\Y-\X)\|_{1}
+\frac{\lambda}{2}
\left(
\|\mathcal{U}\|_{W,*}
+\|\mathcal{V}\|_{W,*}
\right),
\\[-1mm]
&\mathrm{s.t.}\;
\X=\mathcal{U}*\mathcal{V}
\end{aligned}
\)
\\
\addlinespace[2pt]

SPTF
&
\(
\displaystyle
\begin{aligned}
\min_{\mathcal{U},\mathcal{V},\X}\;&
\|\mathcal{P}_{\OmegaSet}(\Y-\X)\|_{1}
+\frac{\lambda}{2}
\left(
\|\mathcal{U}\|_{S_{p_1}}^{p_1}
+\|\mathcal{V}\|_{S_{p_2}}^{p_2}
\right),
\\[-1mm]
&\mathrm{s.t.}\;
\X=\mathcal{U}*\mathcal{V},
\quad
\frac{1}{p}=\frac{1}{p_1}+\frac{1}{p_2}
\end{aligned}
\)
\\
\addlinespace[2pt]

WTSP~\cite{LiuEtAl2020WeightedTSchatten}
&
\(
\displaystyle
\begin{aligned}
\min_{\X}\;&
\|\mathcal P_{\OmegaSet}(\Y-\X)\|_{1}\\[-1mm]
&+\lambda\|\X\|_{W,S_p}^{p}
\end{aligned}
\)
\\

\addlinespace[2pt]

GTNLN~\cite{ShuEtAl2026GTNLN}
&
\(
\displaystyle
\begin{aligned}
\min_{\X,\mathcal{N}}\;&
\|\nabla(\X)\|_{\mathrm{GTNLN}}
+\lambda\|\mathcal{N}\|_{1},
\\[-1mm]
&\mathrm{s.t.}\;
\mathcal{P}_{\OmegaSet}(\X+\mathcal{N})
=
\mathcal{P}_{\OmegaSet}(\Y)
\end{aligned}
\)
\\

\bottomrule
\end{tabular}
\end{table}

\subsection{Computational Complexity}
\label{subsec:vwsptf-complexity}

Let $r_k$ denote the active factor width. Reconstructing the tensor,
computing the observed residual, and forming the IRLS coefficients cost
$\mathcal O(n_1n_2n_3r_k+|\OmegaSet|)$. Pair energies, regularization
coefficients, and pruning statistics cost
$\mathcal O((n_1+n_2)n_3r_k)$. The two block updates require
\[
\mathcal O\!\left(
n_1n_2n_3r_k+(n_1+n_2)n_3r_k^2+n_3r_k^3
\right)
\]
per accepted trial. If at most $N_{\mathrm{bt}}$ backtracking trials are
used for either block, this term is multiplied by $N_{\mathrm{bt}}$.
Under Assumption~\ref{ass:vwsptf-convergence}, $N_{\mathrm{bt}}$ is
uniformly bounded. Let $C_{\mathcal L}(n_3)$ denote the cost of applying the transform or
its inverse to one length-$n_3$ tube. These operations contribute
\[
\mathcal O\!\left(
[n_1n_2+(n_1+n_2)r_k]\,C_{\mathcal L}(n_3)
\right).
\]
For a dense transform matrix, $C_{\mathcal L}(n_3)=\mathcal O(n_3^2)$;
structured transforms may reduce this to $\mathcal O(n_3\log n_3)$.
\WSpTFII{} therefore avoids compact SVDs in its main factor updates, while finite
pruning can reduce $r_k$ as inactive components are removed. This structural
advantage is not, by itself, a claim of lower wall-clock time for every tensor
size. The storage complexity is
$\mathcal O(n_1n_2n_3+(n_1+n_2)r_kn_3)$.

\section{Experiments}
\label{sec:experiments}
We evaluate the proposed \WSpTFI{} and \WSpTFII{} models on synthetic tensor
completion and three real-data applications: color-image restoration,
hyperspectral image inpainting, and printed circuit board (PCB) defect detection. The experiments are
designed to assess reconstruction accuracy, computational behavior, parameter
sensitivity, and robustness under different degradation conditions. All
experiments are implemented in MATLAB R2021b on a Windows 10 workstation
equipped with an Intel Core i7-11800H CPU at 2.30 GHz and 8 GB of RAM.
For controlled matrix comparisons, we use weighted Schatten-$p$ matrix
factorization I (\WSpMFI{}) and weighted Schatten-$p$ matrix factorization II
(\WSpMFII{}), obtained as the matrix reductions of \WSpTFI{} and \WSpTFII{},
respectively. \WSpTFI{} and \WSpMFI{} use the
symmetric factor-exponent setting $p_1=p_2=2p$, which satisfies
$1/p=1/p_1+1/p_2$.

For the proposed tensor models and their controlled tensor variants, we
use the normalized DCT along the third mode. This
is the transform adopted in all experiments. The DCT keeps all
transform-domain quantities real, avoiding the complex arithmetic associated
with Fourier-domain implementations; its reflexive boundary behavior is also
better matched to nonperiodic terminal frames or spectral bands and can reduce
boundary artifacts. In addition, fast DCT implementations retain low
transform cost. These properties have motivated DCT-based tensor completion
and t-SVD models in prior work
\cite{MadathilGeorge2018DCT,XuZhaoNg2019DCT,LuPengWei2019TransformTNN}.
With the normalized DCT matrix used here,
$\mathbf L^*\mathbf L=\mathbf L\mathbf L^*=\mathbf I$, so
\eqref{eq:transform-condition} holds with $\ell=1$. Competing baselines
otherwise follow the transform conventions of their original implementations.

The matrix baselines comprise robust principal component analysis (RPCA)
\cite{CandesEtAl2011RPCA}, the unifying nuclear-norm/bilinear-factorization
method (UNIFY) \cite{CabralEtAl2013}, the generalized reweighted iterative
nuclear/Frobenius-norm method (GIRNN) \cite{HuangEtAl2020GIRNN}, and
reweighted low-rank matrix factorization (RLMF) \cite{ChenEtAl2022RLTF}.
The tensor baselines comprise the sum-of-nuclear-norms method (SNN)
\cite{LiuEtAl2013SNN}, the TNN method
\cite{LuEtAl2020TRPCA}, the tensor version of reweighted low-rank
factorization (RLTF) \cite{ChenEtAl2022RLTF}, weighted nuclear-norm tensor
factorization (WNNTF), Schatten-$p$ tensor factorization (SPTF), the weighted
t-Schatten-$p$ (WTSP) method \cite{LiuEtAl2020WeightedTSchatten}, and the
gradient tensor nuclear $\ell_1$--$\ell_2$ norm (GTNLN) method
\cite{ShuEtAl2026GTNLN}. WNNTF and SPTF are controlled special cases of
\WSpTFI{} obtained by selecting the corresponding parameter settings. See Tables~\ref{tab:matrix_compared_methods} and
\ref{tab:tensor_compared_methods} for details.

In Tables~\ref{tab:matrix_compared_methods} and
\ref{tab:tensor_compared_methods}, $\|\cdot\|_{w,*}$ and
$\|\cdot\|_{w,F}$ denote the reweighted nuclear and Frobenius norms,
respectively. The symbol $\mathbf{X}_{(i)}$ denotes the mode-$i$ unfolding
of $\X$, with $\lambda_i\geq0$ and $\sum_{i=1}^{3}\lambda_i=1$.
Moreover, $\|\nabla(\X)\|_{\mathrm{GTNLN}}$ denotes the gradient tensor
nuclear $\ell_1$--$\ell_2$ norm applied to the temporal gradient tensor.
The matrix counterparts \WSpMFI{} and \WSpMFII{} are obtained by setting
the third-mode size to $n_3=1$, for which the normalized DCT becomes the
identity transform and the t-product reduces to ordinary matrix
multiplication. They are included as controlled counterparts to assess the
benefit of retaining the tensor structure, rather than as independent tensor
models.

\subsection{Synthetic Data}

We first evaluate completion robustness on synthetic third-order tensors under
Gaussian, Gaussian mixture model (GMM), and Laplace noise. The clean tensor
$\X\in\R^{100\times100\times10}$ is generated with prescribed tubal rank
$r$. In the transform domain, its $i$th frontal slice is constructed as
$\bar{\X}^{(i)}=\mathbf{P}\mathbf{C}^{(i)}\mathbf{Q}^{\top}$,
$i=1,\ldots,n_3$, where
$\mathbf{P},\mathbf{Q}\in\R^{100\times r}$ are shared subspace matrices and
$\mathbf{C}^{(i)}\in\R^{r\times r}$ is the slice-dependent coefficient
matrix, generated randomly so that at least one $\mathbf{C}^{(i)}$ has rank
$r$. By \eqref{eq:transform-tubal-rank}, the resulting tensor
therefore satisfies $\operatorname{rank}_{t}(\X)=r$. The inverse transform
maps the tensor back to the original domain.

The clean tensor is normalized to $[0,1]$, after which a Bernoulli observation
mask with sampling ratio $0.5$ is generated. Thus, approximately half of the
entries are available for reconstruction, with the observations denoted by
$\Y_{\OmegaSet}=\mathcal{P}_{\OmegaSet}(\Y)$. The noisy tensor is generated as
$\Y=\operatorname{clip}(\X+\mathcal{N},0,1)$, where $\mathcal{N}$ is the
noise tensor and $\operatorname{clip}(\cdot,0,1)$ restricts the dynamic range
to $[0,1]$. For Gaussian and Laplace noise, the noise level is controlled by
the signal-to-noise ratio (SNR); the Laplace scale parameter $b$ in
$\mathrm{Laplace}(0,b)$ is selected to match the target SNR. For GMM noise,
each entry of $\mathcal{N}$ is independently drawn from the two-component
zero-mean mixture
$\alpha_1\mathcal{N}(0,\sigma_1^2)+\alpha_2\mathcal{N}(0,\sigma_2^2)$.
We set $\alpha_2=\alpha_1/9=0.1$ and
$\sigma_2^2=100\sigma_1^2$, where $\alpha_k$ and $\sigma_k^2$ are the
mixture probability and variance of the $k$th component, respectively. The
resulting variance is $\sigma^2=\sum_{k=1}^{2}\alpha_k\sigma_k^2$, and the
SNR is defined as
\begin{equation*}
  \mathrm{SNR}
  =
  10\log_{10}
  \left(
  \frac{\|\X\|_F^2}{n_1n_2n_3\sigma^2}
  \right).
\end{equation*}
The Gaussian variance, GMM variance, and Laplace scale are adjusted so that
all three experiments use the same target SNR of $10$ dB.

For the synthetic experiments, the model parameters are set as follows:
\WSpMFI{} uses $\lambda=10$, $p=0.95$, and $q=0.65$;
\WSpMFII{} uses $\lambda=1.2\sqrt{\max(n_1,n_2)n_3}$, $p=0.95$, and
$q=0.55$; \WSpTFI{} uses
$\lambda=0.3\sqrt{\max(n_1,n_2)n_3}$, $p=0.5$, and $q=0.95$; and
\WSpTFII{} uses $\lambda=1.2\sqrt{\max(n_1,n_2)n_3}$, $p=0.5$, and
$q=0.6$.

\subsubsection{Completion under Different Noise Distributions}
\begin{table*}[!t]
\centering
\caption{Quantitative results for synthetic tensor completion under different noise distributions. For the reconstruction-quality metrics, the best and second-best results among all methods are marked in \textbf{bold} and \underline{underlined}, respectively. Lower running times are better.}
\label{tab:synthetic_noise_results}
\scriptsize
\resizebox{\textwidth}{!}{
\begin{tabular}{@{}llccccccccc@{}}
\toprule
\multirow{2}{*}{Method Type} &
\multirow{2}{*}{Method} &
\multicolumn{3}{c}{Gaussian Noise} &
\multicolumn{3}{c}{GMM Noise} &
\multicolumn{3}{c}{Laplace Noise} \\
\cmidrule(lr){3-5}\cmidrule(lr){6-8}\cmidrule(lr){9-11}
& & PSNR & SSIM & Time (s)
& PSNR & SSIM & Time (s)
& PSNR & SSIM & Time (s) \\
\midrule
\multirow{6}{*}{Matrix-Based Methods}
& RPCA & 22.11 & 0.78 & 1.65 & 27.44 & 0.92 & 1.62 & 23.64 & 0.83 & 1.70 \\
& Unifying & 22.47 & 0.79 & 0.29 & 28.47 & 0.94 & 0.31 & 23.99 & 0.84 & 0.32 \\
& GIRNN & 22.42 & 0.79 & 1.71 & 27.87 & 0.93 & 1.68 & 23.84 & 0.84 & 1.75 \\
& RLMF & 24.98 & 0.87 & 0.25 & 27.05 & 0.91 & 0.33 & 25.24 & 0.88 & 0.31 \\
& \WSpMFI{} (Ours) & 25.08 & 0.87 & 0.29 & 28.70 & 0.94 & 0.37 & 26.04 & 0.89 & 0.37 \\
& \WSpMFII{} (Ours) & 24.58 & 0.86 & 2.41 & 28.39 & 0.94 & 4.12 & 25.53 & 0.88 & 3.47 \\
\midrule
\multirow{9}{*}{Tensor-Based Methods}
& SNN & 24.57 & 0.88 & 1.02 & 30.30 & 0.96 & 1.01 & 25.79 & 0.90 & 1.08 \\
& TNN & 25.14 & 0.89 & 1.43 & 31.52 & 0.97 & 1.42 & 26.81 & 0.92 & 1.48 \\
& WTSP & 25.00 & 0.84 & 1.67 & 27.54 & 0.91 & 1.38 & 25.51 & 0.86 & 1.70 \\
& RLTF & 24.70 & 0.86 & 0.52 & 30.81 & 0.97 & 0.53 & 28.55 & 0.94 & 0.55 \\
& WNNTF & 24.16 & 0.85 & 0.53 & 33.44 & 0.98 & 0.58 & 27.44 & 0.92 & 0.58 \\
& SPTF & 26.24 & 0.90 & 0.55 & 30.62 & 0.97 & 0.71 & 27.98 & 0.93 & 0.60 \\
& GTNLN & 22.92 & 0.79 & 0.98 & 25.72 & 0.88 & 0.97 & 23.79 & 0.82 & 0.96 \\
& \WSpTFI{} (Ours) & \underline{26.43} & \underline{0.91} & 0.55 & \textbf{33.51} & \textbf{0.98} & 0.67 & \underline{28.71} & \underline{0.94} & 0.69 \\
& \WSpTFII{} (Ours) & \textbf{27.00} & \textbf{0.92} & 1.30 & \underline{33.44} & \underline{0.98} & 1.05 & \textbf{28.93} & \textbf{0.95} & 1.05 \\
\bottomrule
\end{tabular}}
\end{table*}

We compare the matrix- and tensor-based methods summarized in
Tables~\ref{tab:matrix_compared_methods} and
\ref{tab:tensor_compared_methods}. Reconstruction quality is evaluated by
peak signal-to-noise ratio (PSNR) and structural similarity index (SSIM),
and computational cost is reported as running time. Higher PSNR and SSIM
values indicate better reconstruction quality, whereas lower running times
indicate greater computational efficiency. 

The relative squared error (RSE) is defined as
\begin{equation*}
  \operatorname{RSE}
  =
  \frac{\|\hat{\X}-\X\|_{F}^{2}}{\|\X\|_{F}^{2}},
\end{equation*}
where $\hat{\X}$ denotes the recovered tensor. All metrics are evaluated on
the complete tensor rather than only on the observed entries. The PSNR is
computed as
\begin{equation*}
  \operatorname{PSNR}
  =
  10\log_{10}
  \left(
  \frac{n_1 n_2 n_3\|\X\|_{\infty}^{2}}
  {\|\X-\hat{\X}\|_{F}^{2}}
  \right).
\end{equation*}
The SSIM between $\X$ and $\hat{\X}$ is defined as
\begin{equation*}
  \operatorname{SSIM}(\X,\hat{\X})
  =
  \frac{
  (2\mu_{\X}\mu_{\hat{\X}}+C_1)
  (2\sigma_{\X\hat{\X}}+C_2)
  }
  {
  (\mu_{\X}^{2}+\mu_{\hat{\X}}^{2}+C_1)
  (\sigma_{\X}^{2}+\sigma_{\hat{\X}}^{2}+C_2)
  },
\end{equation*}
where $\mu_{\X}$ and $\mu_{\hat{\X}}$ denote the means,
$\sigma_{\X}^{2}$ and $\sigma_{\hat{\X}}^{2}$ the variances, and
$\sigma_{\X\hat{\X}}$ the covariance of $\X$ and $\hat{\X}$.
The positive constants $C_1$ and $C_2$ stabilize the denominator.

Table~\ref{tab:synthetic_noise_results} summarizes reconstruction quality
and running time under Gaussian, GMM, and Laplace noise. Among the
matrix-based methods, \WSpMFI{} yields the highest PSNR in all three cases:
$25.08$, $28.70$, and $26.04$ dB, respectively. It also attains the highest
or tied-highest SSIM within the matrix group. \WSpMFII{} remains competitive
in reconstruction accuracy, particularly under GMM and Laplace noise, but
requires substantially more computation. In terms of running time, RLMF is
the fastest matrix method under Gaussian and Laplace noise, whereas UNIFY is
the fastest under GMM noise.

Among the tensor-based methods, \WSpTFII{} achieves the highest PSNR and
SSIM under Gaussian and Laplace noise, whereas \WSpTFI{} gives the highest
PSNR under GMM noise. In particular, \WSpTFI{} reaches $33.51$ dB for the
GMM case, compared with $33.44$ dB for both WNNTF and \WSpTFII{}. RLTF is
the fastest tensor method in all three settings. Overall, the proposed
tensor models consistently achieve high reconstruction accuracy across the
three noise distributions, suggesting that the factorized tensor
regularization effectively exploits the multiway structure of the synthetic
data, albeit with a higher computational cost than the fastest baseline.

\subsubsection{Convergence Behavior over 150 Iterations}

To examine empirical convergence behavior, all matrix- and tensor-based
methods are run for $150$ outer iterations using the same clean tensor,
observation mask, and noise realization. Early stopping is disabled to
obtain complete and directly comparable iteration histories. For methods
with multiple alternating updates, one full cycle over the primal variables
is counted as one outer iteration. The reconstruction after iteration $k$ is
denoted by $\hat{\X}^{k}$, and its PSNR is recorded for
$k=1,\ldots,150$.

\begin{figure*}[!t]
    \centering
    \includegraphics[width=\textwidth]{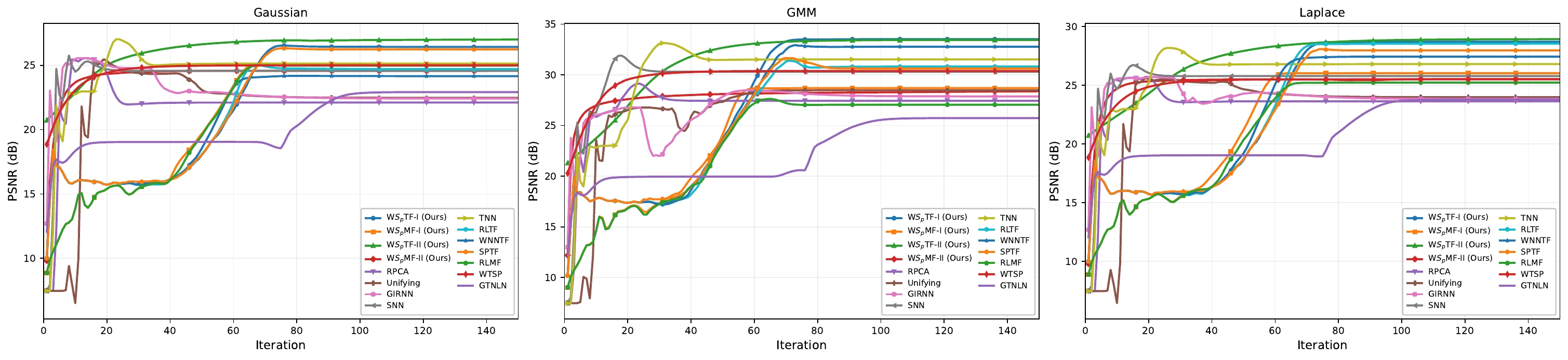}
    \caption{PSNR trajectories of the matrix- and tensor-based methods over
    $150$ outer iterations under Gaussian, GMM, and Laplace noise. The horizontal and vertical axes denote
    the iteration number and reconstruction PSNR, respectively.}
    \label{fig:synthetic_convergence_150}
\end{figure*}

Figure~\ref{fig:synthetic_convergence_150} plots the PSNR trajectories for
Gaussian, GMM, and Laplace noise from left to right. Under Gaussian noise,
\WSpTFI{} reaches a final PSNR of $26.43$ dB, while \WSpTFII{} reaches
$27.00$ dB. Under GMM noise, the corresponding final values are $33.51$
and $33.44$ dB, and under Laplace noise they are $28.71$ and $28.93$ dB,
respectively. The two tensor methods stabilize after approximately
$70$--$80$ outer iterations in these experiments. Some baselines plateau
earlier at lower PSNR values, whereas several trajectories show transient
oscillations during the initial iterations. These empirical curves indicate
stable late-iteration behavior for the proposed tensor methods across the
three tested noise distributions.

\subsubsection{Phase-Transition Analysis}
\begin{figure}[!t]
  \centering
  \includegraphics[width=\columnwidth]{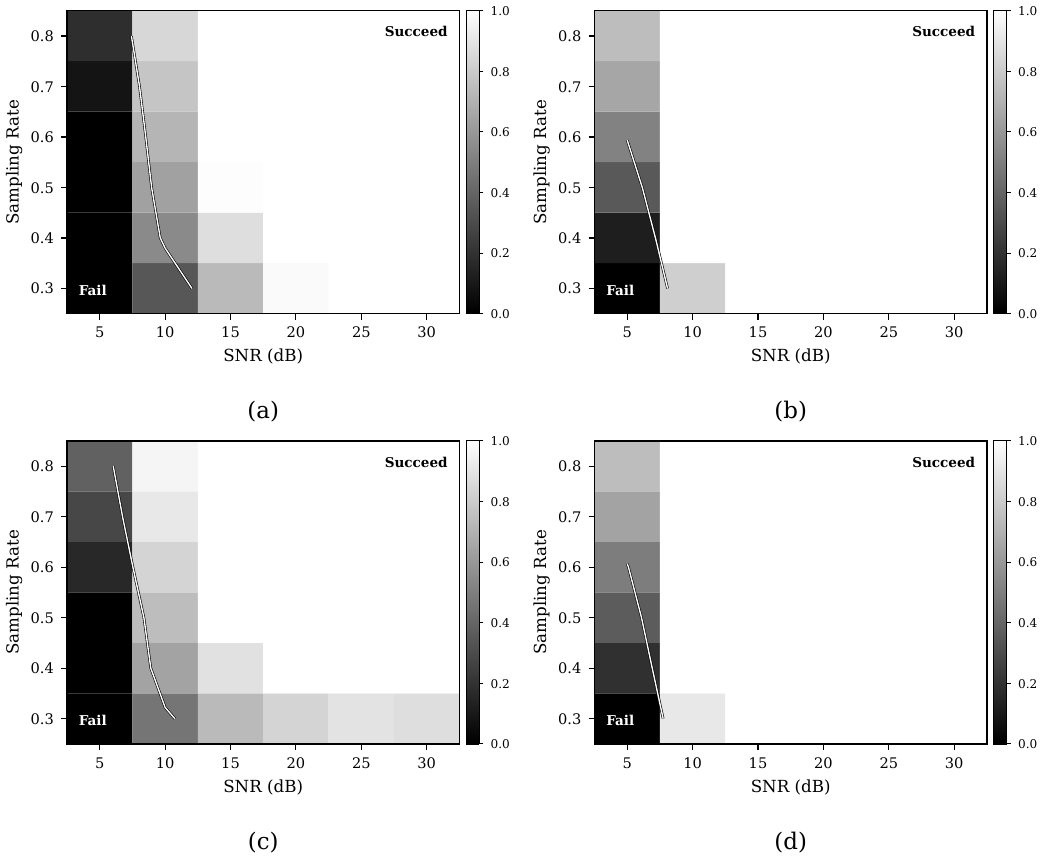}
  \caption{Phase-transition diagrams of the four factorization methods on
  synthetic data: (a) \WSpMFII{}, (b) \WSpTFII{}, (c) \WSpMFI{}, and
  (d) \WSpTFI{}. The horizontal axis denotes input SNR, the vertical axis
  denotes sampling ratio, and gray level represents empirical completion
  probability from $0$ (black) to $1$ (white).}
  \label{fig:synthetic_phase_transition}
\end{figure}

\begin{figure*}[!t]
  \centering
  \includegraphics[width=\textwidth]{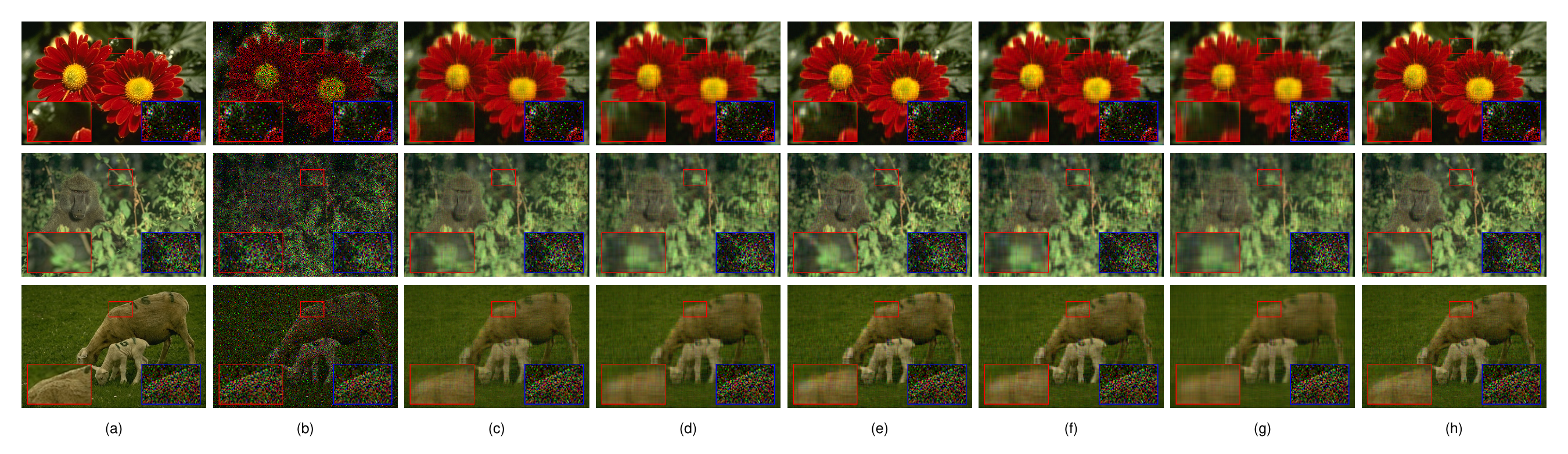}
  \caption{Visual restoration results of the matrix-based methods on BSD
  color images. (a) Original image; (b) incomplete and corrupted
  observation; (c) RPCA; (d) Unifying; (e) GIRNN; (f) RLMF;
  (g) \WSpMFI{} (Ours); and (h) \WSpMFII{} (Ours).}
  \label{fig:bsd-matrix-restoration-results}
\end{figure*}

\begin{figure*}[!t]
  \centering
  \includegraphics[width=\textwidth]{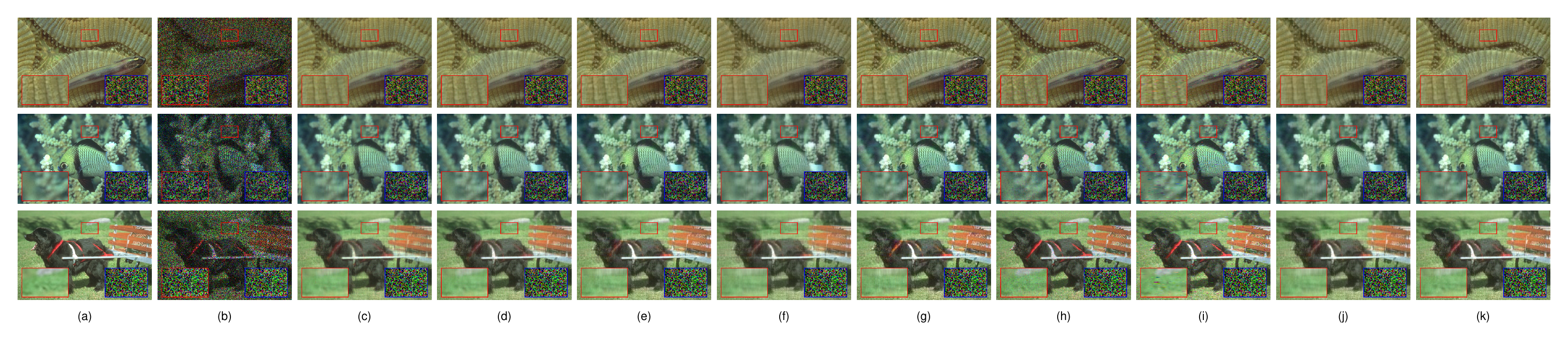}
  \caption{Visual restoration results of the tensor-based methods on BSD
  color images. (a) Original image; (b) incomplete and corrupted
  observation; (c) SNN; (d) TNN; (e) RLTF; (f) WNNTF; (g) SPTF;
  (h) WTSP; (i) GTNLN; (j) \WSpTFI{} (Ours); and (k) \WSpTFII{} (Ours).}
  \label{fig:bsd-tensor-restoration-results}
\end{figure*}

\begin{figure}[!t]
  \centering
  \includegraphics[width=\columnwidth]{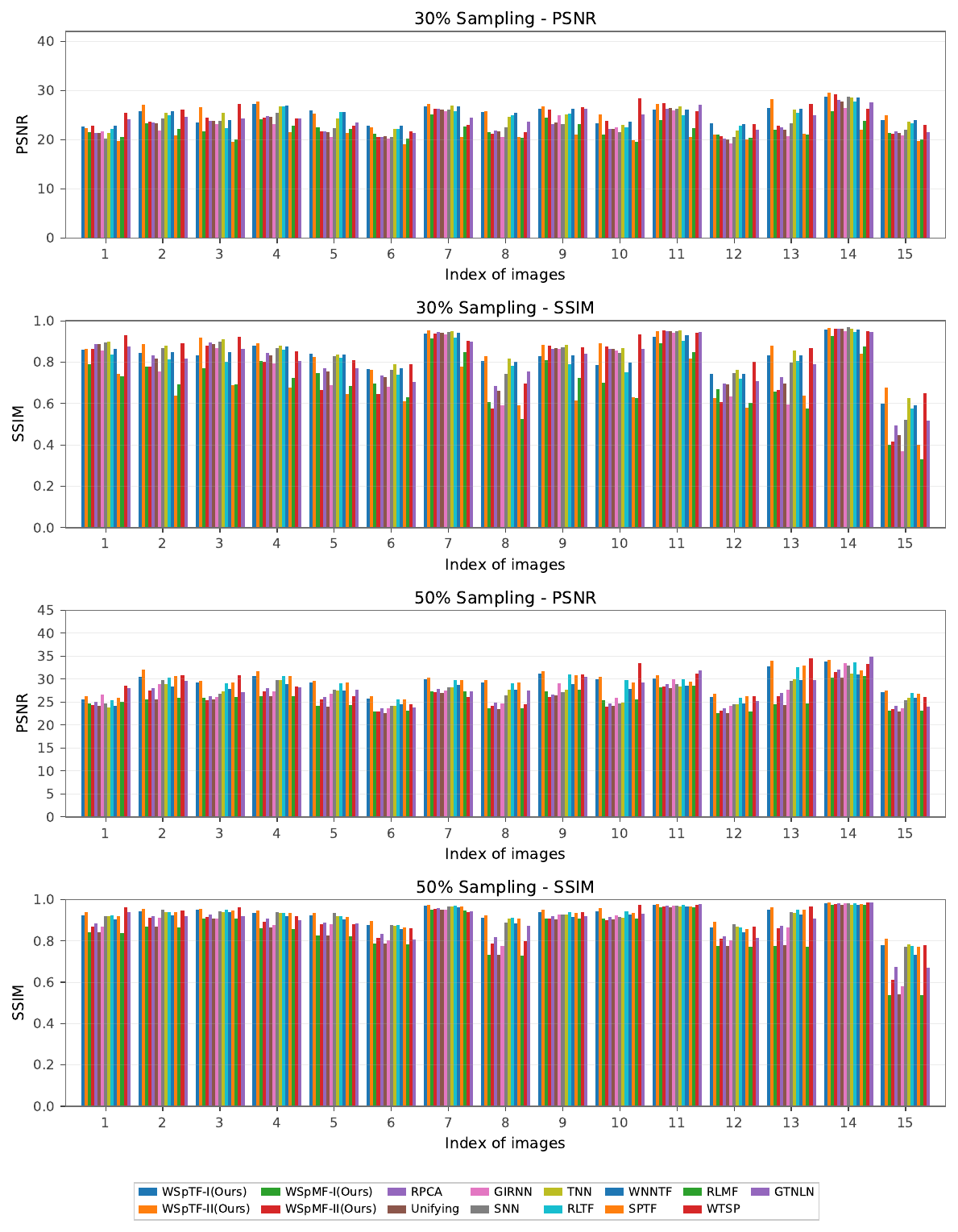}
  \caption{Image-wise PSNR and SSIM for the 15 BSD test images at sampling
ratios of $30\%$ and $50\%$.}
  \label{fig:bsd-color-imagewise-results}
\end{figure}

\begin{figure}[!t]
  \centering
  \includegraphics[width=\columnwidth]{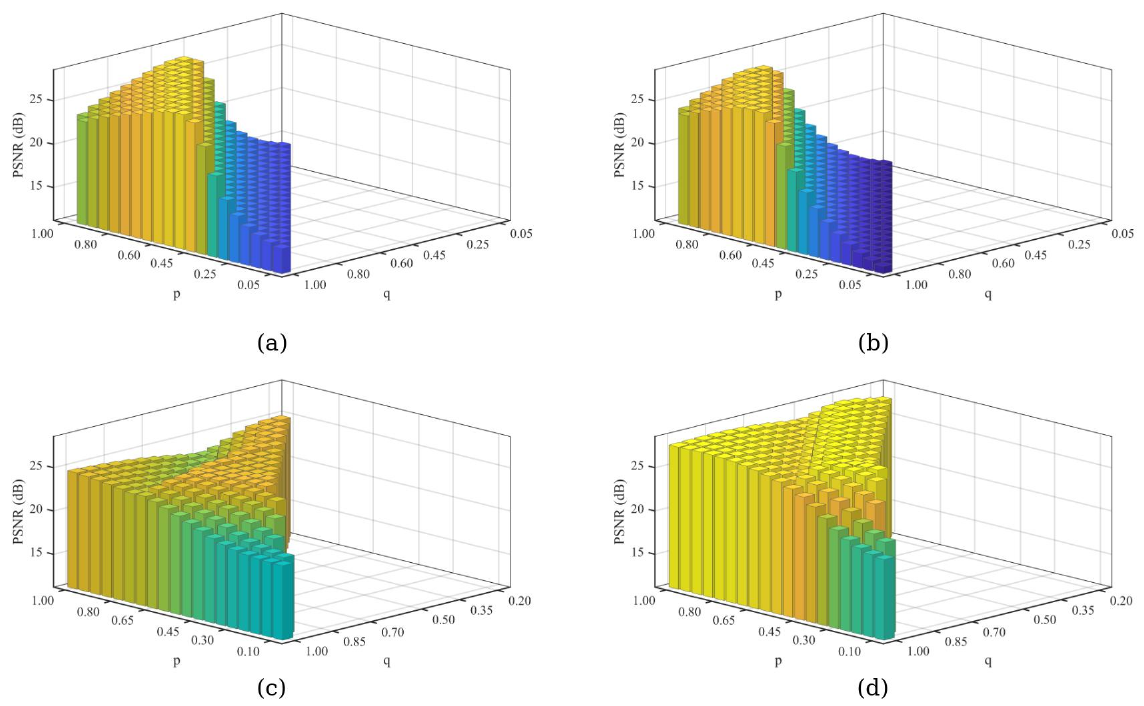}
  \caption{Parameter sensitivity of the four proposed models in the color
  image restoration experiment. The horizontal axes denote $p$ and $q$,
  and the vertical axis denotes PSNR (dB). (a) \WSpTFII{}; (b) \WSpMFII{};
  (c) \WSpMFI{}; and (d) \WSpTFI{}. For \WSpTFI{} and \WSpMFI{}, only parameter pairs
  satisfying $q\geq 1-\frac{4p^2}{1+4p}$ are retained; for \WSpTFII{} and
  \WSpMFII{}, only pairs satisfying $1\leq p+q\leq 2$ are retained.}
  \label{fig:color-parameter-selection}
\end{figure}

We next assess completion stability using phase-transition experiments with
the matrix- and tensor-based factorization methods. The sampling ratio is
varied over $\{0.3,0.4,0.5,0.6,0.7,0.8\}$ and the input SNR over
$\{5,10,15,20,25,30\}$ dB under additive Gaussian noise. For each
sampling-ratio/SNR pair, $100$ independent trials are performed. In each
trial, the low-rank factors, observation mask, and noise realization are
regenerated independently, while the tensor dimensions and prescribed rank
remain fixed.

A trial is considered successful when the relative squared error satisfies
$\operatorname{RSE}(\hat{\X},\X)
=\|\hat{\X}-\X\|_F^2/\|\X\|_F^2<10^{-2}$.
For method $m$, sampling ratio $\rho$, and input SNR $\eta$, the empirical
completion probability is
$\widehat{P}_{m}(\rho,\eta)
=\frac{1}{100}\sum_{s=1}^{100}\mathbb{I}
\left\{
\left\|\hat{\X}_{m,s}^{(\rho,\eta)}-\X_s\right\|_F^2/
\|\X_s\|_F^2<10^{-2}
\right\}$,
where $\mathbb{I}\{\cdot\}$ is the indicator function and
$\hat{\X}_{m,s}^{(\rho,\eta)}$ is the reconstruction produced by method
$m$ in trial $s$. For each method, the resulting $6\times6$ probability
matrix is displayed as a phase diagram, with input SNR on the horizontal
axis, sampling ratio on the vertical axis, and gray level representing the
empirical success probability from $0$ to $1$. A larger high-probability
region, with the transition boundary shifted toward the lower-left corner,
indicates successful completion under fewer observations and stronger noise.

As shown in Fig.~\ref{fig:synthetic_phase_transition}, the empirical completion
probability increases with both sampling ratio and SNR. \WSpTFI{} and
\WSpTFII{} exhibit larger high-probability regions than their matrix
counterparts, with transition boundaries shifted toward lower sampling ratios
and SNRs. When the SNR is approximately $10$ dB or higher and the sampling
ratio is at least $0.4$, the tensor variants enter a predominantly
high-probability region, whereas the matrix variants generally require more
favorable conditions. At the lowest sampling ratio of $0.3$, completion remains
difficult for all four methods. Within the matrix group, \WSpMFII{} exhibits
a broader successful-completion region than \WSpMFI{}. These results indicate
that tensor modeling improves empirical completion stability in the tested
sampling/noise regime, with an additional advantage for the variational
formulation in several challenging settings.

\subsection{Color Image Restoration}
\begin{table*}[!t]
\centering
\caption{Average PSNR, SSIM, FSIM, and running time for color-image restoration at different sampling ratios. For the reconstruction-quality metrics, the best and second-best available results among all methods are marked in \textbf{bold} and \underline{underlined}, respectively. Lower running times are better.}
\label{tab:bsd_color_results}
\scriptsize
\setlength{\tabcolsep}{4.5pt}
\resizebox{\textwidth}{!}{
\begin{tabular}{@{}llcccccccc@{}}
\toprule
\multirow{2}{*}{Method Type} &
\multirow{2}{*}{Method} &
\multicolumn{2}{c}{PSNR (dB)} &
\multicolumn{2}{c}{SSIM} &
\multicolumn{2}{c}{FSIM} &
\multicolumn{2}{c}{Time (s)} \\
\cmidrule(lr){3-4}\cmidrule(lr){5-6}\cmidrule(lr){7-8}\cmidrule(lr){9-10}
& & 30\% Sampling & 50\% Sampling
  & 30\% Sampling & 50\% Sampling
  & 30\% Sampling & 50\% Sampling
  & 30\% Sampling & 50\% Sampling \\
\midrule
\multirow{6}{*}{Matrix-Based Methods}
& RPCA & 23.23 & 26.46 & 0.81 & 0.89 & 0.80 & 0.87 & 6.72 & 7.60\\
& Unifying & 23.14 & 25.19 & 0.80 & 0.84 & 0.80 & 0.83 & 1.55 & 1.86 \\
& GIRNN & 22.53 & 27.04 & 0.76 & 0.87 & 0.78 & 0.86 & 6.51 & 7.67\\
& RLMF & 21.47 & 25.68 & 0.67 & 0.84 & 0.74 & 0.84 & 1.04 & 1.60\\
& \WSpMFI{} (Ours) & 22.76 & 25.47 & 0.74 & 0.84 & 0.77 & 0.83 & 1.21 & 1.65\\
& \WSpMFII{} (Ours) & 23.79 & 25.82 & 0.77 & 0.87 & 0.81 & 0.85 & 23.21 & 54.85\\
\midrule
\multirow{9}{*}{Tensor-Based Methods}
& SNN & 23.42 & 27.40 & 0.83 & 0.92 & 0.79 & 0.89 & 11.80 & 13.39\\
& TNN & 24.84 & 27.32 & \textbf{0.86} & 0.92 & 0.84 & 0.89 & 7.55 & 8.57 \\
& RLTF & 24.50 & 29.28 & 0.80 & 0.92 & 0.81 & 0.91 & 4.34 & 5.29\\
& WNNTF & 25.23 & 27.63 & 0.83 & 0.91 & 0.83 & 0.88 & 3.41 & 4.11\\
& SPTF & 20.53 & 29.18 & 0.66 & 0.92 & 0.72 & 0.91 & 3.46 & 4.17\\
& WTSP & 24.88 & 29.08 & 0.85 & 0.92 & \textbf{0.85} & 0.91 & 6.91 & 8.51\\
& GTNLN & 24.35 & 28.34 & 0.81 & 0.89 & 0.83 & 0.89 & 18.05 & 18.12 \\
& \WSpTFI{} (Ours) & \underline{25.25} & \underline{29.44} & 0.83 & \underline{0.92} & 0.83 & \underline{0.91} & 3.44 & 4.21\\
& \WSpTFII{} (Ours) & \textbf{25.86} & \textbf{30.12} & \underline{0.85} & \textbf{0.94} & \underline{0.85} & \textbf{0.92} & 9.37 & 24.06 \\
\bottomrule
\end{tabular}}
\end{table*}

We next evaluate color-image restoration on the Berkeley Segmentation
Dataset (BSD)\footnote{Dataset source:
\url{https://www.eecs.berkeley.edu/Research/Projects/CS/vision/bsds/}.}.
From the 300 red--green--blue (RGB) images of size $321\times481\times3$, 15 images
are selected at random for evaluation. Each image is represented as a
third-order tensor $\X\in\mathbb{R}^{321\times481\times3}$, with the first
two modes corresponding to spatial dimensions and the third to color
channels. Unlike the synthetic experiment, this setting evaluates completion performance
on natural images containing more complex edges and textures.

Each clean image $\X$ is first corrupted by salt-and-pepper noise with
density $0.05$, producing $\widetilde{\X}$. Missing pixels are then simulated
using a Bernoulli mask,
$\Y=\mathcal{P}_{\OmegaSet}(\widetilde{\X})
=\mathcal{M}\odot\widetilde{\X}$.
Sampling ratios of $30\%$ and $50\%$ are considered. Table~\ref{tab:bsd_color_results} reports the
average PSNR, SSIM, feature similarity index (FSIM), and running time over the 15 test images for both
sampling ratios.

For color-image restoration, \WSpMFI{} uses
$\lambda=\sqrt{\max(n_1,n_2)}$, $p=0.35$, $q=0.8$, and $r=24$;
\WSpMFII{} uses $\lambda=18$, $p=0.85$, $q=0.8$, and $r_0=60$;
\WSpTFI{} uses $\lambda=3\sqrt{\max(n_1,n_2)n_3}$, $p=0.35$, $q=0.8$,
and $r=75$; and \WSpTFII{} uses $\lambda=1$, $p=0.75$,
$q=0.75$, and $r_0=100$.

Qualitative results are presented in
Figs.~\ref{fig:bsd-matrix-restoration-results} and
\ref{fig:bsd-tensor-restoration-results} for the matrix- and tensor-based
methods, respectively. The red and blue boxes identify enlarged regions used
to compare local structure, edge preservation, texture restoration, and residual
corruption.Figure~\ref{fig:bsd-color-imagewise-results} further reports image-wise
PSNR and SSIM values for all 15 test images at sampling ratios of $30\%$ and
$50\%$.

We also examine sensitivity to the exponent parameters $(p,q)$ for
\WSpMFI{}, \WSpMFII{}, \WSpTFI{}, and \WSpTFII{}. For \WSpMFI{} and
\WSpTFI{}, only pairs satisfying
$q\geq 1-\frac{4p^2}{1+4p}$ are considered; for \WSpMFII{} and
\WSpTFII{}, the admissible region is $1\leq p+q\leq2$. For each admissible
pair, all other parameters, the observation mask, and the noise realization
are held fixed, and reconstruction quality is measured by PSNR.
Figure~\ref{fig:color-parameter-selection} visualizes the results as
three-dimensional bar plots with $p$ and $q$ on the horizontal axes and PSNR
on the vertical axis. Broad high-PSNR regions indicate lower sensitivity to
the precise exponent choice, whereas narrow peaks indicate greater parameter
dependence.

FSIM is computed from phase congruency and gradient magnitude. For each
spatial location $p\in\mathcal{D}$, the local similarity is
\begin{equation*}
  \begin{aligned}
  S_L(p) &= S_{PC}(p)S_G(p),\\
  S_{PC}(p)
  &=
  \frac{
  2PC_{\X}(p)PC_{\hat{\X}}(p)+T_1
  }
  {
  PC_{\X}^{2}(p)+PC_{\hat{\X}}^{2}(p)+T_1
  },\\
  S_G(p)
  &=
  \frac{
  2G_{\X}(p)G_{\hat{\X}}(p)+T_2
  }
  {
  G_{\X}^{2}(p)+G_{\hat{\X}}^{2}(p)+T_2
  }.
  \end{aligned}
\end{equation*}
Here, $PC_{\X}(p)$ and $PC_{\hat{\X}}(p)$ denote the phase congruency values, $G_{\X}(p)$ and $G_{\hat{\X}}(p)$ denote the gradient magnitudes, and $T_1,T_2$ are positive constants. The overall FSIM is defined as
\begin{equation*}
\begin{aligned}
  \operatorname{FSIM}(\X,\hat{\X})
  &=
  \frac{
  \sum_{p\in\mathcal{D}}S_L(p)PC_m(p)
  }
  {
  \sum_{p\in\mathcal{D}}PC_m(p)
  },
  \\
  PC_m(p)
  &=
  \max\{PC_{\X}(p),PC_{\hat{\X}}(p)\}.
\end{aligned}
\end{equation*}
where $\mathcal{D}$ is the set of spatial locations.

Both matrix- and tensor-based methods are evaluated in this experiment.
Matrix methods operate on matrix representations of the images, whereas
tensor methods retain the multiway RGB structure. As indicated in
Table~\ref{tab:bsd_color_results}, the best and second-best values are
highlighted across all compared methods.

Table~\ref{tab:bsd_color_results} shows that reconstruction quality generally
improves as the sampling ratio increases from $30\%$ to $50\%$.
\WSpTFII{} achieves the highest average PSNR at both sampling ratios
($25.86$ and $30.12$ dB), while \WSpTFI{} gives the second-highest PSNR
within the tensor group. At $50\%$ sampling, \WSpTFII{} also attains the
highest SSIM and FSIM. These results indicate that the proposed tensor
factorizations are effective at preserving spatial--channel structure under
simultaneous missing pixels and impulse corruption.

\begin{figure}[!t]
  \centering
  \includegraphics[width=\columnwidth]{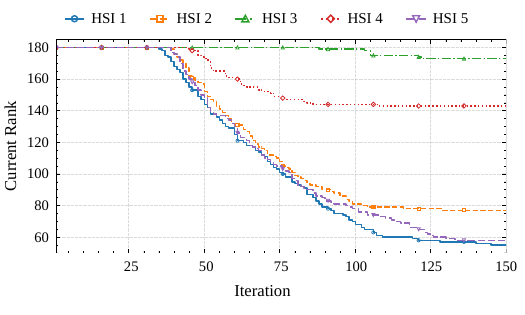}
  \caption{Evolution of the active factor width (current rank) of
  \WSpTFII{} on the five HSI scenes. All runs are initialized with
  $r_0=180$, and the scene-dependent decreases reflect finite pruning of
  low-energy column pairs according to
  \eqref{eq:vwsptf-column-pruning-energy}.}
  \label{fig:hsi_wsptf2_rank_pruning}
\end{figure}

\begin{figure*}[!t]
  \centering
  \includegraphics[width=\textwidth]{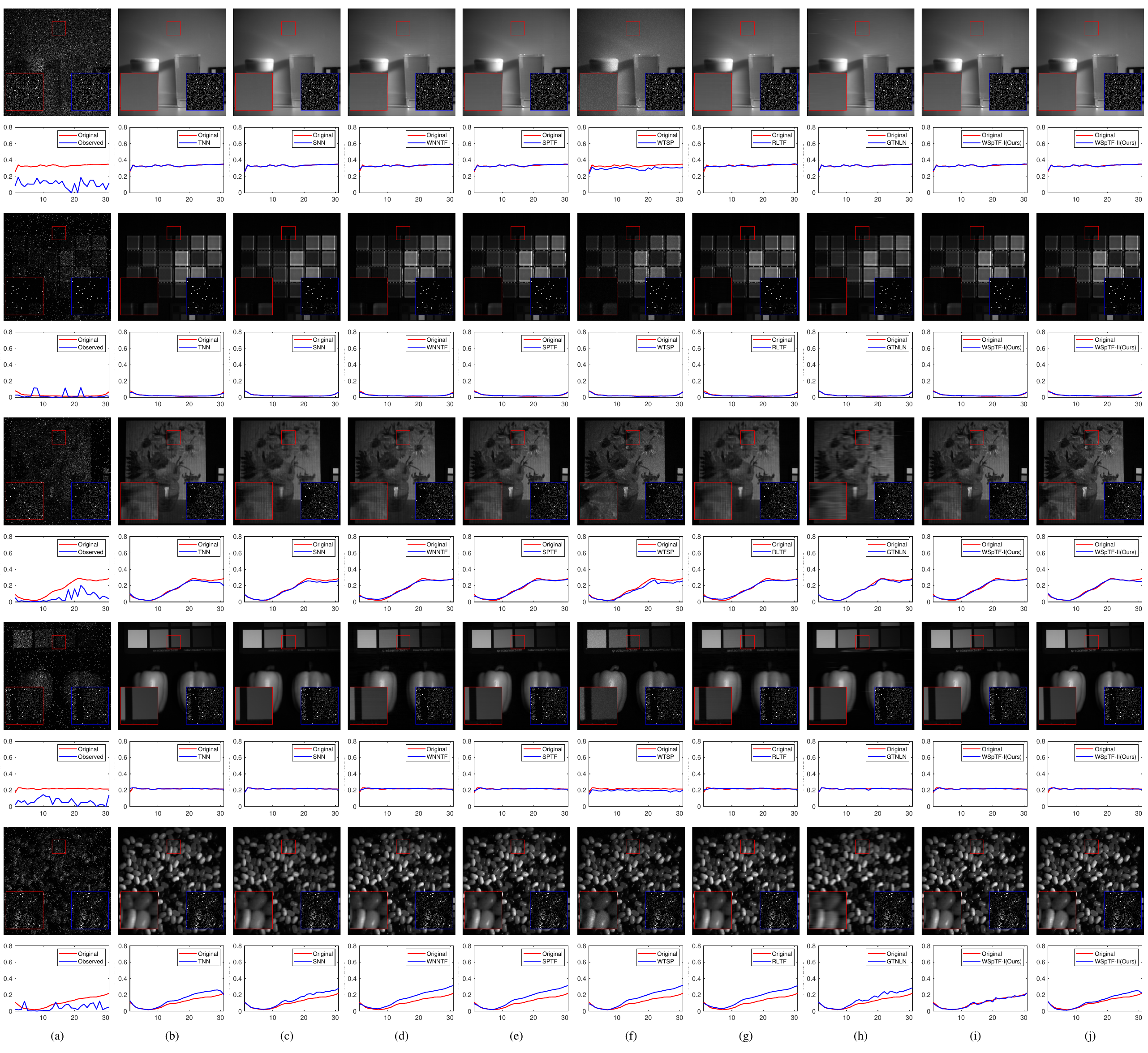}
  \caption{Visual comparison of hyperspectral image inpainting results and
  the corresponding spectral curves on five representative HSI scenes.
  Columns (a)--(j) correspond to Observed, TNN, SNN, WNNTF, SPTF, WTSP,
  RLTF, GTNLN, \WSpTFI{} (Ours), and \WSpTFII{} (Ours), respectively.
  For each scene, the upper row shows the restoration result and the lower
  row shows the corresponding spectral curve. The red and blue boxes indicate
  representative enlarged spatial regions, while the curves compare the
  reconstructed and original spectral responses.}
  \label{fig:hsi_inpainting_visual_comparison}
\end{figure*}

\subsection{Hyperspectral Image Inpainting}
\begin{table*}[!t]
\centering
\caption{Quantitative HSI inpainting results at a sampling ratio of $30\%$ under salt-and-pepper noise densities of $0.05$ and $0.10$. For each scene, metric, and noise density, the best and second-best results are marked in \textbf{bold} and \underline{underlined}, respectively.}
\label{tab:hsi_inpainting_results}
\scriptsize
\renewcommand{\arraystretch}{1.10}
\setlength{\tabcolsep}{1.65pt}
\resizebox{\textwidth}{!}{%
\begin{tabular}{|@{}>{\centering\arraybackslash}m{0.96cm}@{}|c|cc|cc|cc|cc|cc|cc|cc|cc|cc|}
\hline
\multirow{2}{*}{HSI} & \multirow{2}{*}{Index} & \multicolumn{2}{c|}{SNN} & \multicolumn{2}{c|}{TNN} & \multicolumn{2}{c|}{RLTF} & \multicolumn{2}{c|}{WNNTF} & \multicolumn{2}{c|}{SPTF} & \multicolumn{2}{c|}{WTSP} & \multicolumn{2}{c|}{GTNLN} & \multicolumn{2}{c|}{\WSpTFI{}} & \multicolumn{2}{c|}{\WSpTFII{}} \\
\cline{3-20}
& & $0.05$ & $0.10$ & $0.05$ & $0.10$ & $0.05$ & $0.10$ & $0.05$ & $0.10$ & $0.05$ & $0.10$ & $0.05$ & $0.10$ & $0.05$ & $0.10$ & $0.05$ & $0.10$ & $0.05$ & $0.10$ \\
\hline
\multirow{3}{0.96cm}{\centering\includegraphics[width=0.92cm,height=0.92cm,keepaspectratio]{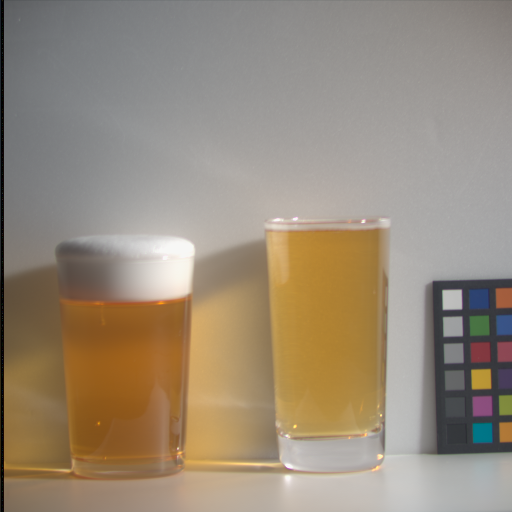}} & PSNR & 35.30 & 34.85 & 34.71 & 34.34 & 35.77 & 35.32 & 37.42 & \underline{36.81} & 37.38 & 36.78 & 31.75 & 31.61 & 35.86 & 34.64 & \textbf{39.17} & \textbf{38.52} & \underline{38.17} & 36.06 \\
 & SSIM & \textbf{0.9783} & \textbf{0.9800} & 0.9644 & 0.9600 & 0.9334 & 0.9300 & 0.9562 & 0.9500 & 0.9558 & 0.9500 & 0.9442 & 0.8700 & 0.9602 & 0.9304 & 0.9695 & \underline{0.9685} & \underline{0.9707} & 0.9600 \\
 & FSIM & 0.9661 & 0.9600 & \underline{0.9673} & \textbf{0.9700} & 0.9584 & 0.9600 & 0.9649 & 0.9600 & 0.9648 & 0.9600 & 0.9441 & 0.9000 & 0.9419 & 0.9215 & \textbf{0.9696} & \underline{0.9689} & 0.9670 & 0.9600 \\
\hline
\multirow{3}{0.96cm}{\centering\includegraphics[width=0.92cm,height=0.92cm,keepaspectratio]{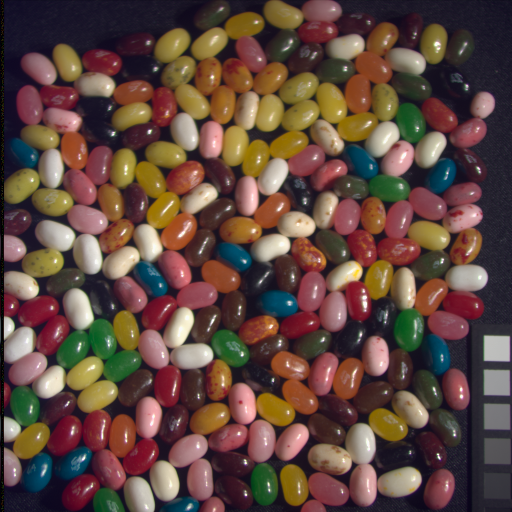}} & PSNR & 28.93 & 28.56 & 31.16 & 30.74 & 34.63 & 33.81 & \underline{35.47} & \underline{34.66} & 35.45 & 34.65 & 29.32 & 29.09 & 26.87 & 26.54 & \textbf{36.91} & \textbf{36.30} & 35.17 & 32.96 \\
 & SSIM & 0.9534 & 0.9500 & 0.9568 & 0.9500 & 0.9517 & 0.9400 & 0.9610 & \underline{0.9600} & 0.9608 & \underline{0.9600} & 0.9381 & 0.9200 & 0.9019 & 0.8703 & \underline{0.9687} & \textbf{0.9669} & \textbf{0.9690} & 0.9400 \\
 & FSIM & 0.9360 & 0.9300 & 0.9540 & 0.9500 & 0.9559 & 0.9500 & 0.9621 & \underline{0.9600} & 0.9620 & \underline{0.9600} & 0.9270 & 0.9200 & 0.8782 & 0.8618 & \textbf{0.9695} & \textbf{0.9677} & \underline{0.9690} & \underline{0.9600} \\
\hline
\multirow{3}{0.96cm}{\centering\includegraphics[width=0.92cm,height=0.92cm,keepaspectratio]{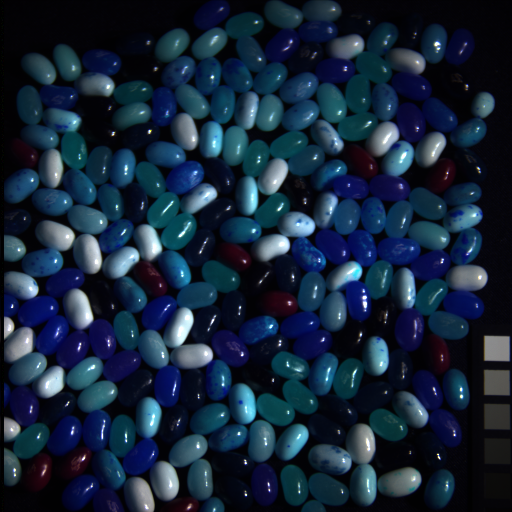}} & PSNR & 24.69 & 24.05 & 26.44 & 26.07 & 29.19 & 28.62 & 29.78 & 29.32 & 29.77 & 29.31 & 29.67 & \underline{29.44} & 25.01 & 24.54 & \underline{30.78} & \textbf{30.59} & \textbf{30.85} & 29.27 \\
 & SSIM & 0.8872 & 0.8600 & 0.8857 & 0.8800 & 0.8855 & 0.8700 & 0.9035 & 0.8900 & 0.9032 & 0.8900 & \textbf{0.9417} & \textbf{0.9300} & 0.8399 & 0.7961 & \underline{0.9207} & \underline{0.9174} & 0.9168 & 0.8900 \\
 & FSIM & 0.8853 & 0.8700 & 0.9096 & 0.9000 & 0.9164 & 0.9100 & 0.9256 & 0.9200 & 0.9254 & 0.9200 & \underline{0.9439} & \underline{0.9300} & 0.8395 & 0.8221 & 0.9340 & \textbf{0.9319} & \textbf{0.9461} & \underline{0.9300} \\
\hline
\multirow{3}{0.96cm}{\centering\includegraphics[width=0.92cm,height=0.92cm,keepaspectratio]{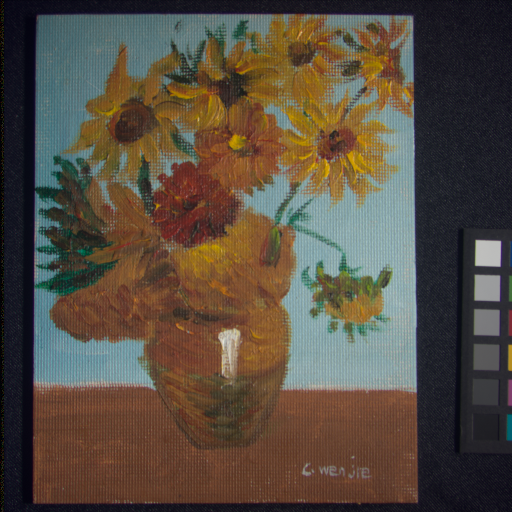}} & PSNR & 30.75 & 30.26 & 31.65 & 31.30 & 32.18 & 31.60 & 33.08 & 32.59 & 33.06 & 32.58 & 30.97 & 30.80 & 28.81 & 28.31 & \underline{34.20} & \textbf{33.91} & \textbf{35.79} & \underline{32.82} \\
 & SSIM & 0.9114 & 0.9000 & 0.9140 & \underline{0.9100} & 0.8952 & 0.8900 & 0.9142 & \underline{0.9100} & 0.9140 & \underline{0.9100} & 0.8710 & 0.8500 & 0.7867 & 0.7522 & \underline{0.9278} & \textbf{0.9250} & \textbf{0.9451} & 0.9000 \\
 & FSIM & 0.9154 & 0.9100 & 0.9259 & 0.9200 & 0.9170 & 0.9100 & 0.9297 & \underline{0.9300} & 0.9295 & \underline{0.9300} & 0.9066 & 0.9000 & 0.8293 & 0.8206 & \underline{0.9397} & \textbf{0.9376} & \textbf{0.9575} & \underline{0.9300} \\
\hline
\multirow{3}{0.96cm}{\centering\includegraphics[width=0.92cm,height=0.92cm,keepaspectratio]{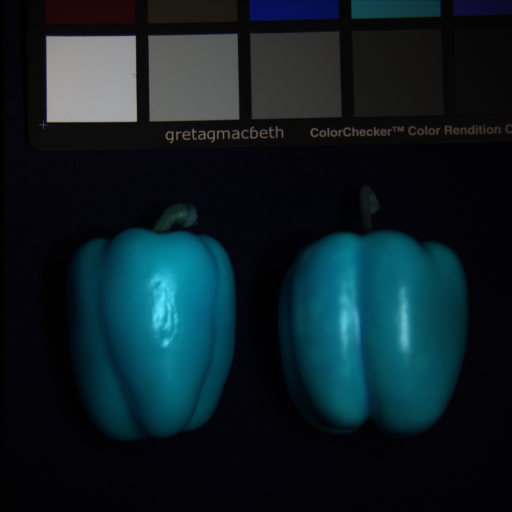}} & PSNR & 36.94 & 36.47 & 34.73 & 34.32 & 36.93 & 36.25 & 37.86 & \underline{37.54} & 37.80 & 37.51 & 36.36 & 36.13 & 37.22 & 35.64 & \textbf{41.71} & \textbf{41.41} & \underline{40.24} & 35.27 \\
 & SSIM & \textbf{0.9778} & \textbf{0.9800} & 0.9550 & 0.9500 & 0.9327 & 0.9300 & 0.9421 & 0.9400 & 0.9413 & 0.9400 & 0.9723 & 0.9500 & 0.9648 & 0.9311 & \underline{0.9752} & \underline{0.9749} & 0.9739 & 0.9400 \\
 & FSIM & 0.9569 & 0.9500 & 0.9635 & \underline{0.9600} & 0.9361 & 0.9300 & 0.9371 & 0.9400 & 0.9367 & 0.9400 & 0.9509 & 0.9300 & 0.9420 & 0.9175 & \underline{0.9662} & \textbf{0.9661} & \textbf{0.9701} & \underline{0.9600} \\
\hline
\multirow{3}{*}{Average} & PSNR & 31.32 & 30.84 & 31.74 & 31.35 & 33.74 & 33.12 & 34.72 & \underline{34.18} & 34.69 & 34.17 & 31.61 & 31.41 & 30.75 & 29.93 & \textbf{36.55} & \textbf{36.15} & \underline{36.04} & 33.28 \\
 & SSIM & 0.9416 & \underline{0.9340} & 0.9352 & 0.9300 & 0.9197 & 0.9120 & 0.9354 & 0.9300 & 0.9350 & 0.9300 & 0.9335 & 0.9040 & 0.8907 & 0.8560 & \underline{0.9524} & \textbf{0.9505} & \textbf{0.9551} & 0.9260 \\
 & FSIM & 0.9320 & 0.9240 & 0.9441 & 0.9400 & 0.9368 & 0.9320 & 0.9439 & 0.9420 & 0.9437 & 0.9420 & 0.9345 & 0.9160 & 0.8862 & 0.8687 & \underline{0.9558} & \textbf{0.9544} & \textbf{0.9619} & \underline{0.9480} \\
\hline
\end{tabular}}
\end{table*}

We evaluate hyperspectral image (HSI) inpainting on the CAVE multispectral
image dataset\footnote{Dataset source:
\url{http://www1.cs.columbia.edu/CAVE/databases/multispectral/}.}.
Each data cube has size $512\times512\times31$ and contains 31 spectral
bands. Five scenes are selected for evaluation. For each scene, the spectral-band images are
ordered by band index and stacked into
$\X_{\mathrm{raw}}\in\mathbb{R}^{512\times512\times31}$ without spatial
cropping. Each cube is globally normalized to $[0,255]$ according to
$\X=255\big(\X_{\mathrm{raw}}-\min(\X_{\mathrm{raw}})\big)/
\big(\max(\X_{\mathrm{raw}})-\min(\X_{\mathrm{raw}})\big)$,
which preserves relative intensity variations across spectral bands.

The degradation combines missing entries with impulse noise.
Salt-and-pepper corruption is applied at densities $\nu=0.05$ and
$\nu=0.10$. A binary observation mask
$\mathcal{M}\in\{0,1\}^{n_1\times n_2\times n_3}$ is then generated
independently with $\mathcal{M}_{ijk}\sim\operatorname{Bernoulli}(\rho)$,
where $\rho$ denotes the sampling ratio and is set to $0.3$ and $0.5$ .
For noise density $\nu$, the observed tensor is
$\Y=\mathcal{M}\odot\operatorname{SP}_{\nu}(\X)$, where
$\operatorname{SP}_{\nu}(\cdot)$ denotes salt-and-pepper corruption.
All methods use identical masks and corrupted observations under each
experimental condition.

Only tensor-based methods are compared in this experiment. Each method is
applied directly to the full HSI cube to preserve joint spatial--spectral
structure.

For HSI inpainting, the parameters of \WSpTFI{} are set to
$p=0.25$, $q=0.9$, and $r=120$, with
$\lambda=80\sqrt{\max(n_1,n_2)n_3}$. For \WSpTFII{}, we set
$p=0.85$, $q=0.2$, and the initial factor width to $r_0=180$, with
$\lambda=\sqrt{\max(n_1,n_2)n_3}$. The deliberately overestimated
initial width is used to further examine the column-pruning behavior
of \WSpTFII{} rather than imposing a fixed target rank.

Figure~\ref{fig:hsi_wsptf2_rank_pruning} plots the evolution of the active
factor width over the iterations for the five HSI scenes. Although all runs
start from $r_0=180$, the active width is subsequently reduced in a
scene-dependent manner by the finite column-pruning rule in
\eqref{eq:vwsptf-column-pruning-energy}. In particular, several scenes exhibit
substantial removal of redundant column pairs, whereas others retain a larger
fraction of the initialized components. This behavior indicates that the
pruning mechanism does not force a common rank across different HSI cubes;
instead, it adapts the retained factor width to the learned paired-column
energies, thereby reducing the sensitivity of \WSpTFII{} to an overestimated
initial rank.

Reconstruction quality is evaluated by PSNR, SSIM, and FSIM. SSIM and FSIM
are computed independently for each spectral band and then averaged across
bands. Quantitative results are reported for each noise density and averaged
over the five scenes, with scene-level values retained to assess consistency
across different spatial and spectral structures. For qualitative
evaluation, false-color images are formed from three representative spectral
bands. Figure~\ref{fig:hsi_inpainting_visual_comparison} compares the
reconstructed images and spectral curves, emphasizing impulse-noise removal,
completion of missing spatial details, and preservation of spectral
information.

\begin{table*}[!t]
\centering
\caption{F-measure comparison for PCB defect detection. For each PCB group and defect category, the best and second-best results are marked in \textbf{bold} and \underline{underlined}, respectively.}
\label{tab:pcb_defect_detection_results}
\small
\renewcommand{\arraystretch}{1.18}
\setlength{\tabcolsep}{5.0pt}
\begin{tabular}{@{}clccccccc@{}}
\toprule
\multirow{2}{*}{\textbf{PCB}} &
\multirow{2}{*}{\textbf{Method}} &
\makecell{\textbf{Missing}\\\textbf{hole}} &
\makecell{\textbf{Mouse}\\\textbf{bite}} &
\makecell{\textbf{Open}\\\textbf{circuit}} &
\makecell{\textbf{Short}\\\textbf{circuit}} &
\textbf{Spur} &
\makecell{\textbf{Spurious}\\\textbf{copper}} &
\textbf{Average} \\
\midrule
\multirow{9}{*}{PCB1} & \WSpTFI{} (Ours) & \underline{0.9705} & \textbf{0.8652} & 0.8298 & 0.8583 & \underline{0.7838} & \textbf{0.9603} & \underline{0.8780} \\
 & \WSpTFII{} (Ours) & \textbf{0.9750} & \underline{0.8641} & \textbf{0.9041} & \textbf{0.9135} & \textbf{0.8092} & 0.9389 & \textbf{0.9008} \\
 & WNNTF & 0.9690 & 0.8090 & 0.8625 & \underline{0.8850} & 0.6986 & 0.9487 & 0.8621 \\
 & SPTF & 0.9703 & 0.8467 & 0.8351 & 0.8583 & 0.7500 & \underline{0.9594} & 0.8700 \\
 & SNN & 0.7656 & 0.7340 & 0.7042 & 0.6843 & 0.6115 & 0.7233 & 0.7038 \\
 & TNN & 0.8928 & 0.7012 & \underline{0.8842} & 0.8212 & 0.5691 & 0.7535 & 0.7703 \\
 & RLTF & 0.9703 & 0.8467 & 0.8298 & 0.8583 & 0.7500 & \underline{0.9594} & 0.8691 \\
 & WTSP & 0.9608 & 0.8019 & 0.7996 & 0.8221 & 0.6986 & 0.8712 & 0.8257 \\
 & GTNLN & 0.5337 & 0.6528 & 0.5619 & 0.5574 & 0.5765 & 0.3034 & 0.5310 \\
\midrule
\multirow{9}{*}{PCB2} & \WSpTFI{} (Ours) & \underline{0.8324} & \textbf{0.9951} & \textbf{0.9971} & \textbf{0.9892} & 0.8336 & \textbf{0.9904} & \textbf{0.9396} \\
 & \WSpTFII{} (Ours) & \textbf{0.9390} & 0.9449 & 0.9104 & 0.9210 & \textbf{0.9430} & 0.9455 & \underline{0.9340} \\
 & WNNTF & 0.7792 & 0.9313 & 0.8924 & 0.9133 & 0.8009 & 0.9497 & 0.8778 \\
 & SPTF & 0.7357 & 0.8824 & 0.8292 & 0.8586 & 0.7645 & 0.9184 & 0.8315 \\
 & SNN & 0.7902 & 0.7629 & 0.7945 & 0.7480 & 0.6022 & 0.5808 & 0.7131 \\
 & TNN & 0.7823 & 0.9294 & 0.8976 & 0.9133 & 0.8026 & 0.9497 & 0.8792 \\
 & RLTF & 0.7823 & 0.9146 & 0.9046 & 0.9116 & 0.8056 & 0.9376 & 0.8761 \\
 & WTSP & 0.6864 & \underline{0.9515} & \underline{0.9383} & \underline{0.9790} & \underline{0.8466} & \underline{0.9699} & 0.8953 \\
 & GTNLN & 0.6316 & 0.5033 & 0.5803 & 0.4872 & 0.2405 & 0.2984 & 0.4569 \\
\bottomrule
\end{tabular}
\end{table*}

Table~\ref{tab:hsi_inpainting_results} shows complementary behavior of the two proposed tensor models under the tested impulse-noise levels. At $\nu=0.05$, \WSpTFI{} achieves the highest average PSNR of $36.55$ dB, while \WSpTFII{} attains the highest average SSIM and FSIM of $0.9551$ and $0.9619$, respectively; its average PSNR is $36.04$ dB, which is the second-best result. At $\nu=0.10$, \WSpTFI{} gives the best average PSNR, SSIM, and FSIM, with values of $36.15$ dB, $0.9505$, and $0.9544$, respectively. These results indicate that the two formulations provide complementary reconstruction advantages across the tested HSI scenes and degradation settings.
\begin{figure*}[!t]
  \centering
  \includegraphics[width=\textwidth]{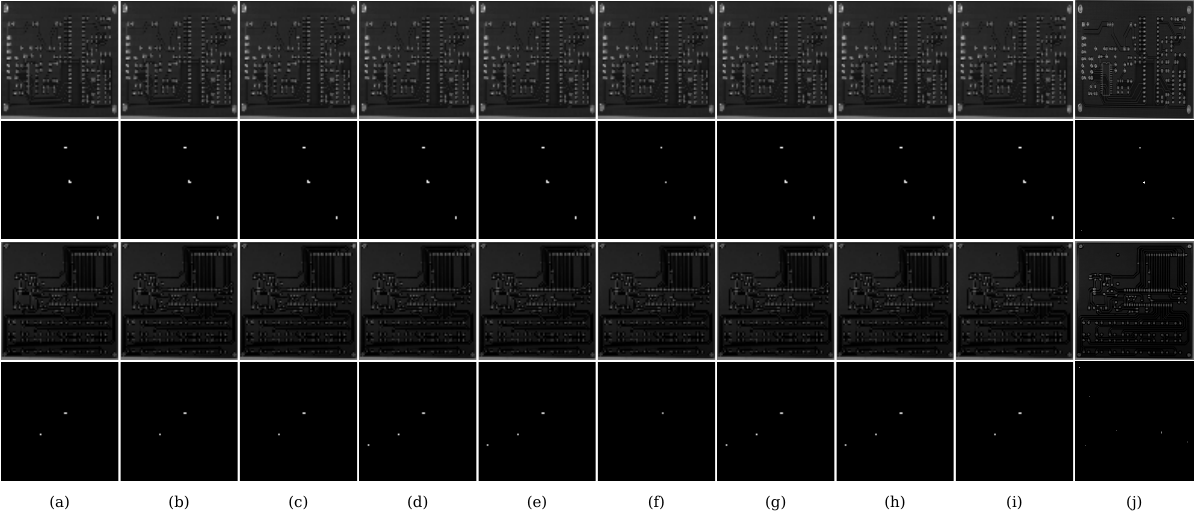}
  \caption{Visual comparison results of different methods on PCB defect
  detection. The first two rows correspond to the first frontal slice of
  PCB1, and the last two rows correspond to the tenth frontal slice of
  PCB2. The first and third rows show the original defective image in
  column (a) and the recovered backgrounds in columns (b)--(j); the
  second and fourth rows show the corresponding ground-truth mask in
  column (a) and predicted defect masks in columns (b)--(j).
  (a) Original/ground truth; (b) \WSpTFI{}; (c) WNNTF; (d) SPTF;
  (e) \WSpTFII{}; (f) SNN; (g) TNN; (h) RLTF; (i) WTSP; (j) GTNLN.}
  \label{fig:pcb_defect_detection_results}
\end{figure*}

\subsection{PCB Defect Detection}

Finally, we evaluate the proposed tensor factorization models on PCB defect detection using the PCB database\footnote{Dataset
source: \url{https://robotics.pkusz.edu.cn/resources/dataset/}.}. The test
images are organized into two groups, PCB1 and PCB2. Each group contains six
defect categories---missing hole, mouse bite, open circuit, short circuit,
spur, and spurious copper---with five defective images per category.

After grayscale conversion, the 30 defective images in each group are
stacked along the third mode to form
$\Y^{(s)}\in\mathbb{R}^{256\times256\times30}$, $s\in\{1,2\}$, with each
frontal slice corresponding to one PCB image. A random intensity offset in
$[-0.2,0.2]$ is added to each defective image to emulate illumination
variation.

Each method decomposes the defective-image tensor into a low-rank
background and a sparse defect component. Let $\widehat{\X}^{(s)}$ denote
the recovered background. The defect-response tensor is defined by the
absolute residual
$\mathcal{S}^{(s)}=\left|\Y^{(s)}-\widehat{\X}^{(s)}\right|$.
For consistent evaluation, both ground-truth and predicted masks are obtained
using the same fixed hard threshold of $5$. For the $k$th image in PCB group
$s$, the ground-truth mask is
$\mathcal{G}^{(s)}_k=\mathbf{1}\!\left(\left|\Y^{(s)}_k-\mathbf{R}^{(s)}\right|>5\right)$,
where $\mathbf{R}^{(s)}$ is the corresponding defect-free reference image
and $\mathbf{1}(\cdot)$ is the indicator function. The predicted mask is
$\widehat{\mathcal{G}}^{(s)}_k=\mathbf{1}\!\left(\mathcal{S}^{(s)}_k>5\right)$.

Only tensor-based methods are compared. All methods use identical PCB
tensors, reference images, masks, threshold values, and evaluation procedures.

For PCB defect detection, the parameters of \WSpTFI{} are set to
$p=0.4$, $q=0.95$, and $r=220$, with
$\lambda=2\sqrt{\max(n_1,n_2)n_3}$. For \WSpTFII{}, we set
$p=0.55$, $q=0.5$, and the initial factor width to $r_0=220$, with
$\lambda=0.25\sqrt{\max(n_1,n_2)n_3}$.

Detection performance is measured by the F-measure between the predicted and
ground-truth masks. Let TP, FP, and FN denote the numbers of true-positive,
false-positive, and false-negative pixels, respectively. Precision and recall
are defined as
\begin{equation*}
\begin{aligned}
  \operatorname{Precision}
  &=
  \frac{\mathrm{TP}}{\mathrm{TP}+\mathrm{FP}},
  \\
  \operatorname{Recall}
  &=
  \frac{\mathrm{TP}}{\mathrm{TP}+\mathrm{FN}},
\end{aligned}
\end{equation*}
and the F-measure is
\begin{equation*}
  F
  =
  \frac{
  2\,\operatorname{Precision}\operatorname{Recall}
  }{
  \operatorname{Precision}+\operatorname{Recall}
  }.
\end{equation*}
Higher F-measure values indicate better detection performance. For each
defect category, the F-measure is averaged over its five images; the group
average is then computed over the six category-level scores.

As reported in Table~\ref{tab:pcb_defect_detection_results}, the two
proposed methods together achieve the highest F-measure in every defect
category across PCB1 and PCB2. On PCB1, \WSpTFII{} gives the highest average
F-measure ($0.9008$) and ranks first for missing hole, open circuit, short
circuit, and spur; \WSpTFI{} ranks first for mouse bite and spurious copper.
On PCB2, \WSpTFI{} gives the highest average F-measure ($0.9396$) and ranks
first for mouse bite, open circuit, short circuit, and spurious copper,
whereas \WSpTFII{} performs best for missing hole and spur. The results show
consistent defect-localization performance across the six defect categories.

For qualitative comparison, the first frontal slice of PCB1 and the tenth
frontal slice of PCB2 are selected as representative examples in
Fig.~\ref{fig:pcb_defect_detection_results}. The figure displays the original
defective images, ground-truth masks, recovered backgrounds, and predicted
defect masks for all methods.

The qualitative results are consistent with the F-measure evaluation.
The proposed methods recover comparatively clean backgrounds while retaining
compact responses at the principal defect locations. Their predicted masks
suppress much of the background structure and preserve the dominant defective
regions, in agreement with the quantitative results in
Table~\ref{tab:pcb_defect_detection_results}.

\section{Conclusion}
\label{sec:conclusion}

This paper developed two complementary exact factorization routes for robust
low-rank tensor completion under the t-product framework. On the common admissible
parameter region, \WSpTFI{} and \WSpTFII{} represent the same weighted
Schatten-$p$ norm-based spectral regularizer up to a positive scaling constant, as made
explicit in Remark~\ref{rem:common-spectral-regularizer}. \WSpTFI{} realizes
this regularizer through iteratively reweighted factor-spectral regularization with flexible,
possibly asymmetric factor exponents, thereby retaining a direct connection
between the spectrum of the reconstructed tensor and the spectra of its
compact factors. \WSpTFII{} provides a second exact route through a
transform-domain paired-column variational representation, leading to
SVD-free main factor updates and an energy-based mechanism for removing
redundant factor components. The two models therefore differ in
factor-space parameterization and computational tradeoffs, not in the target
spectral regularizer.

The theoretical guarantees are aligned with the corresponding algorithms. For
\WSpTFI{}, under the assumptions of Theorem~\ref{thm:wsptf-convergence}, the
iteratively reweighted ADMM-type scheme converges conditionally to KKT points
of the limiting fixed-weight split problem. For \WSpTFII{}, once finite
pruning is inactive and the stated assumptions hold, the damped IRLS--BSUM
updates yield sufficient decrease, asymptotic regularity, vanishing
first-order residuals, and stationary accumulation points for the
fixed-$\delta$ smoothed objective. Experiments on synthetic completion,
color-image restoration, hyperspectral inpainting, and PCB defect detection
consistently demonstrate the robustness of the proposed factorization
strategies under incomplete and corrupted observations.

Several limitations remain. Both models are nonconvex and retain sensitivity
to initialization, factor width, and regularization parameters. The
\WSpTFI{} convergence result is conditional and characterizes a limiting
frozen-weight problem, whereas the \WSpTFII{} stationarity result applies to a
fixed smooth approximation of the robust loss. Future work will therefore
focus on a more direct majorization/stationarity theory for the original
reweighted \WSpTFI{} objective, data-driven rank and parameter selection,
large-scale and parallel implementations.

\appendices
\section{Proofs for the \WSpTFI{} Model}
\label{app:proof-thm-tensor-factorization}
\subsection{Proof of Theorem~\ref{thm:tensor-factorization}}

The proof is carried out slice by slice under the invertible transform fixed in Section~\ref{sec:notations} and satisfying \eqref{eq:transform-condition}. Define
\begin{equation*}
  h(t)=c(t+\varepsilon)^{q-1}t^p,\qquad t\geq0,
\end{equation*}
and let $\psi(x)=h(e^x)$. With $y=e^x>0$,
\begin{equation*}
  \psi'(x)
  =
  c\,y^p(y+\varepsilon)^{q-2}
  \left[p\varepsilon+(p+q-1)y\right],
\end{equation*}
and
\begin{equation}
\begin{aligned}
  \psi''(x)
  =
  c\,y^p(y+\varepsilon)^{q-3}
  \big[
    &p^2\varepsilon^2\\
    &+\{2p(p+q-1)+q-1\}\varepsilon y\\
    &+(p+q-1)^2y^2
  \big].
\end{aligned}
  \label{eq:app-thm1-psi-second}
\end{equation}
The lower bound on $q$ gives
\[
  p+q-1\geq\frac{p}{1+4p}>0,
\]
so $\psi'(x)>0$. The discriminant of the quadratic polynomial
in \eqref{eq:app-thm1-psi-second} is
\begin{equation*}
  \varepsilon^2(q-1)
  \left[4p^2+4pq-4p+q-1\right].
\end{equation*}
Because $q\leq1$ and
\[
  q\geq1-\frac{4p^2}{1+4p}
  \iff
  4p^2+4pq-4p+q-1\geq0,
\]
the discriminant is nonpositive. Hence $\psi''(x)\geq0$, so
$h\circ\exp$ is increasing and convex.

Therefore, the assumptions of
Lemma~\ref{lem:wsptf-product-singular-value} are satisfied by
the function $h$ in the present theorem.

Let $\X=\mathcal U*\mathcal V$ be feasible and put
$d=\min\{n_1,n_2,r\}$. For each transform-domain slice,
$\bar{\X}^{(i)}=\bar{\mathcal U}^{(i)}
\bar{\mathcal V}^{(i)}$. Since
$\operatorname{rank}(\bar{\X}^{(i)})\leq r$,
Lemma~\ref{lem:wsptf-product-singular-value} gives
\begin{equation*}
\begin{aligned}
  &\sum_j w_{ij}^{\X}
  \sigma_j\!\left(\bar{\X}^{(i)}\right)^p\\
  &\leq
  \sum_{j=1}^{d}
  w_{ij}^{\mathcal U,\mathcal V}
  \left[
    \sigma_j\!\left(\bar{\mathcal U}^{(i)}\right)
    \sigma_j\!\left(\bar{\mathcal V}^{(i)}\right)
  \right]^p.
\end{aligned}
\end{equation*}
Since $p/p_1+p/p_2=1$, Young's inequality yields
\begin{equation*}
  \frac{1}{p}(uv)^p
  \leq
  \frac{1}{p_1}u^{p_1}
  +
  \frac{1}{p_2}v^{p_2}.
\end{equation*}
Combining the last two inequalities and adding any nonnegative
unpaired factor terms gives
\begin{equation*}
\begin{aligned}
  &\frac{1}{p}\sum_j w_{ij}^{\X}
  \sigma_j\!\left(\bar{\X}^{(i)}\right)^p\\
  &\leq
  \frac{1}{p_1}\sum_j
  w_{ij}^{\mathcal U,\mathcal V}
  \sigma_j\!\left(\bar{\mathcal U}^{(i)}\right)^{p_1}\\
  &\quad+
  \frac{1}{p_2}\sum_j
  w_{ij}^{\mathcal U,\mathcal V}
  \sigma_j\!\left(\bar{\mathcal V}^{(i)}\right)^{p_2}.
\end{aligned}
\end{equation*}
Summing over $i$ and scaling by $1/\ell$ proves the lower-bound direction.

For attainability, let
$\bar{\X}^{(i)}=\mathbf P_i\boldsymbol\Sigma_i\mathbf Q_i^\top$
be a compact SVD with rank $r_i\leq r$. Choose
$\mathbf H_i\in\R^{r\times r_i}$ with
$\mathbf H_i^\top\mathbf H_i=\mathbf I_{r_i}$ and define
\begin{equation*}
\begin{aligned}
  \bar{\mathcal U}_{\star}^{(i)}
  &=
  \mathbf P_i\boldsymbol\Sigma_i^{p/p_1}\mathbf H_i^\top,\\
  \bar{\mathcal V}_{\star}^{(i)}
  &=
  \mathbf H_i\boldsymbol\Sigma_i^{p/p_2}\mathbf Q_i^\top.
\end{aligned}
\end{equation*}
Then their product is $\bar{\X}^{(i)}$, and
\begin{equation*}
\begin{aligned}
  \sigma_j\!\left(\bar{\mathcal U}_{\star}^{(i)}\right)^{p_1}
  &=
  \sigma_j\!\left(\bar{\X}^{(i)}\right)^p,\\
  \sigma_j\!\left(\bar{\mathcal V}_{\star}^{(i)}\right)^{p_2}
  &=
  \sigma_j\!\left(\bar{\X}^{(i)}\right)^p.
\end{aligned}
\end{equation*}
Moreover,
\begin{equation*}
  \sigma_j\!\left(\bar{\mathcal U}_{\star}^{(i)}\right)
  \sigma_j\!\left(\bar{\mathcal V}_{\star}^{(i)}\right)
  =
  \sigma_j\!\left(\bar{\X}^{(i)}\right),
\end{equation*}
so the tensor and factor weights coincide. Equality therefore
holds in both inequalities. Applying $\mathcal L^{-1}$ gives a
real feasible factor pair attaining
$\frac1p\|\X\|_{W_\X,S_p}^{p}$, which proves the theorem.

\subsection{Proof of Theorem~\ref{thm:wsptf-convergence}}
\label{app:convergence}

We emphasize the role of the assumptions before giving the algebra.
Boundedness of $\mathcal Z_1^k$ and $\mathcal Z_2^k$, together with the
$\X$-subproblem optimality condition, yields bounded multipliers; the
geometric growth of $\mu_k$ makes the resulting increment bounds summable;
exact block minimization supplies the spectral optimality inclusions; and
the two factor-block residual assumptions are used only in the final
factor-stationarity equations. Accordingly, the conclusion is a KKT
statement for the limiting frozen-weight problem
\eqref{eq:wsptf-limiting-fixed-weight}.

The multiplier updates are
\begin{equation*}
\begin{aligned}
  \mathcal Z_1^{k+1}
  &=
  \mathcal Z_1^k+\mu_k
  (\widehat{\mathcal U}^{k+1}-\mathcal U^{k+1}),\\
  \mathcal Z_2^{k+1}
  &=
  \mathcal Z_2^k+\mu_k
  (\widehat{\mathcal V}^{k+1}-\mathcal V^{k+1}),\\
  \mathcal Z_3^{k+1}
  &=
  \mathcal Z_3^k+\mu_k
  (\mathcal U^{k+1}*\mathcal V^{k+1}-\X^{k+1}).
\end{aligned}
\end{equation*}
The $\X$-subproblem optimality condition is
\begin{equation*}
  0\in
  \lambda_2\partial
  \|\mathcal M\odot(\X^{k+1}-\Y)\|_1-\mathcal Z_3^{k+1}.
\end{equation*}
Thus $\mathcal Z_3^k$ vanishes on unobserved entries and is
entrywise bounded by $\lambda_2$ on observed entries. Hence
$\{\mathcal Z_3^k\}$ is bounded.

The previous multiplier updates give
\begin{equation*}
\begin{aligned}
  \widehat{\mathcal U}^{k}
  &=
  \mathcal U^k+\mu_{k-1}^{-1}
  (\mathcal Z_1^k-\mathcal Z_1^{k-1}),\\
  \X^k
  &=
  \mathcal U^k*\mathcal V^k-\mu_{k-1}^{-1}
  (\mathcal Z_3^k-\mathcal Z_3^{k-1}).
\end{aligned}
\end{equation*}
Substitution into the first-order condition of the
$\mathcal U$ least-squares update gives
\begin{equation*}
  \mathcal U^{k+1}-\mathcal U^k
  =
  \mu_k^{-1}\mathcal Q_U^k,
\end{equation*}
where
\begin{equation*}
\begin{aligned}
  \mathcal Q_U^k
  ={}&
  \big[
    \{\rho(\mathcal Z_3^{k-1}-\mathcal Z_3^k)
      -\mathcal Z_3^k\}*(\mathcal V^k)^\top\\
  &\quad+
    \rho(\mathcal Z_1^k-\mathcal Z_1^{k-1})
    +\mathcal Z_1^k
  \big]
  *
  \left(
    \mathcal V^k*(\mathcal V^k)^\top+\mathcal I
  \right)^{-1}.
\end{aligned}
\end{equation*}
To make the boundedness of $\mathcal Q_U^k$ explicit, consider each
frontal slice in the transform domain. For every $i$ and $k$,
\begin{equation*}
\begin{aligned}
  \bar{\mathcal V}^{(i),k}
  \big(\bar{\mathcal V}^{(i),k}\big)^\top+\mathbf I
  &\succeq \mathbf I,\\
  \left\|
  \left[
  \bar{\mathcal V}^{(i),k}
  \big(\bar{\mathcal V}^{(i),k}\big)^\top+\mathbf I
  \right]^{-1}
  \right\|_2
  &\leq 1.
\end{aligned}
\end{equation*}
Hence the inverse tensor operator in $\mathcal Q_U^k$ is uniformly
bounded. Together with boundedness of the primal variables and of
$\mathcal Z_1^k,\mathcal Z_3^k$, this gives a constant
$\vartheta_U<\infty$, independent of $k$, such that
$\|\mathcal Q_U^k\|_F\leq\vartheta_U$.

We now spell out the geometric-series argument. For arbitrary $n>k$,
the telescoping sum, the increment formula, and
$\mu_j=\mu_k\rho^{j-k}$ give
\begin{equation*}
\begin{aligned}
  \|\mathcal U^n-\mathcal U^k\|_F
  &\leq
  \sum_{j=k}^{n-1}
  \|\mathcal U^{j+1}-\mathcal U^j\|_F\\
  &\leq
  \vartheta_U
  \sum_{j=k}^{n-1}\frac{1}{\mu_j}\\
  &=
  \frac{\vartheta_U}{\mu_k}
  \sum_{t=0}^{n-k-1}\rho^{-t}\\
  &\leq
  \frac{\vartheta_U}{\mu_k}
  \sum_{t=0}^{\infty}\rho^{-t}\\
  &=
  \frac{\vartheta_U}{\mu_k}
  \frac{1}{1-\rho^{-1}}
  =
  \frac{\vartheta_U\rho}{\mu_k(\rho-1)}.
\end{aligned}
\end{equation*}
Because $\rho>1$, $\mu_k=\mu_0\rho^k\to\infty$, so the final bound
tends to zero as $k\to\infty$, uniformly for all $n>k$. Equivalently,
for every $\eta>0$ there exists $K$ such that
$\|\mathcal U^n-\mathcal U^k\|_F<\eta$ whenever $n>k\geq K$.
Thus $\{\mathcal U^k\}$ is Cauchy.

Likewise,
\begin{equation*}
  \mathcal V^{k+1}-\mathcal V^k
  =
  \mu_k^{-1}\mathcal Q_V^k,
\end{equation*}
where
\begin{equation*}
\begin{aligned}
  \mathcal Q_V^k
  ={}&
  \left(
    (\mathcal U^{k+1})^\top*\mathcal U^{k+1}
    +\mathcal I
  \right)^{-1}\\
  &*
  \big[
    (\mathcal U^{k+1})^\top*
    \{
      \rho(\mathcal Z_3^{k-1}-\mathcal Z_3^k)
      -\mathcal Z_3^k
      -\mathcal Q_U^k*\mathcal V^k
    \}\\
  &\quad+
    \rho(\mathcal Z_2^k-\mathcal Z_2^{k-1})
    +\mathcal Z_2^k
  \big].
\end{aligned}
\end{equation*}
The same inverse bound applies to
$(\mathcal U^{k+1})^\top*\mathcal U^{k+1}+\mathcal I$: in every
transform-domain frontal slice the corresponding matrix is bounded below
by $\mathbf I$, so its inverse has spectral norm at most one. Since
$\{\mathcal U^k\}$, the multipliers, and $\{\mathcal Q_U^k\}$ are
bounded, there is a constant $\vartheta_V<\infty$ such that
$\|\mathcal Q_V^k\|_F\leq\vartheta_V$. Consequently, for $n>k$,
\begin{equation*}
\begin{aligned}
  \|\mathcal V^n-\mathcal V^k\|_F
  &\leq
  \vartheta_V\sum_{j=k}^{n-1}\frac{1}{\mu_j}\\
  &=
  \frac{\vartheta_V}{\mu_k}
  \sum_{t=0}^{n-k-1}\rho^{-t}\\
  &\leq
  \frac{\vartheta_V\rho}{\mu_k(\rho-1)}
  \longrightarrow0.
\end{aligned}
\end{equation*}
Hence $\{\mathcal V^k\}$ is Cauchy as well.

The multiplier updates now give
\begin{equation}
\begin{aligned}
  \widehat{\mathcal U}^{k+1}-\mathcal U^{k+1}
  &\longrightarrow0,\\
  \widehat{\mathcal V}^{k+1}-\mathcal V^{k+1}
  &\longrightarrow0,\\
  \mathcal U^{k+1}*\mathcal V^{k+1}-\X^{k+1}
  &\longrightarrow0.
\end{aligned}
  \label{eq:app-wsptf-primal-residuals}
\end{equation}
Because finite-dimensional Euclidean spaces are complete, the two
Cauchy sequences have limits
$\mathcal U^k\to\mathcal U^\star$ and
$\mathcal V^k\to\mathcal V^\star$. The first two relations in
\eqref{eq:app-wsptf-primal-residuals} then imply
$\widehat{\mathcal U}^k\to\mathcal U^\star$ and
$\widehat{\mathcal V}^k\to\mathcal V^\star$. Moreover, the transform-induced
tensor product is a continuous bilinear mapping, and hence
\[
  \mathcal U^k*\mathcal V^k
  \longrightarrow
  \mathcal U^\star*\mathcal V^\star.
\]
Combining this with the third residual relation gives
$\X^k\to\mathcal U^\star*\mathcal V^\star$. Thus all five primal
sequences converge, and in particular are Cauchy.

The two weight sequences are generated from
$(\widehat{\mathcal U}^k,\widehat{\mathcal V}^k)$ and
$(\widehat{\mathcal U}^{k+1},\widehat{\mathcal V}^k)$, respectively.
Both input pairs converge to the same factor limit. Since singular
values depend continuously on matrix entries and $\varepsilon>0$
keeps the argument of the weighting function in
\eqref{eq:wsptf-admm-weight}--\eqref{eq:wsptf-half-weight} strictly
positive, the weight mapping is continuous at the limit. Consequently,
\begin{equation}
  W^k\longrightarrow W^\star,
  \qquad
  W^{k+1/2}\longrightarrow W^\star.
  \label{eq:app-wsptf-weight-convergence}
\end{equation}

The primal sequence is convergent, while
$\mathcal Z_1^k,\mathcal Z_2^k$ are bounded by assumption and
$\mathcal Z_3^k$ was proved bounded above. Hence the full primal-dual
sequence is bounded and therefore has accumulation points. Let
$\mathcal S^{k_j}\to\mathcal S^\star$ be any such accumulation
subsequence. The primal limit is unique, and
\eqref{eq:app-wsptf-primal-residuals} gives
\begin{equation}
  \widehat{\mathcal U}^{\star}=\mathcal U^{\star},
  \quad
  \widehat{\mathcal V}^{\star}=\mathcal V^{\star},
  \quad
  \mathcal U^{\star}*\mathcal V^{\star}=\X^{\star}.
  \label{eq:app-feasible}
\end{equation}
The exact spectral updates satisfy
\begin{equation*}
  0\in
  \lambda_1\partial
  \left(
    \frac{1}{p_1}
    \|\widehat{\mathcal U}^{k+1}\|_{W^k,S_{p_1}}^{p_1}
  \right)+\mathcal Z_1^{k+1},
\end{equation*}
and
\begin{equation*}
  0\in
  \lambda_1\partial
  \left(
    \frac{1}{p_2}
    \|\widehat{\mathcal V}^{k+1}\|_{W^{k+1/2},S_{p_2}}^{p_2}
  \right)+\mathcal Z_2^{k+1}.
\end{equation*}
A small index shift is needed before taking limits. The displayed
optimality conditions are written at iteration $k+1$, whereas the
chosen accumulation subsequence is indexed by $k_j$. For all
sufficiently large $j$, set $k=k_j-1$. Then the three relevant
conditions are
\begin{equation*}
  0\in
  \lambda_1\partial
  \left(
    \frac{1}{p_1}
    \|\widehat{\mathcal U}^{k_j}\|_{W^{k_j-1},S_{p_1}}^{p_1}
  \right)+\mathcal Z_1^{k_j},
\end{equation*}
\begin{equation*}
  0\in
  \lambda_1\partial
  \left(
    \frac{1}{p_2}
    \|\widehat{\mathcal V}^{k_j}\|_{W^{k_j-\frac12},S_{p_2}}^{p_2}
  \right)+\mathcal Z_2^{k_j},
\end{equation*}
and
\begin{equation*}
  0\in
  \lambda_2\partial
  \|\mathcal M\odot(\X^{k_j}-\Y)\|_1-\mathcal Z_3^{k_j}.
\end{equation*}
For completeness, we also make explicit the closed-graph step for the
spectral inclusions. Define
\[
\begin{aligned}
  \Psi_U(\widehat{\mathcal U},W)
  &:=
  \frac{1}{p_1}
  \|\widehat{\mathcal U}\|_{W,S_{p_1}}^{p_1},\\
  \Psi_V(\widehat{\mathcal V},W)
  &:=
  \frac{1}{p_2}
  \|\widehat{\mathcal V}\|_{W,S_{p_2}}^{p_2}.
\end{aligned}
\]
For positive finite weights these finite-dimensional weighted spectral
functions are jointly continuous in their tensor argument and in $W$,
and the corresponding partial limiting-subdifferential mappings
$(\widehat{\mathcal U},W)\mapsto
\partial_{\widehat{\mathcal U}}\Psi_U(\widehat{\mathcal U},W)$ and
$(\widehat{\mathcal V},W)\mapsto
\partial_{\widehat{\mathcal V}}\Psi_V(\widehat{\mathcal V},W)$ have
closed graphs. The required function-value convergence follows from
this joint continuity. Therefore, using
$\widehat{\mathcal U}^{k_j}\to\widehat{\mathcal U}^\star$,
$\widehat{\mathcal V}^{k_j}\to\widehat{\mathcal V}^\star$,
$\mathcal Z_m^{k_j}\to\mathcal Z_m^\star$, and
\eqref{eq:app-wsptf-weight-convergence}, we may pass to the limit in
the shifted inclusions and obtain
\begin{equation*}
  0\in
  \lambda_1\partial
  \left(
    \frac{1}{p_1}
    \|\widehat{\mathcal U}^{\star}\|_{W^\star,S_{p_1}}^{p_1}
  \right)+\mathcal Z_1^\star,
\end{equation*}
\begin{equation*}
  0\in
  \lambda_1\partial
  \left(
    \frac{1}{p_2}
    \|\widehat{\mathcal V}^{\star}\|_{W^\star,S_{p_2}}^{p_2}
  \right)+\mathcal Z_2^\star.
\end{equation*}
The $\ell_1$ subdifferential also has a closed graph; hence the shifted
$\X$-optimality condition gives
\begin{equation*}
  0\in
  \lambda_2\partial
  \|\mathcal M\odot(\X^\star-\Y)\|_1-\mathcal Z_3^\star,
  \qquad
  (\mathbf1-\mathcal M)\odot\mathcal Z_3^\star=0.
\end{equation*}
Finally, the assumed factor-block residual convergence gives
\begin{equation*}
  -\mathcal Z_1^\star+
  \mathcal Z_3^\star*(\mathcal V^\star)^\top=0,
\end{equation*}
and
\begin{equation*}
  -\mathcal Z_2^\star+
  (\mathcal U^\star)^\top*\mathcal Z_3^\star=0.
\end{equation*}
Together with \eqref{eq:app-feasible}, these relations form the
complete KKT system of the limiting fixed-weight problem
\eqref{eq:wsptf-limiting-fixed-weight}.


\section{Proofs for the \WSpTFII{} Model}
\label{app:vwsptf-proof}

\subsection{Proof of Theorem~\ref{thm:vwsptf-variational}}

Let $s=\min\{n_1,n_2\}$. We first verify the scalar properties
needed below. For $t>0$,
\begin{equation*}
  \phi'(t)
  =
  t^{p-1}(t+\epsilon)^{q-2}
  \left[(p+q-1)t+p\epsilon\right]
  >0,
\end{equation*}
and
\begin{equation*}
\begin{aligned}
  \phi''(t)
  =
  &\;
  t^{p-2}(t+\epsilon)^{q-3}
  \big[
    p(p-1)\epsilon^2
    +
    2p(p+q-2)\epsilon t
    \\
    &\qquad\qquad
    +
    (p+q-1)(p+q-2)t^2
  \big].
\end{aligned}
\end{equation*}
Because $0<p\leq1$, $0<q\leq1$, and
$1\leq p+q\leq2$, every coefficient in the square brackets is
nonpositive. Thus $\phi$ is nondecreasing and concave on
$(0,\infty)$. In addition, $\phi(0)=0$ and $\phi$ is continuous at
zero, so these properties extend to the whole interval
$[0,\infty)$.

Define
\[
  \Psi(\mathbf{z})
  =
  \sum_{j=1}^{2s}\phi(z_j),
  \qquad
  \mathbf{z}\in\R_+^{2s}.
\]
The function $\Psi$ is permutation invariant, coordinatewise
nondecreasing, and concave. Hence
Lemma~\ref{lem:ips-concave-singular-value} applies to each
transform-domain frontal slice.

Consider any feasible factorization with the dimensions specified
in Theorem~\ref{thm:vwsptf-variational}. For each $i$,
\begin{equation*}
  \bar{\X}^{(i)}
  =
  \bar{\mathcal{U}}^{(i)}
  \bar{\mathcal{V}}^{(i)}
  =
  \sum_{j=1}^{r}
  \bar{\mathbf{u}}_{j}^{(i)}
  \left(\bar{\mathbf{v}}_{j}^{(i)}\right)^\top .
\end{equation*}
Applying Lemma~\ref{lem:ips-concave-singular-value} successively to
this sum, using $\phi(0)=0$ and the fact that a rank-one matrix has
the single nonzero singular value
$\|\bar{\mathbf{u}}_{j}^{(i)}\|_2
 \|\bar{\mathbf{v}}_{j}^{(i)}\|_2$, yields
\begin{equation*}
  \sum_{j=1}^{s}
  \phi\!\left(\sigma_j(\bar{\X}^{(i)})\right)
  \leq
  \sum_{j=1}^{r}
  \phi\!\left(
    \|\bar{\mathbf{u}}_{j}^{(i)}\|_2
    \|\bar{\mathbf{v}}_{j}^{(i)}\|_2
  \right).
\end{equation*}
The arithmetic--geometric mean inequality gives
\[
  \|\bar{\mathbf{u}}_{j}^{(i)}\|_2
  \|\bar{\mathbf{v}}_{j}^{(i)}\|_2
  \leq
  \frac{
    \|\bar{\mathbf{u}}_{j}^{(i)}\|_2^2+
    \|\bar{\mathbf{v}}_{j}^{(i)}\|_2^2
  }{2}
  =
  t_j^{(i)}.
\]
Since $\phi$ is nondecreasing,
\begin{equation*}
  \sum_{j=1}^{s}
  \phi\!\left(\sigma_j(\bar{\X}^{(i)})\right)
  \leq
  \sum_{j=1}^{r}\phi(t_j^{(i)}).
\end{equation*}
Summing over $i$ and scaling by $1/\ell$ proves that every feasible factorization
satisfies
\begin{equation}
  \rho_{p,q,\epsilon}(\X)
  \leq
  R_{p,q,\epsilon}(\mathcal{U},\mathcal{V}).
  \label{eq:app-vwsptf-any-factor}
\end{equation}

It remains to show attainability. Let
$r_i=\operatorname{rank}(\bar{\X}^{(i)})$ and take a compact SVD
\[
  \bar{\X}^{(i)}
  =
  \sum_{j=1}^{r_i}
  \sigma_j(\bar{\X}^{(i)})
  \mathbf{a}_{j}^{(i)}
  \left(\mathbf{b}_{j}^{(i)}\right)^\top.
\]
Because $r\geq\max_i r_i$, define, for
$j=1,\ldots,r_i$,
\begin{equation*}
  \bar{\mathbf{u}}_{j}^{(i)}
  =
  \sqrt{\sigma_j(\bar{\X}^{(i)})}\,
  \mathbf{a}_{j}^{(i)},
  \qquad
  \bar{\mathbf{v}}_{j}^{(i)}
  =
  \sqrt{\sigma_j(\bar{\X}^{(i)})}\,
  \mathbf{b}_{j}^{(i)},
\end{equation*}
and set the remaining column pairs to zero. Since $\mathcal L$ is real
and invertible, these real transform-domain slices define real
tensors $\mathcal{U}$ and $\mathcal{V}$ through $\mathcal L^{-1}$.
They satisfy $\X=\mathcal{U}*\mathcal{V}$ and
\[
  t_j^{(i)}
  =
  \sigma_j(\bar{\X}^{(i)})
  \quad
  (j=1,\ldots,r_i).
\]
Consequently,
\[
  R_{p,q,\epsilon}(\mathcal{U},\mathcal{V})
  =
  \rho_{p,q,\epsilon}(\X).
\]
Together with \eqref{eq:app-vwsptf-any-factor}, this proves both
the equality and attainment of the minimum in
\eqref{eq:vwsptf-variational-equivalence}.

\subsection{Proof of Theorem~\ref{thm:vwsptf-monotonic}}

Fix $k\geq k_0$. Let
\begin{equation*}
  r_{abc}
  =
  (\mathcal{U}*\mathcal{V})_{abc}
  -
  \Y_{abc}.
\end{equation*}
For the smoothed absolute-value function $h(r)=\sqrt{r^2+\delta^2}$, define $g(s)=\sqrt{s+\delta^2}$ with $s=r^2$. Since $g(s)$ is concave on $s\geq0$, we have
\begin{equation*}
  g(s)
  \leq
  g(s^k)
  +
  g'(s^k)(s-s^k).
\end{equation*}
Substituting $s=r^2$ gives
\begin{equation}
  \sqrt{r^2+\delta^2}
  \leq
  \sqrt{(r^k)^2+\delta^2}
  +
  \frac{r^2-(r^k)^2}
  {2\sqrt{(r^k)^2+\delta^2}}.
  \label{eq:app-vwsptf-irls-majorization}
\end{equation}
Rearranging the right-hand side of
\eqref{eq:app-vwsptf-irls-majorization} gives, for each
$(a,b,c)\in\OmegaSet$,
\begin{align*}
  \sqrt{r_{abc}^2+\delta^2}
  \leq{}&
  \frac{r_{abc}^2}
  {2\sqrt{(r_{abc}^{k})^2+\delta^2}}
  +
  \sqrt{(r_{abc}^{k})^2+\delta^2}\\
  &\quad
  -
  \frac{(r_{abc}^{k})^2}
  {2\sqrt{(r_{abc}^{k})^2+\delta^2}}.
\end{align*}
After summing over the observed index set, all terms that do
not contain the new residual $r_{abc}$ are collected into
\begin{align*}
  C_k
  :={}&
  \sum_{\OmegaSet}
  \left[
    \sqrt{(r_{abc}^{k})^2+\delta^2}
    -
    \frac{(r_{abc}^{k})^2}
    {2\sqrt{(r_{abc}^{k})^2+\delta^2}}
  \right]\\
  ={}&
  \frac{1}{2}
  \sum_{\OmegaSet}
  \frac{(r_{abc}^{k})^2+2\delta^2}
  {\sqrt{(r_{abc}^{k})^2+\delta^2}}.
\end{align*}
Thus, the smoothed data fidelity term admits the quadratic
upper bound
\begin{equation}
  \sum_{\OmegaSet}
  \sqrt{r_{abc}^2+\delta^2}
  \leq
  C_k
  +
  \frac{1}{2}
  \sum_{\OmegaSet}
  m_{abc}^{k} r_{abc}^2 ,
  \label{eq:app-vwsptf-data-upper}
\end{equation}
where
\begin{equation*}
  m_{abc}^{k}
  =
  \frac{1}{\sqrt{(r_{abc}^{k})^2+\delta^2}}
\end{equation*}
and $C_k$ depends only on
$(\mathcal{U}^{k},\mathcal{V}^{k})$ and is therefore independent
of the optimization variables $(\mathcal{U},\mathcal{V})$.

For the regularization term, the concavity of $\phi$ gives
the following supporting-line inequality. When $p<1$, the current
energy is positive by the choice of $k_0$; when $p=1$, $\phi'$ extends
continuously to $t=0$. Thus, for every retained column pair,
\begin{equation}
  \phi(t)
  \leq
  \phi(t^k)
  +
  \phi'(t^k)(t-t^k).
  \label{eq:app-vwsptf-reg-upper-single}
\end{equation}
Using \eqref{eq:vwsptf-phi-derivative}, the corresponding
column-pair weight is
\[
  \omega_j^{(i),k}
  =
  \phi'\!\left(t_j^{(i),k}\right),
\]
which is exactly the weight in
\eqref{eq:vwsptf-opt-reg-weight}. Applying
\eqref{eq:app-vwsptf-reg-upper-single} to every column pair and
rearranging gives
\begin{align*}
  \phi\!\left(t_j^{(i)}\right)
  \leq{}&
  \phi\!\left(t_j^{(i),k}\right)
  +
  \omega_j^{(i),k}
  \left(t_j^{(i)}-t_j^{(i),k}\right)\\
  ={}&
  \omega_j^{(i),k}t_j^{(i)}
  +
  \phi\!\left(t_j^{(i),k}\right)
  -
  \omega_j^{(i),k}t_j^{(i),k}.
\end{align*}
Therefore, after summing over all transform-domain slices and
column pairs, the terms independent of the new variables are
collected into
\[
  C_k'
  :=
  \frac{1}{\ell}
  \sum_{i=1}^{n_3}
  \sum_{j=1}^{r}
  \left[
    \phi\!\left(t_j^{(i),k}\right)
    -
    \omega_j^{(i),k}t_j^{(i),k}
  \right].
\]
Hence,
\begin{equation}
  R_{p,q,\epsilon}(\mathcal{U},\mathcal{V})
  \leq
  C_k'
  +
  \frac{1}{\ell}
  \sum_{i=1}^{n_3}
  \sum_{j=1}^{r}
  \omega_j^{(i),k} t_j^{(i)},
  \label{eq:app-vwsptf-reg-upper}
\end{equation}
where $C_k'$ depends only on
$(\mathcal{U}^{k},\mathcal{V}^{k})$ and is therefore independent
of the optimization variables.

Combining \eqref{eq:app-vwsptf-data-upper} and
\eqref{eq:app-vwsptf-reg-upper}, and recalling the coefficient
$\lambda$ in $F_{\delta}$, gives the surrogate
\begin{equation}
\begin{aligned}
  G^k(\mathcal{U},\mathcal{V})
  :={}&
  C_k+\lambda C_k'
  +
  \frac{1}{2}
  \sum_{\OmegaSet}
  m_{abc}^{k}
  \left[
    (\mathcal{U}*\mathcal{V})_{abc}-\Y_{abc}
  \right]^2\\
  &+
  \frac{\lambda}{\ell}
  \sum_{i=1}^{n_3}
  \sum_{j=1}^{r}
  \omega_j^{(i),k}t_j^{(i)}\\
  ={}&
  C_k+\lambda C_k'
  +
  \frac{1}{2}
  \sum_{\OmegaSet}
  m_{abc}^{k}
  \left[
    (\mathcal{U}*\mathcal{V})_{abc}-\Y_{abc}
  \right]^2\\
  &+
  \frac{\lambda}{2\ell}
  \sum_{i=1}^{n_3}
  \sum_{j=1}^{r}
  \omega_j^{(i),k}
  \left(
    \left\|\bar{\bm{u}}_j^{(i)}\right\|_2^2
    +
    \left\|\bar{\bm{v}}_j^{(i)}\right\|_2^2
  \right).
\end{aligned}
\label{eq:app-vwsptf-surrogate}
\end{equation}
The preceding scalar tangent inequalities hold globally and become
equalities at the current residuals and column-pair energies.
Therefore, the explicitly constructed surrogate satisfies
\begin{equation}
\begin{aligned}
  G^k(\mathcal{U},\mathcal{V})
  &\geq
  F_{\delta}(\mathcal{U},\mathcal{V}),
  \qquad
  \forall\,(\mathcal{U},\mathcal{V}),\\
  G^k(\mathcal{U}^{k},\mathcal{V}^{k})
  &=
  F_{\delta}(\mathcal{U}^{k},\mathcal{V}^{k}).
\end{aligned}
\label{eq:app-vwsptf-surrogate-properties}
\end{equation}
Moreover, $\delta>0$ implies that every $m_{abc}^k$ is finite,
and Assumption~\ref{ass:vwsptf-convergence} guarantees that every
active $\omega_j^{(i),k}$ is finite. For fixed $k$, the residuals
are bilinear functions of $(\mathcal{U},\mathcal{V})$ and the
quantities $t_j^{(i)}$ are quadratic functions of the corresponding
column pairs. Hence the explicit expression
\eqref{eq:app-vwsptf-surrogate} shows that $G^k$ is continuous.

It remains to verify tangency in the first derivative. For the data
term, at the current residual $r_{abc}^k$,
\[
  \left.
  \frac{\mathrm{d}}{\mathrm{d}r}
  \left(
    \frac{1}{2}m_{abc}^k r^2
  \right)
  \right|_{r=r_{abc}^k}
  =
  m_{abc}^k r_{abc}^k
  =
  \left.
  \frac{\mathrm{d}}{\mathrm{d}r}
  \sqrt{r^2+\delta^2}
  \right|_{r=r_{abc}^k}.
\]
For the regularizer, the affine tangent majorizer satisfies
\[
  \left.
  \frac{\mathrm{d}}{\mathrm{d}t}
  \left[
    \phi(t_j^{(i),k})
    +
    \phi'(t_j^{(i),k})
    (t-t_j^{(i),k})
  \right]
  \right|_{t=t_j^{(i),k}}
  =
  \phi'(t_j^{(i),k}).
\]
Applying the chain rule to the residual and column-pair energy maps
gives
\begin{equation}
  \nabla G^k(\mathcal{U}^k,\mathcal{V}^k)
  =
  \nabla F_{\delta}(\mathcal{U}^k,\mathcal{V}^k).
  \label{eq:app-vwsptf-first-order-consistency}
\end{equation}
Thus, continuity, global majorization, tightness, and first-order
consistency follow directly from the specific construction of $G^k$.

We now connect the accepted slice matrices in
Algorithm~\ref{alg:vwsptf-irls-bsum} to the block curvatures used in the
analysis. Under column-wise vectorization in the transform domain,
right multiplication by $\mathbf H_U^{(i),k}$ and left multiplication by
$\mathbf H_V^{(i),k}$ induce
\begin{equation*}
\begin{aligned}
\mathbf B_U^k
&=
\operatorname{blkdiag}_{i=1}^{n_3}
\left(
\mathbf H_U^{(i),k}\otimes\mathbf I_{n_1}
\right),\\
\mathbf B_V^k
&=
\operatorname{blkdiag}_{i=1}^{n_3}
\left(
\mathbf I_{n_2}\otimes\mathbf H_V^{(i),k}
\right).
\end{aligned}
\end{equation*}
By the transform normalization relation in \eqref{eq:transform-normalization}, both the
input and output Frobenius norms acquire the same factor $1/\sqrt{\ell}$ when
mapped back to tensor coordinates. Hence the induced operator norms and the
spectral bounds used below are unchanged.

The positive lower curvature has a direct source in the active
regularization weights. On the fixed-dimensional tail, boundedness of the
factor sequence gives a finite upper bound on all retained pair energies.
If $p<1$, Assumption~\ref{ass:vwsptf-convergence} also bounds them below by
$\underline t>0$; if $p=1$, $\phi'$ is positive and continuous at zero.
Since \eqref{eq:vwsptf-phi-derivative} is strictly positive on the relevant
compact energy interval, there exists $\underline\omega>0$ such that every
active $\omega_j^{(i),k}\ge\underline\omega$. Hence
\[
\mathbf H_U^{(i),k}\succeq
\lambda\underline\omega\,\mathbf I,
\qquad
\mathbf H_V^{(i),k}\succeq
\lambda\underline\omega\,\mathbf I.
\]
Together with the accepted upper-curvature bounds in
Assumption~\ref{ass:vwsptf-convergence}, this yields constants
$0<\mu_U\le L_U<\infty$ and $0<\mu_V\le L_V<\infty$ satisfying
\[
\mu_U\mathbf I\preceq\mathbf B_U^k\preceq L_U\mathbf I,
\qquad
\mu_V\mathbf I\preceq\mathbf B_V^k\preceq L_V\mathbf I.
\]

Let
\[
\Delta\mathcal U^k=\mathcal U^{k+1}-\mathcal U^k,
\qquad
\Delta\mathcal V^k=\mathcal V^{k+1}-\mathcal V^k,
\]
and define the intermediate point
\[
\mathcal Z^{k+\frac12}
=
(\mathcal U^{k+1},\mathcal V^k).
\]
With the weights fixed during the $k$th outer cycle, the actual damped
updates \eqref{eq:vwsptf-u-update}--\eqref{eq:vwsptf-v-update} are exactly
\begin{equation}
\Delta\mathcal U^k
=
-\eta_U(\mathbf B_U^k)^{-1}
\nabla_{\mathcal U}G^k(\mathcal U^k,\mathcal V^k),
\label{eq:app-vwsptf-u-damped-operator}
\end{equation}
and
\begin{equation}
\Delta\mathcal V^k
=
-\eta_V(\mathbf B_V^k)^{-1}
\nabla_{\mathcal V}G^k(\mathcal Z^{k+\frac12}).
\label{eq:app-vwsptf-v-damped-operator}
\end{equation}

For the $\mathcal U$ block, write the accepted quadratic upper bound as
\[
\begin{aligned}
Q_U^k(\mathcal U\mid\mathcal V^k)
={}&
G^k(\mathcal U^k,\mathcal V^k)
+
\left\langle
\nabla_{\mathcal U}G^k(\mathcal U^k,\mathcal V^k),
\mathcal U-\mathcal U^k
\right\rangle\\
&+
\frac12
\|\mathcal U-\mathcal U^k\|_{\mathbf B_U^k}^2 .
\end{aligned}
\]
Substituting \eqref{eq:app-vwsptf-u-damped-operator} gives
\begin{align*}
&Q_U^k(\mathcal U^{k+1}\mid\mathcal V^k)
-
Q_U^k(\mathcal U^k\mid\mathcal V^k)\notag\\
&\qquad=
-\left(\frac1{\eta_U}-\frac12\right)
\|\Delta\mathcal U^k\|_{\mathbf B_U^k}^2\notag\\
&\qquad\le
-\frac{\mu_U(2-\eta_U)}{2\eta_U}
\|\Delta\mathcal U^k\|_F^2 .
\end{align*}
Here $0<\eta_U\le1$ makes the coefficient strictly positive. Since the
accepted quadratic majorizes the $\mathcal U$ block of $G^k$ and is tight
at $\mathcal U^k$,
\begin{equation}
G^k(\mathcal U^k,\mathcal V^k)
-
G^k(\mathcal Z^{k+\frac12})
\ge
\frac{c_U}{2}
\|\Delta\mathcal U^k\|_F^2,
\label{eq:app-vwsptf-u-surrogate-descent}
\end{equation}
where
\[
c_U=\mu_U\frac{2-\eta_U}{\eta_U}>0.
\]

Likewise, the accepted $\mathcal V$-block quadratic is
\[
\begin{aligned}
Q_V^k(\mathcal V\mid\mathcal Z^{k+\frac12})
={}&
G^k(\mathcal Z^{k+\frac12})
+
\left\langle
\nabla_{\mathcal V}G^k(\mathcal Z^{k+\frac12}),
\mathcal V-\mathcal V^k
\right\rangle\\
&+
\frac12
\|\mathcal V-\mathcal V^k\|_{\mathbf B_V^k}^2 .
\end{aligned}
\]
Using \eqref{eq:app-vwsptf-v-damped-operator},
\begin{align}
&G^k(\mathcal Z^{k+\frac12})
-
G^k(\mathcal U^{k+1},\mathcal V^{k+1})\notag\\
&\qquad\ge
\frac{c_V}{2}
\|\Delta\mathcal V^k\|_F^2,
\label{eq:app-vwsptf-v-sufficient-decrease}
\end{align}
with
\[
c_V=\mu_V\frac{2-\eta_V}{\eta_V}>0.
\]
Thus the quantitative decrease constants are determined by the accepted
curvature lower bounds and the retained damping parameters; they are not
additional sufficient-descent assumptions.

Finally, global majorization and tightness
\eqref{eq:app-vwsptf-surrogate-properties} give
$F_\delta(x^{k+1})\le G^k(x^{k+1})$ and
$G^k(x^k)=F_\delta(x^k)$. Adding
\eqref{eq:app-vwsptf-u-surrogate-descent} and
\eqref{eq:app-vwsptf-v-sufficient-decrease} therefore yields
\begin{equation}
\begin{aligned}
F_\delta(x^k)-F_\delta(x^{k+1})
\ge{}&
\frac{c_U}{2}\|\Delta\mathcal U^k\|_F^2\\
&+
\frac{c_V}{2}\|\Delta\mathcal V^k\|_F^2.
\end{aligned}
\label{eq:app-vwsptf-total-sufficient-decrease}
\end{equation}
This is \eqref{eq:vwsptf-main-sufficient-decrease}. Since
$F_\delta\ge0$, summing \eqref{eq:app-vwsptf-total-sufficient-decrease}
from $k_0$ to $N$ and then letting $N\to\infty$ gives
\[
\sum_{k=k_0}^{\infty}
\left(
\|\Delta\mathcal U^k\|_F^2+
\|\Delta\mathcal V^k\|_F^2
\right)<\infty.
\]
Consequently,
\begin{equation}
\|\Delta\mathcal U^k\|_F\to0,
\qquad
\|\Delta\mathcal V^k\|_F\to0.
\label{eq:app-vwsptf-asymptotic-regularity}
\end{equation}
The objective values are nonincreasing on the tail and bounded below, so
they converge. A finite prefix does not affect this conclusion.

\subsection{Proof of Theorem~\ref{thm:vwsptf-stationary}}

Choose $k_0$ large enough that pruning has stopped and, when $p<1$, all
retained pair energies satisfy the nondegeneracy condition in
Assumption~\ref{ass:vwsptf-convergence}. Theorem~\ref{thm:vwsptf-monotonic}
and \eqref{eq:app-vwsptf-asymptotic-regularity} give
\[
\|\Delta\mathcal U^k\|_F\to0,
\qquad
\|\Delta\mathcal V^k\|_F\to0.
\]
Thus the asymptotic regularity needed below is a consequence of the actual
damped updates.

Let
\[
x^k=(\mathcal U^k,\mathcal V^k),
\qquad
\mathcal Z^{k+\frac12}
=
(\mathcal U^{k+1},\mathcal V^k).
\]
For the $\mathcal U$ block, define the first-order residual of the accepted
quadratic at the damped point by
\[
\begin{aligned}
\bm e_U^k
&:=
\nabla_{\mathcal U}
Q_U^k(\mathcal U^{k+1}\mid\mathcal V^k)\\
&=
\nabla_{\mathcal U}G^k(x^k)
+
\mathbf B_U^k\Delta\mathcal U^k.
\end{aligned}
\]
Using \eqref{eq:app-vwsptf-u-damped-operator},
\begin{equation*}
\bm e_U^k
=
-\frac{1-\eta_U}{\eta_U}
\mathbf B_U^k\Delta\mathcal U^k.
\end{equation*}
Hence
\[
\|\bm e_U^k\|_F
\le
\frac{1-\eta_U}{\eta_U}
L_U\|\Delta\mathcal U^k\|_F
\longrightarrow0.
\]
In particular, residual vanishing is derived from the damped update and
Theorem~\ref{thm:vwsptf-monotonic}, rather than assumed.

Similarly, for the $\mathcal V$ block,
\[
\begin{aligned}
\bm e_V^k
&:=
\nabla_{\mathcal V}
Q_V^k
\left(
\mathcal V^{k+1}
\,\middle|\,
\mathcal Z^{k+\frac12}
\right)\\
&=
\nabla_{\mathcal V}
G^k(\mathcal Z^{k+\frac12})
+
\mathbf B_V^k\Delta\mathcal V^k,
\end{aligned}
\]
and \eqref{eq:app-vwsptf-v-damped-operator} gives
\begin{equation*}
\bm e_V^k
=
-\frac{1-\eta_V}{\eta_V}
\mathbf B_V^k\Delta\mathcal V^k.
\end{equation*}
Therefore
\[
\|\bm e_V^k\|_F
\le
\frac{1-\eta_V}{\eta_V}
L_V\|\Delta\mathcal V^k\|_F
\longrightarrow0.
\]

For the $\mathcal U$ block, rearranging the residual identity and using
first-order consistency
\eqref{eq:app-vwsptf-first-order-consistency} yields
\[
\begin{aligned}
\|\nabla_{\mathcal U}F_\delta(x^k)\|_F
&=
\|\nabla_{\mathcal U}G^k(x^k)\|_F\\
&\le
\|\bm e_U^k\|_F
+
L_U\|\Delta\mathcal U^k\|_F
\longrightarrow0.
\end{aligned}
\]
For the $\mathcal V$ block, the residual relation first gives
\[
\left\|
\nabla_{\mathcal V}G^k(\mathcal Z^{k+\frac12})
\right\|_F
\le
\|\bm e_V^k\|_F
+
L_V\|\Delta\mathcal V^k\|_F
\longrightarrow0.
\]
The uniform block-Lipschitz property in
Assumption~\ref{ass:vwsptf-convergence} then implies, for some
$L_G<\infty$ independent of $k$,
\[
\left\|
\nabla_{\mathcal V}G^k(x^k)
-
\nabla_{\mathcal V}G^k(\mathcal Z^{k+\frac12})
\right\|_F
\le
L_G\|\Delta\mathcal U^k\|_F
\longrightarrow0.
\]
Combining this with
\eqref{eq:app-vwsptf-first-order-consistency} gives
\[
\|\nabla_{\mathcal V}F_\delta(x^k)\|_F\longrightarrow0.
\]

Now let $x^{k_j}\to x^\star
=(\mathcal U^\star,\mathcal V^\star)$ be any accumulation subsequence,
whose existence follows from boundedness. If $p<1$, continuity of the
retained energy maps and
$t_j^{(i),k_j}\ge\underline t$ imply that all retained energies remain
strictly positive in a neighborhood of $x^\star$; if $p=1$, $\phi'$ is
continuous also at zero. Since $\delta>0$, the smoothed data term is
continuously differentiable everywhere. Hence $F_\delta$ is continuously
differentiable on a neighborhood of $x^\star$.

Passing to the limit along the subsequence gives
\[
\nabla_{\mathcal U}F_\delta(x^\star)=\mathbf0,
\qquad
\nabla_{\mathcal V}F_\delta(x^\star)=\mathbf0.
\]
Therefore $\nabla F_\delta(x^\star)=\mathbf0$. For a continuously
differentiable function, the limiting subdifferential is the singleton
containing its gradient, so
\[
\mathbf0\in
\partial F_\delta(\mathcal U^\star,\mathcal V^\star).
\]
This proves the theorem.




\begin{thebibliography}{99}

\bibitem{RechtFazelParrilo2010}
B.~Recht, M.~Fazel, and P.~A. Parrilo,
``Guaranteed minimum-rank solutions of linear matrix equations via nuclear
norm minimization,'' \emph{SIAM Rev.}, vol.~52, no.~3, pp.~471--501,
2010, doi: 10.1137/070697835.

\bibitem{NieEtAl2012Schatten}
F.~Nie, H.~Huang, and C.~Ding, ``Low-rank matrix recovery via
efficient Schatten-$p$ norm minimization,'' in \emph{Proc. AAAI Conf.
Artif. Intell.}, 2012, pp.~655--661.

\bibitem{ShangEtAl2016Tractable}
F.~Shang, Y.~Liu, and J.~Cheng, ``Tractable and scalable Schatten
quasi-norm approximations for rank minimization,'' in
\emph{Proc. Artif. Intell. Statist.}, 2016, pp.~620--629.

\bibitem{XuEtAl2017UnifiedConvex}
C.~Xu, Z.~Lin, and H.~Zha, ``A unified convex surrogate for the
Schatten-$p$ norm,'' in \emph{Proc. AAAI Conf. Artif. Intell.}, vol.~31,
no.~1, 2017, pp.~926--932, doi: 10.1609/aaai.v31i1.10646.

\bibitem{KilmerMartin2011}
M.~E. Kilmer and C.~D. Martin, ``Factorization strategies for
third-order tensors,'' \emph{Linear Algebra Appl.}, vol.~435, no.~3,
pp.~641--658, 2011, doi: 10.1016/j.laa.2010.09.020.

\bibitem{KernfeldEtAl2015}
E.~Kernfeld, M.~Kilmer, and S.~Aeron, ``Tensor--tensor products with
invertible linear transforms,'' \emph{Linear Algebra Appl.}, vol.~485,
pp.~545--570, 2015, doi: 10.1016/j.laa.2015.07.021.

\bibitem{ZhangAeron2017}
Z.~Zhang and S.~Aeron, ``Exact tensor completion using t-SVD,''
\emph{IEEE Trans. Signal Process.}, vol.~65, no.~6, pp.~1511--1526,
2017, doi: 10.1109/TSP.2016.2639466.

\bibitem{LuEtAl2020TRPCA}
C.~Lu, J.~Feng, Y.~Chen, W.~Liu, Z.~Lin, and S.~Yan, ``Tensor robust
principal component analysis with a new tensor nuclear norm,''
\emph{IEEE Trans. Pattern Anal. Mach. Intell.}, vol.~42, no.~4,
pp.~925--938, 2020, doi: 10.1109/TPAMI.2019.2891760.

\bibitem{ShangEtAl2016Schatten}
F.~Shang, Y.~Liu, and J.~Cheng, ``Scalable algorithms for tractable
Schatten quasi-norm minimization,'' in \emph{Proc. AAAI Conf. Artif.
Intell.}, vol.~30, no.~1, 2016, pp.~2016--2022,
doi: 10.1609/aaai.v30i1.10266.

\bibitem{GuEtAl2014WNNM}
S.~Gu, L.~Zhang, W.~Zuo, and X.~Feng, ``Weighted nuclear norm
minimization with application to image denoising,'' in
\emph{Proc. IEEE Conf. Comput. Vis. Pattern Recognit.}, 2014,
pp.~2862--2869, doi: 10.1109/CVPR.2014.366.

\bibitem{LuEtAl2016IRNN}
C.~Lu, J.~Tang, S.~Yan, and Z.~Lin, ``Nonconvex nonsmooth low-rank
minimization via iteratively reweighted nuclear norm,''
\emph{IEEE Trans. Image Process.}, vol.~25, no.~2, pp.~829--839,
2016, doi: 10.1109/TIP.2015.2511584.

\bibitem{XieEtAl2016WSNM}
Y.~Xie, S.~Gu, Y.~Liu, W.~Zuo, W.~Zhang, and L.~Zhang, ``Weighted
Schatten $p$-norm minimization for image denoising and background
subtraction,'' \emph{IEEE Trans. Image Process.}, vol.~25, no.~10,
pp.~4842--4857, 2016, doi: 10.1109/TIP.2016.2599290.

\bibitem{YangEtAl2022WeightedTensorSchatten}
M.~Yang, Q.~Luo, W.~Li, and M.~Xiao, ``Nonconvex 3D array image data
recovery and pattern recognition under tensor framework,''
\emph{Pattern Recognit.}, vol.~122, Art. no.~108311, 2022,
doi: 10.1016/j.patcog.2021.108311.

\bibitem{SrebroEtAl2004MMMF}
N.~Srebro, J.~D.~M. Rennie, and T.~S. Jaakkola,
``Maximum-margin matrix factorization,'' in \emph{Adv. Neural Inf. Process. Syst.},
vol.~17, 2004, pp.~1329--1336.

\bibitem{CabralEtAl2013}
R.~Cabral, F.~De~la~Torre, J.~P. Costeira, and A.~Bernardino,
``Unifying nuclear norm and bilinear factorization approaches for
low-rank matrix decomposition,'' in \emph{Proc. IEEE Int. Conf.
Comput. Vis.}, 2013, pp.~2488--2495,
doi: 10.1109/ICCV.2013.309.

\bibitem{HastieEtAl2015ALS}
T.~Hastie, R.~Mazumder, J.~D. Lee, and R.~Zadeh,
``Matrix completion and low-rank SVD via fast alternating least squares,''
\emph{J. Mach. Learn. Res.}, vol.~16, no.~104, pp.~3367--3402, 2015.

\bibitem{ShangEtAl2020Unified}
F.~Shang, Y.~Liu, F.~Shang, H.~Liu, L.~Kong, and L.~Jiao,
``A unified scalable equivalent formulation for Schatten quasi-norms,''
\emph{Mathematics}, vol.~8, no.~8, Art. no.~1325, 2020,
doi: 10.3390/math8081325.

\bibitem{ZhouEtAl2018TensorFactorization}
P.~Zhou, C.~Lu, Z.~Lin, and C.~Zhang,
``Tensor factorization for low-rank tensor completion,''
\emph{IEEE Trans. Image Process.}, vol.~27, no.~3, pp.~1152--1163,
2018, doi: 10.1109/TIP.2017.2762595.

\bibitem{ChenEtAl2022RLTF}
L.~Chen, X.~Jiang, X.~Liu, and M.~Haardt, ``Reweighted low-rank
factorization with deep prior for image restoration,''
\emph{IEEE Trans. Signal Process.}, vol.~70, pp.~3514--3529, 2022,
doi: 10.1109/TSP.2022.3183466.

\bibitem{JiangEtAl2023FactorTensorNorm}
W.~Jiang, J.~Zhang, C.~Zhang, L.~Wang, and H.~Qi,
``Robust low tubal rank tensor completion via factor tensor norm
minimization,'' \emph{Pattern Recognit.}, vol.~135, Art.~no.~109169, 2023,
doi: 10.1016/j.patcog.2022.109169.

\bibitem{HeAtia2023Correntropy}
Y.~He and G.~K. Atia,
``Robust low-tubal-rank tensor completion based on tensor factorization and
maximum correntropy criterion,'' \emph{IEEE Trans. Neural Netw. Learn. Syst.},
vol.~35, no.~10, pp.~14603--14617, 2024,
doi: 10.1109/TNNLS.2023.3280086.

\bibitem{LiuEtAl2024FGD}
Z.~Liu, Z.~Han, Y.~Tang, X.-L.~Zhao, and Y.~Wang,
``Low-tubal-rank tensor recovery via factorized gradient descent,''
\emph{IEEE Trans. Signal Process.}, vol.~72, pp.~5470--5483, 2024,
doi: 10.1109/TSP.2024.3504292.

\bibitem{DuEtAl2021UTF}
S.~Du, Q.~Xiao, Y.~Shi, R.~Cucchiara, and Y.~Ma,
``Unifying tensor factorization and tensor nuclear norm approaches for
low-rank tensor completion,'' \emph{Neurocomputing}, vol.~458,
pp.~204--218, 2021, doi: 10.1016/j.neucom.2021.06.020.

\bibitem{GiampourasEtAl2020Variational}
P.~Giampouras, R.~Vidal, A.~Rontogiannis, and B.~D. Haeffele,
``A novel variational form of the Schatten-$p$ quasi-norm,'' in
\emph{Adv. Neural Inf. Process. Syst.}, vol.~33, 2020.

\bibitem{GiampourasEtAl2019AIRLS}
P.~V. Giampouras, A.~A. Rontogiannis, and K.~D. Koutroumbas,
``Alternating iteratively reweighted least squares minimization for low-rank
matrix factorization,'' \emph{IEEE Trans. Signal Process.}, vol.~67, no.~2,
pp.~490--503, 2019, doi: 10.1109/TSP.2018.2883921.

\bibitem{FanEtAl2019FGSR}
J.~Fan, L.~Ding, Y.~Chen, and M.~Udell, ``Factor group-sparse
regularization for efficient low-rank matrix recovery,'' in
\emph{Adv. Neural Inf. Process. Syst.}, vol.~32, 2019,
pp.~5104--5114.

\bibitem{OrnhagEtAl2020Bilinear}
M.~V. \"Ornhag, C.~Olsson, and A.~Heyden,
``Bilinear parameterization for differentiable rank-regularization,'' in
\emph{Proc. IEEE/CVF Conf. Comput. Vis. Pattern Recognit. Workshops},
2020, pp.~346--347.

\bibitem{WangEtAl2024SparsityTensor}
Z.-Y.~Wang, H.~C.~So, and A.~M.~Zoubir,
``Low-rank tensor completion via novel sparsity-inducing regularizers,''
\emph{IEEE Trans. Signal Process.}, vol.~72, pp.~3519--3534, 2024,
doi: 10.1109/TSP.2024.3424272.


\bibitem{CandesEtAl2011RPCA}
E.~J. Cand\`es, X.~Li, Y.~Ma, and J.~Wright, ``Robust principal
component analysis?'' \emph{J. ACM}, vol.~58, no.~3, Art. no.~11,
2011, doi: 10.1145/1970392.1970395.

\bibitem{LiuEtAl2014ExactRank}
L.~Liu, W.~Huang, and D.-R. Chen, ``Exact minimum rank approximation
via Schatten-$p$ norm minimization,'' \emph{J. Comput. Appl. Math.},
vol.~267, pp.~218--227, 2014,
doi: 10.1016/j.cam.2014.02.018.

\bibitem{ZhangEtAl2018LRRSchatten}
H.~Zhang, J.~Yang, F.~Shang, C.~Gong, and Z.~Zhang, ``LRR for
subspace segmentation via tractable Schatten-$p$ norm minimization
and factorization,'' \emph{IEEE Trans. Cybern.}, vol.~49, no.~5,
pp.~1722--1734, 2019, doi: 10.1109/TCYB.2018.2811761.

\bibitem{Foucart2018ConcaveMirsky}
S.~Foucart, ``Concave Mirsky inequality and low-rank recovery,''
\emph{SIAM J. Matrix Anal. Appl.}, vol.~39, no.~1, pp.~99--103, 2018,
doi: 10.1137/16M1090004.

\bibitem{PengEtAl2014Reweighted}
Y.~Peng, J.~Suo, Q.~Dai, and W.~Xu, ``Reweighted low-rank matrix
recovery and its application in image restoration,'' \emph{IEEE
Trans. Cybern.}, vol.~44, no.~12, pp.~2418--2430, 2014,
doi: 10.1109/TCYB.2014.2305919.

\bibitem{HuangEtAl2020GIRNN}
Y.~Huang, G.~Liao, Y.~Xiang, L.~Zhang, J.~Li, and A.~Nehorai,
``Low-rank approximation via generalized reweighted iterative nuclear
and Frobenius norms,'' \emph{IEEE Trans. Image Process.}, vol.~29,
pp.~2244--2257, 2020, doi: 10.1109/TIP.2019.2949383.

\bibitem{LuPengWei2019TransformTNN}
C.~Lu, X.~Peng, and Y.~Wei, ``Low-rank tensor completion with a new tensor
nuclear norm induced by invertible linear transforms,'' in \emph{Proc. IEEE/CVF
Conf. Comput. Vis. Pattern Recognit. (CVPR)}, 2019, pp.~5996--6004,
doi: 10.1109/CVPR.2019.00615.

\bibitem{LiuEtAl2025TTSP}
J.~Liu, L.~Lu, P.~Wang, H.~Liu, Y.~Huang, and Y.~Zang,
``Tensor truncated Schatten-$p$ norm approximation tensor completion
algorithm,'' \emph{IET Image Process.}, vol.~19, no.~1, Art.~no.~e70171,
2025, doi: 10.1049/ipr2.70171.

\bibitem{RennieSrebro2005}
J.~D. Rennie and N.~Srebro, ``Fast maximum margin matrix
factorization for collaborative prediction,'' in \emph{Proc. Int.
Conf. Mach. Learn.}, 2005, pp.~713--719,
doi: 10.1145/1102351.1102441.

\bibitem{WenEtAl2012LowRankFactorization}
Z.~Wen, W.~Yin, and Y.~Zhang, ``Solving a low-rank factorization
model for matrix completion by a nonlinear successive over-relaxation
algorithm,'' \emph{Math. Program. Comput.}, vol.~4, no.~4,
pp.~333--361, 2012, doi: 10.1007/s12532-012-0044-1.

\bibitem{FengEtAl2013ORPCA}
J.~Feng, H.~Xu, and S.~Yan,
``Online robust PCA via stochastic optimization,'' in
\emph{Adv. Neural Inf. Process. Syst.}, vol.~26, 2013.

\bibitem{ShangEtAl2015RBF}
F.~Shang, Y.~Liu, H.~Tong, J.~Cheng, and H.~Cheng,
``Robust bilinear factorization with missing and grossly corrupted
observations,'' \emph{Inf. Sci.}, vol.~307, pp.~53--72, 2015,
doi: 10.1016/j.ins.2015.02.026.

\bibitem{SunLuo2016Factorization}
R.~Sun and Z.-Q. Luo,
``Guaranteed matrix completion via non-convex factorization,''
\emph{IEEE Trans. Inf. Theory}, vol.~62, no.~11, pp.~6535--6579, 2016.

\bibitem{HaeffeleVidal2020}
B.~D. Haeffele and R.~Vidal, ``Structured low-rank matrix factorization:
Global optimality, algorithms, and applications,''
\emph{IEEE Trans. Pattern Anal. Mach. Intell.}, vol.~42, no.~6,
pp.~1468--1482, 2020, doi: 10.1109/TPAMI.2019.2900306.

\bibitem{TaoEtAl2022ColumnL20}
T.~Tao, Y.~Qian, and S.~Pan,
``Column $\ell_{2,0}$-norm regularized factorization model of low-rank matrix
recovery and its computation,'' \emph{SIAM J. Optim.}, vol.~32, no.~2,
pp.~959--988, 2022, doi: 10.1137/20M136205X.

\bibitem{LiuEtAl2020WeightedTSchatten}
M.~Liu, X.~Zhang, and L.~Tang, ``Weighted t-Schatten-$p$ norm
minimization for real color image denoising,'' \emph{IEEE Access},
vol.~8, pp.~150350--150359, 2020,
doi: 10.1109/ACCESS.2020.3016777.

\bibitem{HornJohnson1991}
R.~A. Horn and C.~R. Johnson, \emph{Topics in Matrix Analysis}.
Cambridge, U.K.: Cambridge Univ. Press, 1991,
doi: 10.1017/CBO9780511840371.

\bibitem{Thompson1976}
R.~C. Thompson, ``Convex and concave functions of singular values of
matrix sums,'' \emph{Pacific J. Math.}, vol.~66, no.~1,
pp.~285--290, 1976.

\bibitem{MadathilGeorge2018DCT}
B.~Madathil and S.~N. George, ``DCT based weighted adaptive multi-linear data
completion and denoising,'' \emph{Neurocomputing}, vol.~318, pp.~120--136,
2018, doi: 10.1016/j.neucom.2018.08.038.

\bibitem{XuZhaoNg2019DCT}
W.-H.~Xu, X.-L.~Zhao, and M.~K. Ng, ``A fast algorithm for cosine transform
based tensor singular value decomposition,'' arXiv:1902.03070, 2019.

\bibitem{LiuEtAl2013SNN}
J.~Liu, P.~Musialski, P.~Wonka, and J.~Ye, ``Tensor completion for
estimating missing values in visual data,'' \emph{IEEE Trans. Pattern
Anal. Mach. Intell.}, vol.~35, no.~1, pp.~208--220, 2013,
doi: 10.1109/TPAMI.2012.39.

\bibitem{ShuEtAl2026GTNLN}
H.~Shu, J.~Li, T.~Lei, and L.~Sun,
``Robust tensor completion via gradient tensor nuclear $\ell_1$--$\ell_2$
norm for traffic data recovery,'' \emph{IEEE Trans. Intell. Transp. Syst.},
2026, doi: 10.1109/TITS.2026.3659755.
\end{thebibliography}
\end{document}